\documentclass[ijoc,sglanonrev]{informs4_arXiv}
\usepackage{eqndefns-left} % For checking the display equation width and equation environment definitions %
\RequirePackage{tgtermes}
\RequirePackage{newtxtext}
\RequirePackage{newtxmath}
\RequirePackage{bm}
\RequirePackage{endnotes}

\OneAndAHalfSpacedXII % Current default line spacing
\usepackage{amsfonts}
\usepackage{indentfirst}
\usepackage{subfigure}
\usepackage{caption}
\usepackage{hyperref}
\usepackage{booktabs}
\usepackage{multirow}
\usepackage{threeparttable}
\usepackage{verbatim}
\usepackage{pifont}
\newcommand{\minimize}{\mathop{\mathrm{min}}}

\usepackage{algorithm}
\usepackage{algpseudocode}
\usepackage{tikz}
\usepackage{natbib}
 \bibpunct[, ]{(}{)}{,}{a}{}{,}%
 \def\bibfont{\small}%
\EquationsNumberedThrough    % Default: (1), (2), ...
\TheoremsNumberedThrough     % Preferred (Theorem 1, Lemma 1, Theorem 2)
\ECRepeatTheorems  %  

\MANUSCRIPTNO{IJOC-0001-2024.00}

\begin{document}
%%%%%%%%%%%%%%%%

% Outcomment only when entries are known. Otherwise leave as is and
%   default values will be used.
%\setcounter{page}{1}
%\VOLUME{00}%
%\NO{0}%
%\MONTH{Xxxxx}% (month or a similar seasonal id)
%\YEAR{0000}% e.g., 2005
%\FIRSTPAGE{000}%
%\LASTPAGE{000}%
%\SHORTYEAR{00}% shortened year (two-digit)
%\ISSUE{0000} %
%\LONGFIRSTPAGE{0001} %
%\DOI{10.1287/xxxx.0000.0000}%

% Author's names for the running heads
% Sample depending on the number of authors;
% \RUNAUTHOR{Jones}
% \RUNAUTHOR{Jones and Wilson}
% \RUNAUTHOR{Jones, Miller, and Wilson}
% \RUNAUTHOR{Jones et al.} % for four or more authors
% Enter authors following the given pattern:
%\RUNAUTHOR{}
\RUNAUTHOR{Wang, Y., et al.}

% Title or shortened title suitable for running heads. Sample:
% \RUNTITLE{Predictive Maintenance in Manufacturing}
% Enter the (shortened) title:
\RUNTITLE{Matrix Completion with Graph Information: Nonconvex Optimization Approach}

% Full title. Sample:
% \TITLE{Optimal Resource Allocation in Humanitarian Logistics: A Stochastic Programming Approach}
% Enter the full title:
\TITLE{Matrix Completion with Graph Information: A Provable Nonconvex Optimization Approach}

% Block of authors and their affiliations starts here:
% NOTE: Authors with same affiliation, if the order of authors allows,
%   should be entered in ONE field, separated by a comma.
%   \EMAIL field can be repeated if more than one author
\ARTICLEAUTHORS{%
%\AUTHOR{John Doe,\textsuperscript{a} Jane Smith,\textsuperscript{b}}
%\AFF{\textsuperscript{a}Department of Industrial Engineering, University of XYZ, \EMAIL{john.doe@xyz.edu; \textsuperscript{b}Department of Computer Science, University of ABC, \EMAIL{jane.smith@abc.edu}} 
\AUTHOR{Yao Wang}
\AFF{School of Management, Xi'an Jiaotong University, Xi'an, China, \EMAIL{yao.s.wang@gmail.com}}

\AUTHOR{Yiyang Yang}
\AFF{School of Management, Xi'an Jiaotong University, Xi'an, China, \EMAIL{yyyang817@gmail.com}}

\AUTHOR{Kaidong Wang}
\AFF{School of Management, Xi'an Jiaotong University, Xi'an, China, \EMAIL{wangkd13@gmail.com}}

\AUTHOR{Shanxing Gao}
\AFF{School of Management, Xi'an Jiaotong University, Xi'an, China, \EMAIL{gaozn@mail.xjtu.edu.cn}}

\AUTHOR{Xiuwu Liao}
\AFF{School of Management, Xi'an Jiaotong University, Xi'an, China, \EMAIL{liaoxiuwu@mail.xjtu.edu}}
% Enter all authors
} % end of the block

\ABSTRACT{%
% Enter your abstract
We consider the problem of matrix completion with graphs as side information depicting the interrelations between variables. The key challenge lies in leveraging the graph’s similarity structure to enhance matrix recovery. Existing approaches, primarily based on graph Laplacian regularization, suffer from several limitations: (1) they focus only  on the similarity between neighboring variables, while overlooking  long-range correlations; (2) they are highly sensitive to false edges in the graphs and (3) they lack theoretical guarantees regarding statistical and computational complexities. To address these issues, we propose in this paper a novel graph regularized matrix completion algorithm called GSGD, based on preconditioned projected gradient descent approach. We demonstrate that GSGD effectively captures the higher-order correlation information behind the graphs, and achieves superior robustness and stability against the false edges. Theoretically, we prove that GSGD achieves linear convergence to the global optimum  with near-optimal sample complexity, providing the first theoretical guarantees for both recovery accuracy and  efficacy in the perspective of nonconvex optimization. Our numerical experiments on both synthetic and real-world data further validate that GSGD achieves superior recovery accuracy and scalability compared with several popular alternatives.}%

%\FUNDING{This research was supported by [grant number, funding agency].}

%Supplemental Material:
%Data Ethics & Reproducibility Note:

% Sample
%\KEYWORDS{Stochastic programming, Decision support,Uncertainty, Disaster response, Optimization}

% Fill in data. If unknown, outcomment the field
%\HISTORY{Received: Month DD, YYYY; Accepted: Month DD, YYYY; Published Online: Month DD, YYYY}

\maketitle
%%%%%%%%%%%%%%%%%%%%%%%%%%%%%%%%%%%%%%%%%%%%%%%%%%%%%%%%%%%%%%%%%%%%%%

% Text of your paper here

\section{Introduction}\label{introduction}
Aiming to recover the missing
entries from partial observations, low-rank matrix completion has attracted increasing attentions in recent years,  and been successfully applied across various domains such as recommender systems \citep{muter2017incorporating}, bioinformatics \citep{chen2018predicting}, and intelligent transportation system \citep{lei2022bayesian}.
The classical low-rank matrix completion problem can be formulated as recovering a rank-$ r $ matrix $ X\in \mathbb{R}^{m\times n} $ for which only a subset of its entries $ X_{ij}, \forall (i,j)\in \Omega $, are observed, where $ \Omega \subset \{1,2,...,m\}\times \{1,2,...,n\} $ is the set of known entries' positions and $ |\Omega|  \ll mn $. Taking the low-rankness and consistency with the partial observations into consideration, a general formulation for matrix completion is given by the following rank minimization form: 
	$ \min_{Z\in \mathbb{R}^{m\times n}} \text{rank}(Z)   ~~\text{s.t.} ~Z_{ij} = X_{ij},  \forall (i,j)\in \Omega $,
which can further be relaxed by a convex nuclear norm based approach:
	$ \min_{Z\in \mathbb{R}^{m\times n}} \|Z\|_*   ~~\text{s.t.} ~Z_{ij} = X_{ij},  \forall (i,j)\in \Omega $,
where $ \|\cdot\|_* $ denotes the nuclear norm. This convex relaxation provides significant convenience for algorithm design and theoretical analysis, while the per-iteration cost of computing SVD (Singular Value Decomposition) may increase rapidly as the dimension of the problem increases, making the algorithms rather slow  for problems with large size.  

Following its success in the Netflix competition, matrix factorization has gained widespread popularity, particularly in recommender systems. A popular factorization based formulation for matrix completion can be stated as:
$
	\min_{W\in \mathbb{R}^{m\times r}, H\in \mathbb{R}^{n\times r}} \frac{1}{2}\sum_{(i,j)\in \Omega} \big(X_{ij} - (WH^T)_{ij}\big)^2,
$
where $ W $ and $ H $ are commonly interpreted as the latent feature matrices of variables (e.g., users, items). This model is a nonconvex fourth-order polynomial optimization problem, and can be solved to stationary points by standard optimization algorithms such as alternating minimization \citep{jain2013low} and gradient descent method \citep{sun2016guaranteed}. Factorization-based algorithms can achieve good performance and high efficiency, particularly for large-scale problems, as they significantly reduce per-iteration computation costs and storage requirements. However, the theoretical understanding of these algorithms remains limited, largely due to the challenges of nonconvex optimization. Only recently, with the development of new analytical tools, has there been a growing interest in advancing the theory and algorithms of nonconvex optimization \citep{chi2019nonconvex}.

In many real-world scenarios, in addition to the partial observations of the underlying matrix data, we also have access
to supplementary information about the variables involved,
known as side information \citep{farias2019learning}. Generally speaking, common side
information can be broadly categorized into two types, i.e.,
features \citep{bertsimas2023interpretable} and graphs \citep{banerjee2016online}. Features capture the attributes of each variable (e.g., directors and genres of movies), while graphs represent the relationships between variables (e.g., the social network of users). It is natural to utilize side information as prior knowledge to enhance the prediction accuracy of matrix completion. Actually, over the past few years there has been considerable research on investigating matrix completion with features, which is usually called  inductive matrix completion \citep{zilber2022inductive}. In contrast, research on the graph side information is relatively limited \citep{dong2021riemannian}. This may be due to that the complex topological structure of graphs poses significant challenges for the quality measurement and analysis of the graph information.

This research focuses on the problem of matrix completion with graph information, which can be formulated as recovering a rank-$ r $ matrix $ X\in \mathbb{R}^{m\times n} $ from its partial observations $ X_{ij}, \forall (i,j)\in \Omega $, where we additionally have access to the similarity graphs $ G_1 = (V_1, E_1), G_2 = (V_2, E_2) $  representing the correlations among the rows and columns of $ X $, respectively. Obviously, the core of this problem lies in effectively characterizing the graph smoothness of the matrix, namely, two
rows (columns) connected by an edge in the graph are ``close'' to each other in the Euclidean distance.  For a long time, graph Laplacian regularization has served as the standard approach for incorporating graph information into matrix recovery problems~\citep{rao2015collaborative, dong2021riemannian}. Nevertheless, this approach has some inherent limitations: (1) it only captures the first-order smoothness of graphs, without considering higher-order smoothness, i.e., long-range correlations among variables; (2) it is sensitive to the noise (false edge), putting a high demand on the quality of the graph; (3) relevant research generally lacks theoretical guarantees regarding statistical and computational complexities. 

To address the aforementioned limitations, we propose a new graph regularized matrix completion algorithm which demonstrates superior effectiveness and efficiency compared  to the graph Laplacian regularization based methods. Precisely, we define a new matrix that explicitly captures the higher-order correlation information underlying the similarity graph, based on which we derive a preconditional projected gradient descent  algorithm incorporating higher-order graph information. Our main contributions can be summarized as follows:

1. Algorithmically, the proposed method fully exploits the higher-order smoothness of the graph to enhance the recovery performance, while achieving better robustness and stability against false edges in the graph. Additionally, we introduce a new initialization method that incorporates graph information, effectively enhancing the convergence speed of the algorithm.

2. Theoretically, we establish the first theoretical guarantee in terms of statistical and iteration complexities from the perspective of nonconvex optimization,  effectively bridging a gap in the theoretical examination of the problem involving
matrix recovery with graph information. The theoretical results demonstrate that  the proposed algorithm achieves a linear convergence rate independent of the condition number of the low-rank matrix  at near-optimal sample complexity, which provides theoretical guarantees for both recovery accuracy and efficacy. The core of our analysis lies in the innovative introduction of a rigorous quality measure for similarity graphs and a graph incoherence condition to prevent  ill-posedness and  ensure reliable estimation of the low-rank matrix, offering general tools that can be applied to other related problems involving graph information.

3. Experimentally, we examine the performance of the proposed algorithm in extensive numerical experiments including synthetic and two large-scale real-world data sets. We demonstrate the strong capability and stability in exploiting the graph information, the robustness against false edges, and the effectiveness of higher-order graph smoothness and new initialization approach of our method. Furthermore, we highlight the superior recovery accuracy and scalability of the proposed algorithm to some state-of-the-art methods, including graph regularized and graph-agnostic ones.

The outline of this paper is as follows. Section \ref{section algorithm} introduces the proposed algorithm, detailing the update rules, projection operator, and the new initialization method. Section \ref{section theoretical analysis} establishes theoretical guarantees for the algorithm in terms of both statistical and iteration complexities. Sections \ref{section synthetic data} and \ref{section realworld data} evaluate the recovery performance and speed of the proposed algorithm on synthetic and large-scale real-world data sets, highlighting its superior effectiveness and  efficiency compared to state-of-the-art methods. Section \ref{section conclusion} concludes this paper and discusses some potential extensions.

\subsection{Relevant Literature}

\subsubsection{Nonconvex Optimization Based Matrix Completion}
Generally, matrix completion methods based on nonconvex optimization primarily rely on two strategies: alternating minimization and gradient descent. \citep{jain2013low}  provides the first global optimality guarantees with a linear convergence rate based on alternating minimization. Their theoretical results were later improved and extended in \citep{hardt2014fast, zhao2015nonconvex}. 
\citep{sun2016guaranteed} provides the first theoretical analysis demonstrating the linear convergence of the gradient descent approach for $ \ell_{2,\infty} $-norm regularized matrix factorization problems. The $ \ell_{2,\infty} $-norm regularization or projection has become a standard assumption for nonconvex matrix completion ever since to encourage an incoherent solution, e.g., \citep{chen2015fast} and \citep{zheng2016convergence} provide the theoretical guarantees for projected gradient descent to linearly converge to the global optimum. It is worth noting that, the
iteration complexity of these gradient descent approaches scales at least linearly with respect to the condition
number $ \kappa $ of the low-rank matrix, e.g. $ O( \kappa\log(\frac{1}{\epsilon})) $, to reach $ \epsilon $-accuracy, and thus converge slowly for ill-conditioned matrices. In contrast,  alternating minimization  converges at the rate $ O(\log(\frac{1}{\epsilon})) $  independent of $ \kappa $, while the per-iteration computation cost is significantly higher.  Recently, \citep{tong2021accelerating} proposed a new preconditioned gradient decent approach, achieving iteration complexity $ O(\log(\frac{1}{\epsilon})) $ similar as alternating minimization, while maintaining the low per-iteration cost of gradient descent. It is the first algorithm
that provably exhibits such properties across a wide range of low-rank matrix estimation tasks.
\subsubsection{Matrix Completion with Graph Information} The existing matrix completion methods utilizing graph information can be divided into convex optimization-based and nonconvex optimization-based approaches. 
Among the former, a notable work is \citep{kalofolias2014matrix} which introduced a convex optimization model by incorporating graph Laplacian regularization into the nuclear norm minimization problem. Building upon this,  \citep{zhao2014expert} proposed an accelerated proximal gradient approach to solve the graph Laplacian regularized nuclear norm minimization model for question answering problem. Recently, nonconvex optimization methods have gained prominence due to their lower computational cost. An early pioneering work  is \citep{zhou2012kernelized}, which developed a kernelized probabilistic matrix factorization method incorporating external graph information. Following that, \citep{rao2015collaborative} developed a highly scalable algorithm based on alternating minimization to solve the graph Laplacian regularized matrix factorization model and provided a statistical consistency guarantee. However, their theoretical analysis relies on a convex reformulation of the original nonconvex matrix factorization model, resulting in a disconnect between the theory and the algorithm. More recently, \citep{dong2021riemannian} introduced a preconditioned gradient descent algorithm which leverages Riemannian geometry to determine descent directions, achieving faster convergence compared to its counterparts.

Our work differs from prior works in two key aspects: (1) From a theoretical perspective, we provide a pioneering analysis of the statistical and computational complexities of our algorithm, which is the first theoretical guarantee for graph regularized matrix recovery methods within the framework of nonconvex optimization. (2) From the perspective of algorithm design, our method is the first to move beyond conventional graph Laplacian regularization by considering higher-order smoothness and robustness of graphs, leading to enhanced recovery performance and efficiency.
\subsection{Notation}
We use uppercase letters to denote matrices. For any matrix $ A $, we use $ A_{ij} $ to denote its $ (i,j) $-th element, and $ A_{i:} $ and $ A_{:j} $ to denote the $ i $-th row and $ j $-th column of $ A $, respectively. $ \|A\|_F $,$ \|A\|_{\text{op}} $, $ \|A\|_{2,\infty} $, and $ \text{tr}(A) $ denote the Frobenius norm, the spectral norm (i.e., the largest singular value), the $ \ell_{2,\infty} $ norm (i.e., the largest $ \ell_2 $ norm of the rows), and the trace of the matrix $ A $.  $ f(n) = O(g(n)) $ and $ f(n) \gtrsim g(n) $  mean $ |f(n)|/|g(n)| \leq C $ and $ |f(n)|/|g(n)| \geq C $   for some constant $ C>0 $ when $ n $ is sufficiently large, respectively. We use the terminology “with overwhelming probability” to
denote the event happens with probability at least $ 1-c_1n^{-c_2} $, where $ c_1,  c_2>0 $ are some universal
 constants.  For two numbers $ a $ and $ b $, let $ a \vee b = \max \{a,b\}  $ and $ a \wedge b = \min \{a,b\}  $.
For the rank-$ r $ matrix $ X_\star\in \mathbb{R}^{m\times n} $, denote $ U_\star \Sigma_\star V^T_\star $ as its compact singular value decomposition (SVD), where $ U_\star \in \mathbb{R}^{m\times r} $ and  $ V_\star \in \mathbb{R}^{n\times r} $ are orthogonal matrices consisting of $ r $ left and right singular vectors of $  X_\star $, respectively, and $ \Sigma_\star\in \mathbb{R}^{r\times r} $ is a diagonal matrix containing the $ r $ nonzero singular values of $ X_\star $ in non-increasing order, i.e., $ \sigma_1(X_\star)\geq \sigma_2(X_\star)\geq \cdots \geq \sigma_r(X_\star) > 0 $. We define the condition number of $ X_\star $ as 
	$ \kappa := \sigma_1(X_\star)/\sigma_r(X_\star) $,
and the ground truth low-rank factors of $ X_\star $ as
	$ W_\star := U_\star \Sigma_\star^{1/2},
	H_\star := V_\star \Sigma_\star^{1/2} $,
so that $ X_\star = W_\star H_\star^T $. We define a stacked factor matrix as
	$ F_\star :
	=[W^T_\star, H^T_\star]^T \in \mathbb{R}^{(m+n)\times r} $.

\section{Algorithm}
\label{section algorithm}
In this paper, we investigate the graph regularized matrix completion problem which aims to recover a rank-$ r $ matrix $ X_\star\in \mathbb{R}^{m\times n} $ from partial observations $ X_{ij}, \forall (i,j)\in \Omega $, leveraging additional graphs $ G_1 = (V_1, E_1) $ and $ G_2 = (V_2, E_2) $ that encodes the similarity structure among the rows and columns of $ X_\star$, respectively. Supposing that each $ (i,j) \in \Omega $ is independently sampled with probability $ p $, we define the orthogonal projection operator $ \mathcal{P}_\Omega(\cdot) $ which retains only the entries of the matrix lying in the set $ \Omega $, i.e., $ \mathcal{P}_\Omega(X)_{ij} := X_{ij} $ for $ (i, j) \in \Omega $ and $ 0 $ otherwise,
then a graph-agnostic matrix completion model based on matrix factorization can be built as
\begin{equation}\label{matrix decomposition model}
	\minimize_{W\in \mathbb{R}^{m\times r},H\in \mathbb{R}^{n\times r}} \mathcal{L}(W,H) = \frac{1}{2p}\|\mathcal{P}_\Omega(WH^T - X_\star)\|_F^2.
\end{equation}
Given an initialization $ (W_0, H_0) $, (\ref{matrix decomposition model}) can be solved by gradient descent (GD) algorithm as follows:
\begin{equation}\label{GD update}
	\begin{aligned}
		W_{t+1} &= W_t - \eta \nabla_W\mathcal{L}(W_t, H_t)= W_t - \frac{\eta}{p} \mathcal{P}_\Omega(W_tH_t^T - X_\star)H_t,\\
		H_{t+1} &= H_t - \eta \nabla_H\mathcal{L}(W_t, H_t) = H_t - \frac{\eta}{p} \mathcal{P}_\Omega(W_tH_t^T - X_\star)^TW_t,
	\end{aligned}
\end{equation}
where $ \eta > 0 $ is the step size, and $ \nabla_W\mathcal{L}(W_t, H_t) $ and $ \nabla_H\mathcal{L}(W_t, H_t) $ are the gradients of the loss function $ \mathcal{L}(W,H) $ with respect to the factor matrices $ W_t $ and $ H_t $ at the $ t $-th iteration, respectively. Notably, \citep{tong2021accelerating} introduced a preconditioned gradient descent algorithm, Scaled Gradient Descent (ScaledGD), to solve model (\ref{matrix decomposition model}), significantly accelerating the convergence of vanilla GD algorithm. The update rules of ScaledGD for solving (\ref{matrix decomposition model}) are given as follows:
\begin{equation}\label{ScaledGD update}
	\begin{aligned}	
		W_{t+1} &= W_t - \eta \nabla_W\mathcal{L}(W_t, H_t)(H_t^TH_t)^{-1}= W_t - \frac{\eta}{p} \mathcal{P}_\Omega(W_tH_t^T - X_\star)H_t(H_t^TH_t)^{-1},\\
		H_{t+1} &= H_t - \eta \nabla_H\mathcal{L}(W_t, H_t)(W_t^TW_t)^{-1} = H_t - \frac{\eta}{p} \mathcal{P}_\Omega(W_tH_t^T - X_\star)^TW_t(W_t^TW_t)^{-1},
	\end{aligned}
\end{equation}
where $ (H_t^TH_t)^{-1} $ and $ (W_t^TW_t)^{-1} $ act as preconditioners adjusting the search direction to allow larger step sizes. ScaledGD has been theoretically and empirically proven to achieve faster convergence,   sparking extensive subsequent research~\citep{tong2022scaling, jia2024preconditioning} and inspiring our method. 

As mentioned earlier, graph Laplacian regularization is widely employed to characterize the similarity structure among variables inherent in the graphs. Supposing that the Laplacian matrices of graphs $ G_1 $ and $ G_2 $ are $ \widetilde{L}_W\in \mathbb{R}^{m \times m} $ and $ \widetilde{L}_H \in \mathbb{R}^{n \times n}  $, respectively, graph Laplacian regularization for target matrix $ X = WH^T $ can be formulated as follows \citep{rao2015collaborative, dong2021riemannian}:
\begin{equation}\label{graph laplacian}
	\text{tr}(W^T\widetilde{L}_WW), ~~~\text{tr}(H^T\widetilde{L}_HH),
\end{equation}
where $ \text{tr}(\cdot) $ denotes the trace of a matrix.  The rationale behind the graph Laplacian regularization is as follows. Let $ L $ and $ A $ be the Laplacian matrix and adjacent matrix for a similarity graph, then it is easy to verify that the equation holds:
\begin{equation}\label{graph laplacian equation}
	\text{tr}(W^TLW) = \frac{1}{2}\sum_{i,j}A_{ij}(W_{i:}-W_{j:})^2,
\end{equation}
where $ W_{i:} $ and $ W_{j:} $ denote the $ i $-th and $ j $-th rows of $ W $, respectively. (\ref{graph laplacian equation}) indicates that lessening $ \text{tr}(W^TLW) $ enforces $ (W_{i:}-W_{j:})^2 $ to be smaller when $ A_{ij}=1 $, which aligns precisely with the intended effect of graph smoothness.

Graph Laplacian regularization (\ref{graph laplacian}) is often injected into matrix factorization model (\ref{matrix decomposition model}) to leverage the graph information for improved recovery, resulting in the following model:
\begin{equation}\label{matrix completion model}
	\minimize_{W\in \mathbb{R}^{m\times r},H\in \mathbb{R}^{n\times r}}  \frac{1}{2p}\|\mathcal{P}_\Omega(WH^T - X_\star)\|_F^2 + \frac{\beta}{2}\big(\text{tr}(W^T\widetilde{L}_WW) + \text{tr}(H^T\widetilde{L}_HH)\big),
\end{equation}
where $ \beta $ is the trade-off parameter. Model (\ref{matrix completion model}) can be efficiently solved using common nonconvex optimization approaches, such as alternating minimization~\citep{rao2015collaborative} and preconditioned gradient descent~\citep{ dong2021riemannian}. Despite its widespread applications, graph Laplacian regularization is hindered by its outlook of long-range correlations and sensitivity to noisy edges, limiting its full utilization of graph information. Furthermore, the lack of theoretical analysis for the associated optimization algorithms leaves both recovery performance and efficiency unguaranteed.

In the following, we propose a new preconditioned projected gradient descent algorithm for the graph regularized matrix completion problem. We begin by formulating the update rules based on preconditioned gradient descent and demonstrate that these rules introduce a novel graph smoothness regularization, which captures higher-order graph smoothness and offers robustness against false edges, outperforming traditional graph Laplacian regularization. Next, we establish a new graph incoherence condition along with a corresponding projection operation to prevent ill-posedness and ensure reliable estimation of the underlying low-rank matrix. Lastly, we design a novel initialization approach that incorporates graph information to accelerate convergence.

\subsection{Update Rules Based on Preconditioned Gradient Descent}
For the convenience of subsequent discussions, 
for similarity graphs $ G_1 $ and $ G_2 $ with Laplacian matrices $ \widetilde{L}_W $ and $ \widetilde{L}_H $, 
we define the corresponding higher-order graph matrices as follows:
\begin{equation}\label{LWLH form}
	L_W := (1+\beta)I_m -\beta \mathcal{A},~~L_H := (1+\beta)I_n - \beta \mathcal{B},
\end{equation}
where $ I_m $, $ I_n $ are $ m \times m $ and $ n \times n $ identity matrices, respectively, $ \lambda \geq 0 $, $ \beta \geq 0 $ are parameters, and matrices $  \mathcal{A} $ and $ \mathcal{B} $ are defined as $ \mathcal{A} := (I_m + \lambda \widetilde{L}_W)^{-1},	\mathcal{B} := (I_n + \lambda \widetilde{L}_H)^{-1} $, respectively. 
We propose a Graph regularized Scaled Gradient Descent (GSGD) algorithm based on the matrix factorization $ X = WH^T $, which leverages higher-order graph smoothness and demonstrates robustness to false edges. Starting from an initialization $ (W_0, H_0) $, GSGD updates the factor matrices at the $ t $-th iteration according to the following rules:
\begin{equation}\label{GSGD update}
	\begin{aligned}
		W_{t+1} & = W_t - \frac{\eta}{p}L_W\mathcal{P}_\Omega(W_tH_t^T - X_\star)H_t(H_t^TH_t)^{-1},\\
		H_{t+1} &= H_t - \frac{\eta}{p}L_H\mathcal{P}_\Omega(W_tH_t^T - X_\star)^TW_t(W_t^TW_t)^{-1}.
	\end{aligned}
\end{equation}
It can be observed that in the absence of graph information, i.e., $ \widetilde{L}_W = 0 $ and $ \widetilde{L}_H = 0 $, the matrices simplify to $ L_W = I_m $ and $ L_H = I_n $, in which case (\ref{GSGD update}) degenerates into the ScaledGD method  (\ref{ScaledGD update}). 

In the following, we analyze the advantages of the simple form (\ref{GSGD update}) over graph Laplacian regularization from the perspective of optimization objectives. Plugging into  the computational form of $ L_W $ and $ L_H $ from	 (\ref{LWLH form}), update rules (\ref{GSGD update}) can be decomposed into the following equivalent form:
\begin{equation}\label{GSGD update2}
	\begin{aligned}
		W_{t+1} & = W_t - \frac{\eta}{p}\mathcal{P}_\Omega(W_tH_t^T - X_\star)H_t(H_t^TH_t)^{-1} - \frac{\eta}{p}\beta(I_m-\mathcal{A})\mathcal{P}_\Omega(W_tH_t^T - X_\star)H_t(H_t^TH_t)^{-1},\\
		H_{t+1} &= H_t - \frac{\eta}{p}\mathcal{P}_\Omega(W_tH_t^T - X_\star)^TW_t(W_t^TW_t)^{-1} - \frac{\eta}{p}\beta(I_n-\mathcal{B})\mathcal{P}_\Omega(W_tH_t^T - X_\star)^TW_t(W_t^TW_t)^{-1}.
	\end{aligned}
\end{equation}
In (\ref{GSGD update2}), the first terms correspond exactly to the update rules (\ref{ScaledGD update}) of ScaledGD for solving the standard matrix completion model  (\ref{matrix decomposition model}), while the last terms play the role of reorienting the search directions based on the graph information. From the perspective of the target model, it is straightforward  to verify that at the $ (t+1) $-th iterates,  (\ref{GSGD update2}) is equivalent to update the factor matrices $ W_{t+1} $ and $ H_{t+1} $ from the current point $ (W_t, H_t) $ using ScaledGD to optimize the following regularized model:
\begin{equation}\label{GSGD model}
	\begin{aligned}
		&\minimize_{W\in \mathbb{R}^{m\times r}}  \underbrace{\frac{1}{2p}\|\mathcal{P}_\Omega(WH_t^T - X_\star)\|_F^2}_\text{loss function} + \underbrace{\frac{\beta}{2p}\text{tr}\big(\mathcal{P}_\Omega(WH_t^T - X_\star)^T(I_m - \mathcal{A})\mathcal{P}_\Omega(WH_t^T - X_\star)\big)}_\text{graph smoothness regularization},\\
		&\minimize_{H\in \mathbb{R}^{n\times r}}  \underbrace{\frac{1}{2p}\|\mathcal{P}_\Omega(W_tH^T - X_\star)\|_F^2}_\text{loss function} + \underbrace{\frac{\beta}{2p}\text{tr}\big(\mathcal{P}_\Omega(W_tH^T - X_\star)(I_n - \mathcal{B})\mathcal{P}_\Omega(W_tH^T - X_\star)^T\big)}_\text{graph smoothness regularization},
	\end{aligned}
\end{equation}
where $ \beta $ is  the trade-off parameter.
Taking the update of $ W $ as an example, compared to graph Laplacian regularization (\ref{graph laplacian}), (\ref{GSGD model}) introduces a novel graph smoothness regularization. This regularization leverages the graph matrix $ (I_m - \mathcal{A}) $ instead of $ \widetilde{L}_W $ to incorporate graph information and replaces $ W $ with $ \mathcal{P}_\Omega(WH_t^T - X_\star) $ to capture the similarity structure. In the following, we demonstrate how these modifications allow the new regularization to exploit higher-order graph smoothness and enhance robustness to false edges in the graph.

\subsubsection{\boldmath {$ (I_m - \mathcal{A}) $} VS. \boldmath {$ \widetilde{L}_W $}: Higher-order Graph Smoothness}\label{Higher-order Graph Smoothness}
Since $ \widetilde{L}_W $ only considers the existence of direct edges between nodes, graph Laplacian regularization merely enforces proximity between rows that are directly connected, while neglecting long-range effects. We refer to this property as \textit{first-order smoothness}.
In practical applications, first-order smoothness fails to fully exploit graph information due to its binary treatment of edges. A more reasonable approach is to account for the relationship between paths and the degree of association between nodes: shorter paths imply stronger associations, while longer paths suggest weaker ones. In terms of graph smoothness, this implies that the similarity between two rows of the matrix should be inversely correlated with the distance between their corresponding nodes.  Distinguished from  first-order smoothness, we define this property as \textit{higher-order smoothness}.
\begin{figure*}[t]
	\centering 
	\includegraphics[width=0.9\linewidth]{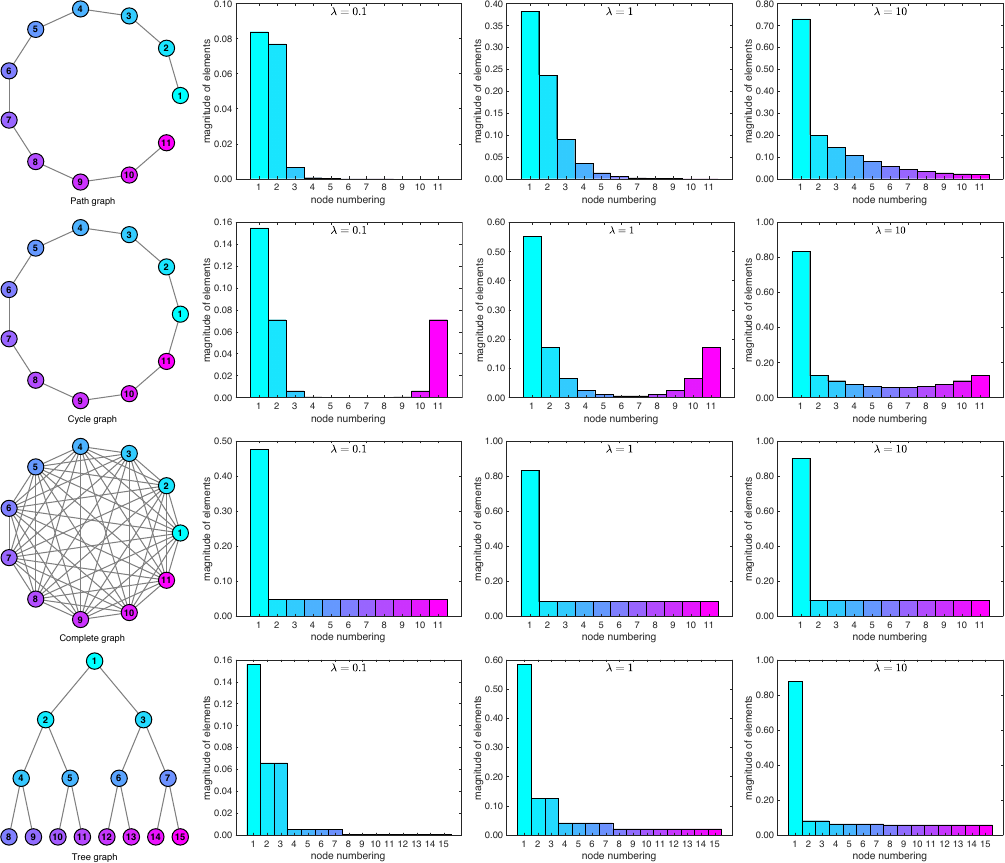}
	\vspace{0mm}
	\caption{Several representative graphs and the corresponding bar plots of $ |(I_M - \mathcal{A})_{1:}| $, i.e., the first row of matrix $ |(I_M - \mathcal{A})| $, with different values of $ \lambda $, where different colors and numbers represent the node identifiers..
	}\label{higher order graph}
	\vspace{0mm}
\end{figure*}

To leverage higher-order smoothness, $ \widetilde{L}_W $ should be replaced with a new matrix which captures the paths between nodes. It can be verified that the elements in the matrix $ (I_m - \mathcal{A}) $ exhibit the following desirable properties:\\
(1) $ (I_m - \mathcal{A})_{ij} \leq 0 $ if $ i \neq j $, and $ (I_m - \mathcal{A})_{ij} > 0 $ if $ i = j $;\\
(2) the sum of elements in each row  is zero, i.e., 
	$ (I_m - \mathcal{A})_{ii} = \sum_{j\neq i}|(I_m - \mathcal{A})_{ij}|, i=1,2,\cdots  $;\\
(3) For two distinct nodes $ i $ and $ j $, the magnitude of $ |(I_m - \mathcal{A})_{ij}| $ is inversely correlated with the distance between them.\\
These properties demonstrate that $ (I_m - \mathcal{A}) $ can serve as an alternative to the Laplacian matrix $ \widetilde{L}_W $ to capture higher-order information. To visualize these properties of $ (I_m - \mathcal{A}) $,  in Figure \ref{higher order graph} we present bar plots of $ |(I_m - \mathcal{A})_{1:}| $, i.e., the first row of the matrix $ |(I_m - \mathcal{A})| $, for several representative graphs and different values of $ \lambda $, where different colors and numbers represent the node identifiers. From Figure \ref{higher order graph}, the following observations can be made:
(1) For each fixed $ \lambda $, for any $ j \neq 1 $, the closer node $ j $ is to node $ 1 $, the larger the corresponding magnitude $ |(I_m - \mathcal{A})_{1j}| $, which effectively captures the higher-order information of node $ 1 $.
(2) The parameter $ \lambda $ controls the degree of association between distant nodes. When $ \lambda $ is small, only nodes very close to node $ 1 $ exhibit larger corresponding magnitudes, while others remain close to zero, which is consistent with first-order information. Conversely, when $ \lambda $ is large, even nodes relatively far from node $ 1 $ have magnitudes greater than zero, reflecting interactions between distant nodes.

With this, we define a higher-order adjacent matrix $ E $ as $ E_{ij} =  |(I_m - \mathcal{A})_{ij}| $ for $ j \neq i $ and $ E_{ij} = 0 $ for $ j=i $, 
then for a matrix $ M $, we have
$
	\text{tr}(M^T(I_m - \mathcal{A}) M) = \frac{1}{2}\sum_{i,j}E_{ij}\|M_{i:} - M_{j:}\|_2^2.
$
Obviously,  lessening $ \text{tr}(M^T(I_m - \mathcal{A}) M)  $ enforces $ \|M_{i:} - M_{j:}\|_2^2 $ to be smaller for larger $ E_{ij} $, which corresponds to smaller distances between nodes $ i $ and $ j $. As a consequence, the higher-order smoothness can be achieved, with the parameter $ \lambda $ governing the extent of higher-order graph information exploitation.

\subsubsection{\boldmath {$ \mathcal{P}_\Omega(WH_t^T - X_\star) $} VS. \boldmath {$ W $}: Robustness to False Edges}\label{analysis robust}
In practical scenarios, accessible graphs are often affected by false edges caused by external interference, posing significant challenges to the graph Laplacian regularization method. This issue arises because rows connected by false edges are typically dissimilar or even highly divergent, yet graph Laplacian regularization compels these rows to be close, resulting in degraded recovery performance. By characterizing the similarity of rows in matrix $ \mathcal{P}_\Omega(WH_t^T - X_\star) $, the new graph smoothness regularization in (\ref{GSGD model}) can be equivalently expressed as:
\begin{equation*}
	\begin{aligned}
		&\text{tr}( \mathcal{P}_\Omega(WH_t^T - X_\star)^T(I_M - \mathcal{A})\mathcal{P}_\Omega(WH_t^T - X_\star))\\
		&= \frac{1}{2}\sum_{i,j}E_{ij}\big\|\big(\mathcal{P}_\Omega(WH_t^T)_{i:} - \mathcal{P}_\Omega(WH_t^T)_{j:}\big) - \big(\mathcal{P}_\Omega(X_\star)_{i:} - \mathcal{P}_\Omega(X_\star)_{j:}\big)\big\|_2^2,
	\end{aligned}
\end{equation*}
which indicates that for large $ E_{ij} $, the new regularization enforces $ \big(\mathcal{P}_\Omega(WH_t^T)_{i:} - \mathcal{P}_\Omega(WH_t^T)_{j:}\big) $ to closely approximate $ \big(\mathcal{P}_\Omega(X_\star)_{i:} - \mathcal{P}_\Omega(X_\star)_{j:}\big) $. For real edges, we have $ \mathcal{P}_\Omega(X_\star)_{i:} \approx \mathcal{P}_\Omega(X_\star)_{j:} $. Consequently, the new regularization enforces $ \mathcal{P}_\Omega(WH_t^T)_{i:} \approx \mathcal{P}_\Omega(WH_t^T)_{j:} $, which is consistent with the behavior of graph Laplacian regularization regardless of $ \mathcal{P}_\Omega(\cdot) $ and $ H_t $. However, for false edges, the new regularization aligns the difference $ \mathcal{P}_\Omega(WH_t^T)_{i:} - \mathcal{P}_\Omega(WH_t^T)_{j:} $ with $ \mathcal{P}_\Omega(X_\star)_{i:} - \mathcal{P}_\Omega(X_\star)_{j:} $ rather than forcing it to $ 0 $, thereby significantly mitigating the adverse effects of false edges.
\begin{figure*}[t]
	\centering 
	\includegraphics[width=0.9\linewidth]{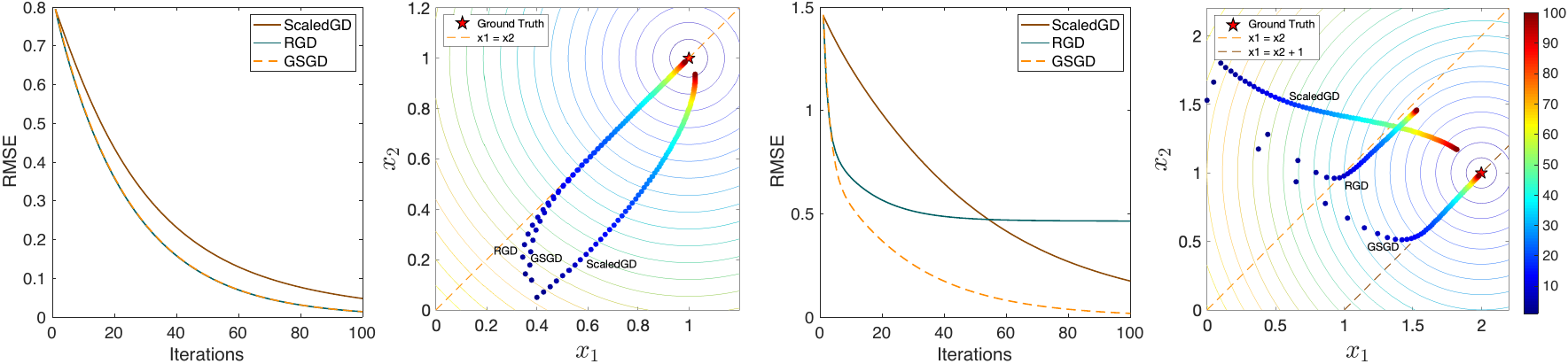}
	\vspace{0mm}
	\caption{The recovery RMSE and trajectory with respect to iterations of ScaledGD, RGD and GSGD on the toy matrix factorization problem, where the color of the trajectory points changes gradually with the number of iterations from $ 1 $ to $ 100 $. Left two: $ x_1 = x_2 = 1 $; right two: $ x_1 = 2 $, $ x_2 = 1 $.
	}\label{opt path}
	\vspace{0mm}
\end{figure*}

As a toy experimental verification, we consider a simple matrix factorization problem: factorizing a target matrix $ X_\star = [x_1, x_2]^T
\in \mathbb{R}^2 $ into factors $ W \in \mathbb{R}^2 $ and $ H \in \mathbb{R} $, with access to a similarity graph $ G $ along the rows of $ X_\star $. Denote the Laplacian matrix of $ G $ as $ L = \begin{bmatrix}
	1& -1 \\ -1 &1
\end{bmatrix} $, indicating that the two rows of $ X_\star $ are connected by an edge. To evaluate the robustness of graph Laplacian regularization and our method, we consider the following two cases: (1) $ x_1 = x_2 = 1 $, where the edge in $ G $ is clearly real; (2) $ x_1 = 2 $, $ x_2 = 1 $, indicating that the edge in $ G $ is false, as the corresponding nodes are not close at all. Given a random initialization, we implement ScaledGD, RGD, and GSGD, where ScaledGD and RGD serve as representatives of the graph-agnostic and graph Laplacian regularized method, respectively. We plot the corresponding recovery RMSE and iteration trajectory in Figure \ref{opt path}. We observe the following: (1) In the case of $ x_1 = x_2 = 1 $ (real edge), there is no significant difference between RGD and GSGD. Both algorithms approach the ground truth along the straight line $ x_1 = x_2 $ and outperform ScaledGD, demonstrating the utility of graph information. (2) In the case of $ x_1 = 2 $, $ x_2 = 1 $ (false edge), RGD is evidently misled by the incorrect graph information, continuing along the straight line $ x_1 = x_2 $. In contrast, GSGD automatically adjusts its route to rapidly approach the ground truth along the straight line $ x_1 = x_2 + 1 $,  highlighting its robustness to false edges. 

\subsection{Graph Incoherence and New Projection Operator}
It has been demonstrated  that if the underlying matrix $ X_\star $ contains mostly zero rows or columns, completing $ X_\star $ becomes impossible unless all its entries are observed \citep{candes2012exact}. To avoid this ill-posedness, it is now standard practice to assume that $ X_\star $ satisfies additional properties  referred to as incoherence \citep{chen2015incoherence}. Standard incoherence condition is defined as follows:
\begin{definition}[Standard incoherence, \citep{chen2015incoherence}]
	A rank-$ r $ matrix $ X_\star\in\mathbb{R}^{m \times n} $ with compact SVD $ X_\star = U_\star\Sigma V_\star^T $ is said to be $ \mu $-incoherent if 
		$ \|U_\star\|_{2,\infty} \leq \sqrt{\frac{\mu r}{m}}, ~\|V_\star\|_{2,\infty} \leq \sqrt{\frac{\mu r}{n}} $.
\end{definition}

Noting that $ \sqrt{\frac{\mu r}{m}} = \sqrt{\frac{\mu }{m}}\|U_\star\|_F $ and $ \sqrt{\frac{\mu r}{n}} = \sqrt{\frac{\mu }{n}}\|V_\star\|_F $, the standard incoherence condition ensures that the information of the row and column spaces of the matrix is not overly concentrated in a few entries. Taking the graph structure of $ X_\star $ into consideration, we extend the standard incoherence condition to the following graph incoherence condition:
\begin{definition}[Graph incoherence]
	A rank-$ r $ matrix $ X_\star\in\mathbb{R}^{m \times n} $ with compact SVD $ X_\star = U_\star\Sigma V_\star^T $ and higher-order graph matrices $ L_W $, $ L_H $ is said to be $ \mu $-graph incoherent if 
		$ \|L_W^{\frac{1}{2}}U_\star\|_{2,\infty} \leq \sqrt{\frac{\mu r}{m}}, ~\|L_H^{\frac{1}{2}}V_\star\|_{2,\infty} \leq \sqrt{\frac{\mu r}{n}} $.
\end{definition}

To enforce the incoherence condition, a common strategy in gradient methods is to perform projection after each gradient updates to maintain small $ \ell_{2,\infty} $ norms of the factor matrices \citep{chen2015fast, tong2021accelerating}. Specifically for our graph regularization algorithm and graph incoherence condition, we first define a new error metric  (i.e., Lyapunov function) to measure the distance between the iterates and the ground truth, based on which we introduce a new projection operator to ensure compliance with the graph incoherence condition. Clearly, considering the update form (\ref{GSGD update}), the new distance metric should properly take the effect of graph information $ L_W $, $ L_H $ and preconditioning $ (W_t^TW_t)^{-1} $, $ (H_t^TH_t)^{-1} $ into account. Furthermore, since the factored representation $ WH^T $ is indistinguishable with respect to an invertible matrix $ Q $, i.e., $ WH^T = (WQ)(HQ^{-T})^T $, the definition of distance metric should also account for the issue of non-uniqueness in factorization.
Guided by these considerations,  we define the following new distance metric:
\begin{definition}[Graph-aware distance metric]
	Given the ground truth stacked factor matrix $ F_\star := 
	[W_\star^T, H_\star^T]^T
	\in \mathbb{R}^{(m+n)\times r}  $, and let $ \text{GL}(r) $  denote the set of invertible matrices in $ \mathbb{R}^{r \times r} $, the distance metric between any factor matrix $ F := 
	[W^T, H^T]^T
	\in \mathbb{R}^{(m+n)\times r}  $ and $ F_\star $ is defined as follows:
	\begin{equation}\label{error metric}
		\text{dist}^2(F,F_\star) := \inf_{Q\in \text{GL}(r)} \big\|L_W^{1/2}(WQ-W_\star)\Sigma_\star^{1/2}\big\|_F^2 + \big\|L_H^{1/2}(HQ^{-T}-H_\star)\Sigma_\star^{1/2}\big\|_F^2.
	\end{equation} 
\end{definition}

The error metric (\ref{error metric}) defines a quadratic distance scaled by $ L_W $, $ L_H $ and $ \Sigma_\star $, where $ L_W $ and $ L_H $ evaluates the higher-order graph smoothness of the factor matrices, and $ \Sigma_\star $ accounts for the preconditioning, as the preconditioners in (\ref{ScaledGD update}) can be approximated by $ W_t^TW_t \approx \Sigma_\star $ and $ H_t^TH_t \approx \Sigma_\star $ for $ W_t \approx W_\star $ and $ H_t \approx H_\star $ in the vicinity of the ground truth.
The design of the new error metric (\ref{error metric}) incorporates both preconditioning and graph information, playing a crucial role in the subsequent algorithmic analysis.  In comparison, the
previously studied distance metrics either omit the diagonal
scaling~\citep{zheng2016convergence} (mainly for GD), or disregard the effect of graph information~\citep{tong2021accelerating} (mainly for ScaledGD), which fail to reveal the advantage of GSGD.

Based on the new distance metric and graph incoherence condition, we then introduce a new projection operator $ \mathcal{P}_B(\cdot) $ for any $ \widetilde{F} = 	[\widetilde{W}^T, \widetilde{H}^T]^T\in \mathbb{R}^{(m+n)\times r} $ as follows:
\begin{equation}\label{projection operator}
	\begin{aligned}
		\mathcal{P}_B(\widetilde{F}) &:= \argmin_{F\in \mathbb{R}^{(m+n)\times r}} \big\|L_W^{\frac{1}{2}}(W - \widetilde{W})(\widetilde{H}^T\widetilde{H})^{\frac{1}{2}}\big\|_F^2 + \big\|L_H^{\frac{1}{2}}(H - \widetilde{H})(\widetilde{W}^T\widetilde{W})^{\frac{1}{2}}\big\|_F^2\\
		&\text{s.t.} ~ \sqrt{m}\big\|L_W^{\frac{1}{2}}W(\widetilde{H}^T\widetilde{H})^{\frac{1}{2}}\big\|_{2,\infty} \leq B,~\sqrt{n}\big\|L_H^{\frac{1}{2}}H(\widetilde{W}^T\widetilde{W})^{\frac{1}{2}}\big\|_{2,\infty} \leq B.
	\end{aligned}
\end{equation}
The operator (\ref{projection operator}) finds a
factorized matrix $ F = [W^T, H^T]^T $  which is closest to $ \widetilde{F} $ while maintaining graph incoherent in a weighted sense. The following proposition demonstrates that this projection can be efficiently computed via a simple closed-form solution. The proofs of all the proposition and theorems presented later are provided in the supplementary material due to page limitations.
\begin{proposition}\label{proposition projection}
	The projection of $ \widetilde{F} $ in (\ref{projection operator}) has the following closed-form solution:
$
		\mathcal{P}_B(\widetilde{F}) = [
			(L_W^{-\frac{1}{2}}\mathcal{W})^T,
			(L_H^{-\frac{1}{2}}\mathcal{H})^T]^T,
$
	where each row of matrices $ \mathcal{W}\in \mathbb{R}^{m\times r} $ and $ \mathcal{H}\in \mathbb{R}^{n\times r} $ can be calculated by 
$
			\mathcal{W}_{i:} = \bigg(1 \wedge \frac{B}{\sqrt{m}\|\widetilde{\mathcal{W}}_{i:}\widetilde{\mathcal{H}}^T\|_2}\bigg)\widetilde{\mathcal{W}}_{i:}, ~~
			\mathcal{H}_{j:} = \bigg(1 \wedge \frac{B}{\sqrt{n}\|\widetilde{\mathcal{H}}_{j:}\widetilde{\mathcal{W}}^T\|_2}\bigg)\widetilde{\mathcal{H}}_{j:}
$
	with $ \widetilde{\mathcal{W}} := L_W^{\frac{1}{2}} \widetilde{W} $ and $ \widetilde{\mathcal{H}} := L_H^{\frac{1}{2}} \widetilde{H} $.
\end{proposition}

\subsection{Graph Spectral Initialization}\label{graph spectral initialization}
In the related researches of nonconvex optimization, it has been widely demonstrated that proper initialization plays a critical role in both the theoretical guarantees and practical performance of gradient-based methods. A common strategy for generating a reasonably good initial estimate is the spectral method. For a matrix $ X \in \mathbb{R}^{m \times n} $, its top-$ r $ SVD is given by $ U_r \Sigma_r V_r^T $, where matrices $ U_r \in \mathbb{R}^{m \times r} $ and $ V_r \in \mathbb{R}^{n \times r} $ consist of the top-$ r $ left and right singular vectors of $ X $, respectively, and $ \Sigma_r \in \mathbb{R}^{r \times r} $ is a diagonal matrix containing the corresponding top-$ r $ singular values. Then, for matrix completion problem, the spectral method offers an initialization $ (W_0, H_0) $ in the following form:
\begin{definition}[Standard spectral initialization]
	For the matrix completion problem, let $ U_0\Sigma_0V_0^T $ denote the top-$ r $ SVD of $ p^{-1}\mathcal{P}_\Omega(X_\star) $, then standard spectral method provides an initialization $ (W_0, H_0) $ such that
		$ W_0 := U_0\Sigma_0^\frac{1}{2},~~H_0 := V_0\Sigma_0^\frac{1}{2} $.
\end{definition}
\begin{figure*}[t]
	\centering 
	\includegraphics[width=0.9\linewidth]{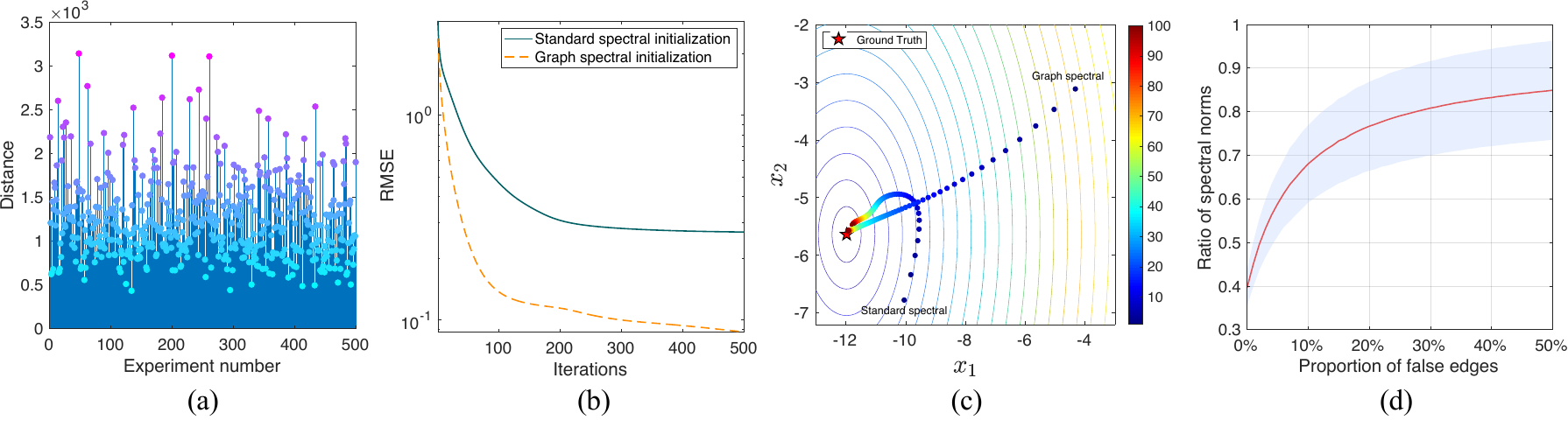}
	\vspace{0mm}
	\caption{(a): the magnitudes of $ \text{distance} :=  \text{distance}_\text{standard} - \text{distance}_\text{graph} $ in $ 500 $ synthetic experiments. (b)(c): the recovery RMSE and iteration trajectories of GSGD with standard spectral and graph spectral initialization, where the color of the trajectory points changes gradually with the number of iterations from $ 1 $ to $ 100 $. (d): Curve of the ratio of spectral norms $ \frac{\|\mathcal{A}X\mathcal{B} - X
	\|_\text{op}}{\|X\|_\text{op}} $ with respect to the proportion of false edges in the graphs.
	}\label{figure inipath}
	\vspace{0mm}
\end{figure*}

This simple strategy has proven highly effective in providing a ``warm start'' for many
nonconvex matrix factorization algorithms. Despite this, for graph regularized matrix recovery problems, the standard spectral initialization fails to incorporate graph information, often resulting in degraded outcomes. To address this limitation, we propose a graph spectral initialization method:
\begin{definition}[Graph spectral initialization]
	For graph regularized matrix completion problem, let $ U_0\Sigma_0V_0^T $ denote the top-$ r $ SVD of matrix $ p^{-1}\mathcal{A}\mathcal{P}_\Omega(X_\star)\mathcal{B} $, where $ \mathcal{A} $, $ \mathcal{B} $ are computed as in (\ref{LWLH form}), then graph spectral method offers an initialization $ (W_0, H_0) $ such that
		$ W_0 := U_0\Sigma_0^\frac{1}{2},~~H_0 := V_0\Sigma_0^\frac{1}{2} $.
\end{definition}

To empirically compare the two initialization methods, we implement GSGD with both standard and graph spectral initialization on synthetic matrices with similarity graphs, where the data generation technique will be detailed in the synthetic data experiments section.  We repeat the experiment $ 500 $ times, each time  calculating the Euclidean distances between ground truth and initial points for the two initialization methods, denoted as $ \text{distance}_\text{standard} $ and $ \text{distance}_\text{graph} $, respectively. To evaluate the relative magnitudes of these distances, we compute $ \text{distance} :=  \text{distance}_\text{standard} - \text{distance}_\text{graph} $ and display the resulting $ 500 $ values in Figure \ref{figure inipath} (a). It can be seen that all  $ 500 $ computed $ \text{distances} $ are greater than zero, indicating that graph spectral initialization consistently produces initial points closer to the ground truth compared to standard spectral initialization. Furthermore, as a case study, we present the recovery RMSE and iteration trajectories from one representative experiment in Figure \ref{figure inipath} (b)(c). It is evident that the trajectory starting from standard spectral initialization often follows a winding path, whereas the one from graph spectral initialization progresses directly toward the target, resulting in faster convergence and improved recovery performance.

Combining the update rules (\ref{GSGD update}), the projection operator $ \mathcal{P}_B(\cdot) $, and the graph spectral initialization, our algorithm is summarized in Algorithm \ref{algorithm MC}. Details regarding its implementation and computational complexity are provided in the appendix due to page limitations.
\begin{algorithm}[H]
	\caption{Graph regularized Scaled Gradient Descent algorithm (GSGD)}
	\label{algorithm MC}
	\begin{algorithmic}[1]
		%\tiny
		\State \textbf{Input:} $X_0$: observed matrix, $\Omega$: set of the indices of the observed entries, $\widetilde{L}_W, \widetilde{L}_H$: the Laplacian matrices of the similarity graphs, $ r $: rank of matrix, $ \beta $, $ \lambda $: model parameter.
		\State \textbf{Output:} $X$: estimated matrix.
		\State Compute $\mathcal{A} $, $ \mathcal{B} $, $  L_W $ and  $ L_H $ by (\ref{LWLH form}).
		\State  Let $ U_0\Sigma_0V_0^T $ be the top-$ r $ SVD of $ \frac{1}{p}\mathcal{A}X_0\mathcal{B} $, and set:
		$
			\begin{bmatrix}
				W_0 \\ 
				H_0
			\end{bmatrix}
			=
			\mathcal{P}_B\bigg(\begin{bmatrix}
				U_0\Sigma_0^\frac{1}{2} \\ 
				V_0\Sigma_0^\frac{1}{2}
			\end{bmatrix}\bigg).
		$
		\State $ t\leftarrow0 $.
		\While{not converged}
		\State  Update $ W_{t+1} $ and $ H_{t+1} $ by: 
		$
			\begin{bmatrix}
				W_{t+1} \\ 
				H_{t+1}
			\end{bmatrix}
			=
			\mathcal{P}_B\bigg(\begin{bmatrix}
				W_t - \frac{\eta}{p}L_W\mathcal{P}_\Omega(W_tH_t^T-X_0)H_t(H_t^TH_t)^{-1} \\ 
				H_t - \frac{\eta}{p}L_H\mathcal{P}_\Omega(W_tH_t^T-X_0)^TW_t(W_t^TW_t)^{-1}
			\end{bmatrix}\bigg).
		$
		\State  $t\leftarrow t+1$.
		\EndWhile
		\State  \textbf{Return} $X = W_tH_t^T$.
	\end{algorithmic}
\end{algorithm}

\section{Theoretical  Analysis}
\label{section theoretical analysis}
In this section we establish the theoretical guarantees in terms of statistical and iteration complexities of GSGD. In (\ref{error metric}) we define a new graph-aware error metric to measure the distance between the iterates and the ground truth. On this basis, we show the contraction of the iterates under the new distance metric, which lies at the core of our analysis. To this end,  we first introduce a definition of $ \psi  $-smoothness to measure the quality of similarity graphs.
\begin{definition}[Graph quality measure]
	Graphs $ G_1 $ and $ G_2 $ are  $ \psi $-smooth on matrix $ X $ if
	$ 
		\frac{\|\mathcal{A}X\mathcal{B} - X
			\|_\text{op}}{\|X\|_\text{op}} \leq \sqrt{\frac{\psi r}{m\wedge n}} $,
	where matrices $ \mathcal{A} $ and $ \mathcal{B} $ are computed by (\ref{LWLH form}).
\end{definition}

It can be verified that the higher the quality of the graph, the smaller the ratio of spectral norms $ \frac{\|\mathcal{A}X\mathcal{B} - X
	\|_\text{op}}{\|X\|_\text{op}} $, as illustrated in Figure \ref{figure inipath} (d) which shows the  experimental trend of the mean and standard deviation of the ratio with respect to the proportion of false edges in the graphs.  Thus $ \psi $ can be used as a measure of the quality: a small $ \psi $ means that the corresponding similarity graphs are quite smooth on matrix $ X $, and vise versa. The following theorem ensures the new projection satisfies both non-expansiveness and graph incoherence under the new error metric.
\begin{theorem}[Property of new projection operator]\label{lemma property}
	Suppose that $ X_\star $ is $ \mu $-graph incoherent with respect to $ \widetilde{L}_W $ and $ \widetilde{L}_H $, and $ \text{dist}(\widetilde{F}, F_\star) \leq \epsilon\sigma_r(X_\star) $ for some $ \epsilon < 1 $. Set the projection radius $ B \geq (1+\epsilon)\sqrt{\mu r (1+\beta)}\sigma_1(X_\star)  $, then $ \mathcal{P}_B(\widetilde{F}) $ satisfies the non-expansiveness
$ 
		\text{dist}(\mathcal{P}_B(\widetilde{F}) , F_\star) \leq \text{dist}(\widetilde{F}, F_\star) $,
	and the graph incoherence condition
		$ \sqrt{m}\|L_W^{\frac{1}{2}}WH^T\|_{2,\infty} \vee \sqrt{n}\|L_H^{\frac{1}{2}}HW^T\|_{2,\infty} \leq B $. 
\end{theorem}

The next theorem guarantees that the iterates of Algorithm \ref{algorithm MC} converge linearly and remain graph incoherent as long as the sample complexity is large enough.

\begin{theorem}[Linear convergence of the iterates]\label{lemma linear converge}
	Suppose that $ X_\star $ is $ \mu $-graph incoherent with respect to $ \widetilde{L}_W $ and $ \widetilde{L}_H $, $ p \geq C\big(\mu r \kappa^4 \vee \frac{\log(m \vee n)}{1+\beta}\big)\mu r /(m \wedge n) $ for some sufficiently large constant $ C $, and set the  projection radius $ B =  C_B\sqrt{\mu r (1+\beta)}\sigma_1(X_\star)  $ for some constant $ C_B \geq 1+0.02(1+\beta) $. Under an event $ \mathcal{E} $ which happens with overwhelming probability, if the parameter $ \beta $ and step size $ \eta $ obey $ 0 < \beta \leq 1 $ and $  0 < \eta \leq \frac{2}{2(1+\beta)+\sqrt{(1+\beta)}} $, and the $ t $-th iterate of Algorithm \ref{algorithm MC}  satisfies  $ \text{dist}(F_t,F_\star) \leq 0.02(1+\beta)\sigma_r(X_\star) $ and the graph incoherence condition 
		$ \sqrt{m}\|L_W^{\frac{1}{2}}W_tH_t^T\|_{2,\infty} \vee \sqrt{n}\|L_H^{\frac{1}{2}}H_tW_t^T\|_{2,\infty} \leq B $,
	then the $ (t+1) $-th iterate $ F_{t+1} $ satisfies  
		$ \text{dist}(F_{t+1},F_\star) \leq (1-\gamma\eta)\text{dist}(F_t,F_\star), ~~~\|W_{t+1}H_{t+1}^T - X_\star\|_F \leq 1.5\text{dist}(F_{t+1},F_\star) $
	and  the graph incoherence condition 
$ 
		\sqrt{m}\|L_W^{\frac{1}{2}}W_{t+1}H_{t+1}^T\|_{2,\infty} \vee \sqrt{n}\|L_H^{\frac{1}{2}}H_{t+1}W_{t+1}^T\|_{2,\infty} \leq B $,
	where $ \gamma $ is a constant between $ 0 $ and $ 1 $.
\end{theorem}

Theorem \ref{lemma linear converge} ensures that, as long as the initialization is close to the ground truth and satisfies the graph incoherence condition, the iterates of Algorithm \ref{algorithm MC} converge linearly and remain graph incoherent. The following theorem demonstrates that such an initialization can be achieved using the proposed graph spectral method.
\begin{theorem}[Graph spectral initialization]\label{lemma initialization}
	Suppose that $ X_\star $ is $ \mu $-graph incoherent with respect to $ \widetilde{L}_W $ and $ \widetilde{L}_H $, and $ G_1 $, $ G_2 $ are $ \psi $-smooth on matrix $ X_\star $. Then with overwhelming probability, the graph spectral initialization before projection $ \widetilde{F}_0 := [W_0^T,H_0^T]^T $ satisfies
	\begin{equation}\label{conclusion in lemma initialization}
		\text{dist}(\widetilde{F}_0,F_\star) \leq C\bigg( \frac{\mu r\log(m \vee n)}{p\sqrt{mn}} + \sqrt{\frac{\mu r\log(m \vee n)}{p(m \wedge n)}} + \sqrt{\frac{\psi r}{p(m \wedge n)}}   \bigg)5\sqrt{r(1+\beta)}\kappa\sigma_r(X_\star).
	\end{equation}
\end{theorem}

It is easy to verified from Theorem \ref{lemma initialization} that as long as $ p \geq C\big( \frac{\mu \log(m \vee n)}{1+\beta} \vee \frac{\psi}{1+\beta}   \big)r^2\kappa^2/(m\wedge n) $ for some sufficiently large constant $ C $, the graph spectral initialization before projection  $ \widetilde{F}_0 $ satisfies $ 	\text{dist}(\widetilde{F}_0,F_\star) \leq 0.02(1+\beta)\sigma_r(X_\star) $. Then Theorem \ref{lemma property} ensures that the graph spectral initialization $ F_0 = \mathcal{P}_B(\widetilde{F}_0) $ satisfies  $ \text{dist}(F_0,F_\star) \leq 0.02(1+\beta)\sigma_r(X_\star) $ and the graph incoherence condition 
	$ \sqrt{m}\|L_W^{\frac{1}{2}}W_0H_0^T\|_{2,\infty} \vee \sqrt{n}\|L_H^{\frac{1}{2}}H_0W_0^T\|_{2,\infty} \leq B $.
As a consequence, we can invoke Theorem \ref{lemma linear converge} to obtain the conclusion of Theorem \ref{theorem completion}, which is our main theoretical results.
\begin{theorem}\label{theorem completion}
	Suppose that $ X_\star $ is $ \mu $-graph incoherent with respect to $ \widetilde{L}_W $ and $ \widetilde{L}_H $,  $ G_1 $ and $ G_2 $ are $ \psi $-smooth on matrix $ X_\star $, and $ p \geq C\big(\mu^2 \kappa^2 \vee \frac{\mu\log(m \vee n)}{1+\beta} \vee \frac{\psi}{1+\beta}\big) r^2\kappa^2 /(m \wedge n) $ for some sufficiently large constant $ C $. Set the  projection radius $ B =  C_B\sqrt{\mu r (1+\beta)}\sigma_1(X_\star)  $ for some constant $ C_B \geq 1+0.02(1+\beta) $. If the parameter $ \beta $ and step size $ \eta $ obey $ 0 < \beta \leq 1 $ and $  0 < \eta \leq \frac{2}{2(1+\beta)+\sqrt{(1+\beta)}} $, 
	then with overwhelming probability, for all $ t \geq 0 $, the iterates of Algorithm \ref{algorithm MC} satisfy
	\begin{equation*}
		\text{dist}(F_{t},F_\star) \leq (1-\gamma\eta)^t0.02(1+\beta)\sigma_r(X_\star), ~~~\|W_{t}H_{t}^T - X_\star\|_F \leq (1-\gamma\eta)^t0.03(1+\beta)\sigma_r(X_\star),
	\end{equation*}
	where $ \gamma $ is a constant between $ 0 $ and $ 1 $.
\end{theorem}
\begin{remark}
	Theorem \ref{theorem completion} demonstrates that our graph regularized matrix completion algorithm GSGD contracts linearly as long as the probability of observations satisfies $ p \gtrsim \big(\mu^2 \kappa^2 \vee \frac{\mu\log(m \vee n)}{1+\beta} \vee \frac{\psi}{1+\beta}\big) r^2\kappa^2 /(m \wedge n) $. It takes at most $ T = O\big(\log(\frac{1}{\epsilon})\big) $ iterations to reach $ \epsilon $-accuracy, i.e., $ \|W_tH_t^T  - X_\star \|_F \leq \epsilon\sigma_r(X_\star) $. In comparison,  to reach $ \epsilon $-accuracy, ScaledGD for general matrix completion takes $ T = O(\log\big(\frac{1}{\epsilon})\big) $ iterations as long as $ p \gtrsim \big(\mu^2 \kappa^2 \vee \mu\log(m \vee n) \big) r^2\kappa^2 /(m \wedge n) $. Thus, when there is not accessible graphs, i.e., $ \beta=0 $, $ \psi=0 $, the sample complexity of GSGD  degrades into that of ScaledGD. And when we have access to the similarity graphs, as long as their quality is  good enough, i.e., $ \psi $ is quite small, GSGD achieves lower sampling complexity than ScaledGD for $ \beta > 0 $, which reflects the effect of graph information in matrix completion problems. Furthermore, it can be seen that the higher the quality of the graphs, the lower the sampling complexity of GSGD, which is also in line with expectations. Regarding the convergence speed, although both ScaledGD and GSGD have an iteration complexity of $ O(\log\big(\frac{1}{\epsilon})\big) $, empirical results indicate that GSGD converges significantly faster than ScaledGD, as will be demonstrated in the numerical experiments.
\end{remark}

\section{Synthetic Data Experiments}
\label{section synthetic data}
In this section we evaluate the performance of our algorithm on synthetic data. To this end, we first generate ground truth matrix $ X_\star\in \mathbb{R}^{m\times n} $ and similarity graphs in the following way. We randomly generate two graphs $ G_1 $ and $ G_2 $ with totally $ m $ and $ n $ vertexes using GSPbox (Graph Signal Processing toolbox)~\citep{perraudin2014gspbox}. Denote the Laplacian matrices of $ G_1 $ and $ G_2 $ as $ \widetilde{L}_W $, $ \widetilde{L}_H $, respectively, and $ \widetilde{L}_W = U_W\Sigma_WU_W^T $, $ \widetilde{L}_H = U_H\Sigma_HU_H^T $  are the singular value decomposition of $ \widetilde{L}_W $ and $ \widetilde{L}_H $. A rank-$ r $ matrix $ X_\star $ smooth on $ G_1 $ and $ G_2 $ is generated by 
	$ Z := UV^T, ~~X_\star := AZB^T $,
where matrices $ U\in \mathbb{R}^{m\times r} $ and $ V\in \mathbb{R}^{n\times r} $ are
independently sampled from Gaussian distribution, and matrices $ A\in \mathbb{R}^{m\times m} $ and $ B\in \mathbb{R}^{n\times n} $ are defined as
	$ A := U_Wg(\Sigma_W), B := U_Hg(\Sigma_H) $
with graph spectral filter $ g(\cdot) $. Here $ A $ and $ B $ transform the random matrix $ Z $ into a graph smooth matrix $ X_\star  $.

In the following subsections, we first validate GSGD from various perspectives in \ref{section similarity graphs} - \ref{section graph spectral initialization}, and then compare GSGD to state-of-the-art algorithms for graph regularized and 
graph-agnostic matrix completion in \ref{section scalability}. All the numerical experiments
are implemented on a desktop computer with Intel Core i9-9900k CPU, 64.0G RAM and MATLAB R2022a.

\subsection{The Exploitation of Graph Information}\label{section similarity graphs}
To compare the ability of graph Laplacian regularization and the proposed method to exploit graph information, we evaluate the recovery performance of RGD and GSGD on synthetic data. Meanwhile, we employ two graph-agnostic matrix completion methods, that is, GD  and ScaledGD as the baselines. We generate the ground truth low-rank matrix $ X_\star \in \mathbb{R}^{1000\times 1000} $ with similarity graphs $ G_1 $ and $ G_2 $ in the way described above.  Denote $ \Omega $ as the set of Bernoulli observed positions with probability $ p $, then  the observation matrix $ Y $ is generated by
	$ Y: = \mathcal{P}_\Omega(X_\star + E) $,
where $ E_{i,j} \sim \mathcal{N}(0, \sigma^2) $ are i.i.d. Gaussian noise. We evaluate the recovery performance of an algorithm by the root mean square error (RMSE) of its  retrieved matrix $ \widehat{X} $:
$
	\text{RMSE} = \sqrt{\frac{1}{|\overline{\Omega}|}\sum_{(i,j)\in \overline{\Omega}}\big(\widehat{X}_{ij} - (X_\star)_{ij}\big)^2},
$
where $ \overline{\Omega} $ denotes the complement of $ \Omega $, i.e., the set of missing positions.
We  consider two scenarios: noise-free observations with $ \sigma = 0 $, and noisy observations with $ \sigma = 0.1 $. For each scenario, we set sampling rate $ p = 10\% $, and run $ 100 $ tests. We illustrate the mean RMSE of various algorithms in Figure~\ref{figure synthetic} (a)(b), where we set the optimal step size for each algorithm, and fix regularization parameter $ \beta= 1 $ for RGD and $ \beta = 1 $, $ \lambda = 1 $ for GSGD.  It can be seen that: 
\begin{figure*}[t]
	\centering 
	\includegraphics[width=0.9\linewidth]{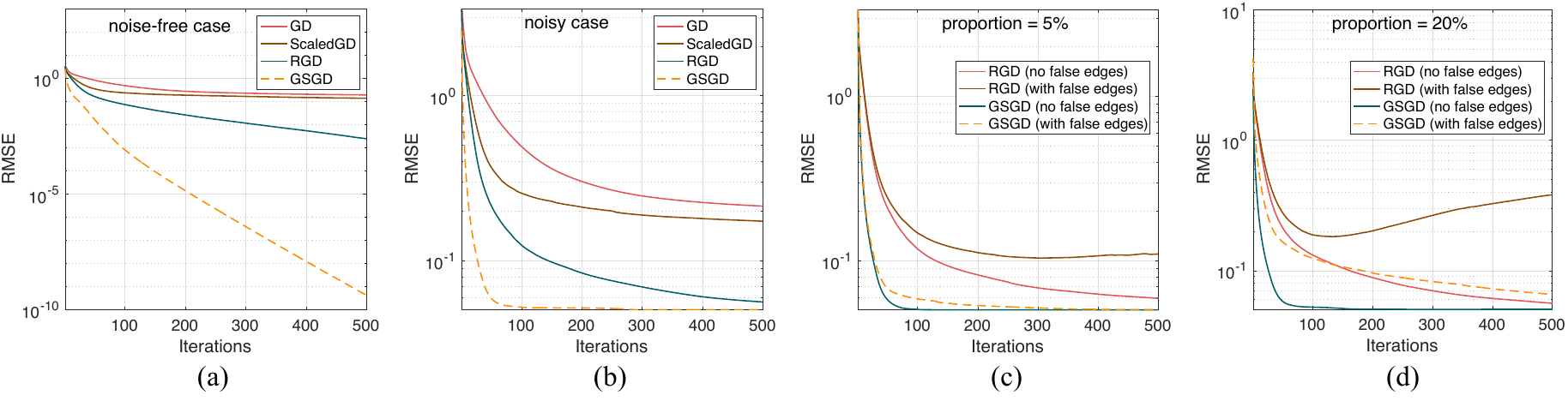}
	\vspace{0mm}
	\caption{(a)(b): The test RMSE  of various algorithms with respect to iteration count for cases with noise-free and noisy observations, where the Y-axis is set to a logarithmic scale for clarity. (c)(d): The synthetic experiments in the presence of false edges with proportion = 5\% and 20\%, respectively.
	}\label{figure synthetic}
	\vspace{0mm}
\end{figure*}
\begin{itemize}
	\item ScaledGD consistently achieves better performance than vanilla GD,  validating the utility of the preconditioners $ (H_t^TH_t)^{-1} $ and $ (W_t^TW_t)^{-1} $ in (\ref{ScaledGD update}). GSGD inherits the well-established preconditioners from ScaledGD, thereby building on a strong foundational model with proven effectiveness.
	\item Owing to the effective utilization of similarity graphs by graph Laplacian regularization, RGD consistently achieves better results in both recovery performance and convergence speed compared to ScaledGD. Nevertheless, as we analyzed earlier, graph Laplacian regularization struggles to fully exploit the potential of graph information,  while GSGD offers significant improvements in this regard. Actually, regardless of the presence of noise, compared to RGD, GSGD demonstrates significant advantages in terms of recovery RMSE and iteration count. Specifically, in the case of noise-free observations, GSGD can achieve exact matrix recovery while reaching the same level of recovery RMSE as RGD with only one-tenth of the iteration count. In the noisy observations case, although exact matrix recovery is no longer attainable, GSGD still requires only one-tenth of the iteration count to achieve a lower RMSE than RGD.
	 This consistent and significant improvement in both recovery accuracy and efficiency highlights the dual advantage of GSGD over graph Laplacian regularization in the capability and stability in exploiting graph information. 
\end{itemize}

\begin{figure*}[t]
	\centering 
	\includegraphics[width=0.9\linewidth]{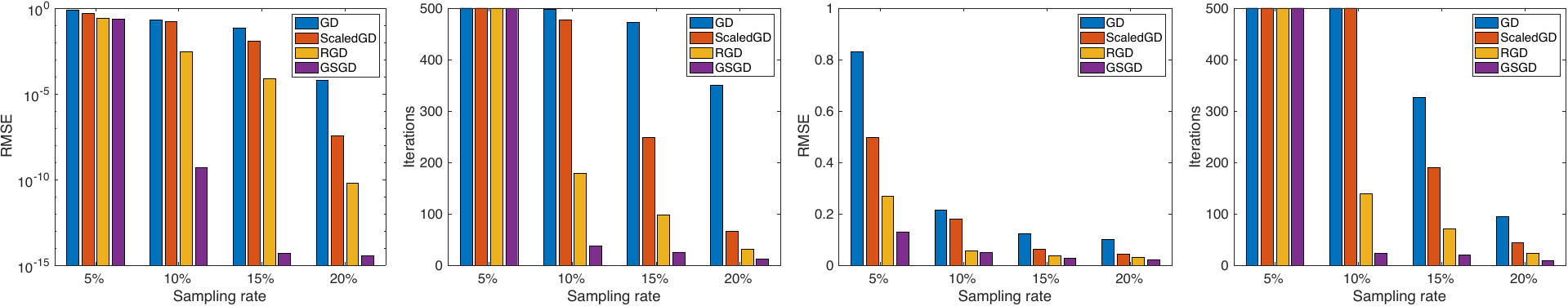}
	\vspace{0mm}
	\caption{The RMSE and the number of iterations required to achieve RMSE $ \leq 0.05 $ (noise-free case) and $ 0.1 $ (noisy case) under sampling rate $ p = 5\%, 10\%, 15\%, 20\% $. Left two: noise-free observations; Right two: noisy observations.
	}\label{figure synthetic various p}
	\vspace{0mm}
\end{figure*}
Furthermore, to evaluate the recovery performance of various algorithms under different sampling rate $ p $, we set $ p = 5\%, 10\%, 15\%, 20\% $ and record the RMSE and number of iterations required to achieve RMSE  $ \leq 0.05 $ for noise-free observations and RMSE  $ \leq 0.1 $ for noisy observations. The results are presented as a bar chart in Figure \ref{figure synthetic various p}. We can see that the RMSE and required iterations of these algorithms tend to decrease with higher sampling rates, which is expected, as more observed data makes it easier to recover the target matrix. Meanwhile, the RMSE and number of iterations for GSGD are significantly lower than those of the other three methods, highlighting the dual advantage of GSGD in both recovery accuracy and speed.

\subsection{Robustness Against False Edges}
\label{section false edges}
To assess the impact of false edges, we evaluate the performance of RGD and GSGD on synthetic data in the presence of false edges. We simulate false edges in the graph by randomly deleting and adding edges in a certain proportion, and compare the RMSE of RGD and GSGD on data with/without false edges shown in Figure~\ref{figure synthetic} (c)(d).  We observe that RGD's performance significantly deteriorates as the proportion of false edges increases, further highlighting the sensitivity of graph Laplacian regularization to false edges. In contrast, GSGD is much less affected, demonstrating its considerable robustness and stability against false edges.  This holds significant importance for the practical application of GSGD.

\subsection{The Role of Higher-order Smoothness}
\label{section higher-order smoothness}
\begin{figure*}[t]
	\centering 
	\includegraphics[width=0.9\linewidth]{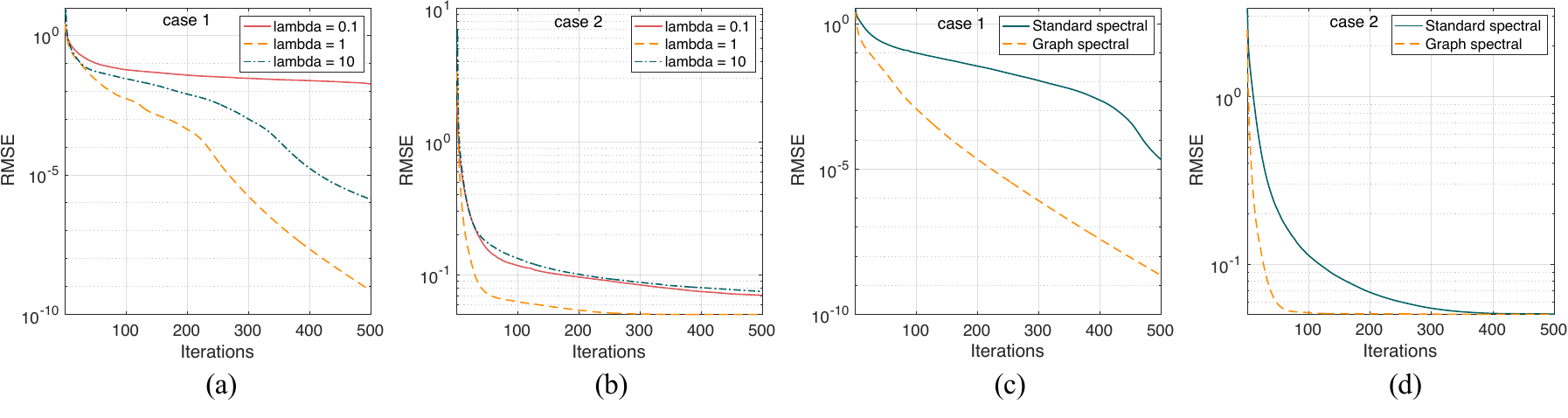}
	\vspace{0mm}
	\caption{(a)(b): Comparison of GSGD with different values of $ \lambda = 0.1, 1, 10 $ on two cases of synthetic data. (c)(d): Comparison of GSGD with standard and graph spectral initialization on two cases of synthetic data.  Case 1: synthetic data with noise-free observations; Case 2: synthetic data with noisy observations.
	}\label{figure higherordergraph ini}
	\vspace{0mm}
\end{figure*}
To evaluate the role of higher-order smoothness, we compare the recovery  performance of GSGD with different values of $ \lambda = 0.1, 1, 10 $ on synthetic data with noise-free (case $ 1 $) and noisy (case $ 2 $) observations, respectively, and report the results in Figure \ref{figure higherordergraph ini} (a)(b). Figure \ref{figure higherordergraph ini} (a)(b) shows that GSGD consistently achieves superior performance at $ \lambda = 1 $ compared to $ \lambda = 0.1 $ and $ \lambda = 10 $ in both cases, indicating that an appropriate level of higher-order smoothness indeed facilitates the improved recovery of target matrix. In all our other experiments, we fix $ \lambda = 1 $ for convenience.

\subsection{The Effectiveness of Graph Spectral Initialization}
\label{section graph spectral initialization}
To verify the advantage of the proposed graph spectral initialization approach over standard spectral initialization, we compare the recovery performance of GSGD with both initialization methods on synthetic data with noise-free (case $ 1 $) and noisy (case $ 2 $) observations, respectively, and show the results in Figure \ref{figure higherordergraph ini} (c)(d). It can be seen that compared to standard spectral initialization, graph spectral initialization significantly improves the convergence speed of the algorithm, which attributes to the effective incorporation of graph information. Specifically, GSGD with graph spectral initialization can achieve the same RMSE as standard spectral initialization with only about one-fourth of the iterations, resulting in significant time savings.

\subsection{Comparison with State-of-the-Art Algorithms}
\label{section scalability}
In this subsection, we compare GSGD to state-of-the-art algorithms for graph regularized and graph-agnostic matrix completion, which includes:
\begin{itemize}
	\item GRALS: Graph Regularized Alternating Least Squares \citep{rao2015collaborative} --- This algorithm is widely recognized as a state-of-the-art graph regularized matrix completion method.
	\item RGD: Riemannian Gradient Descent \citep{dong2021riemannian} --- A newly developed matrix completion algorithm solving a matrix factorization model with graph Laplacian regularization.
	\item ScaledGD: Scaled Gradient Descent \citep{tong2021accelerating} --- A recently proposed matrix completion method which significantly improves the convergence speed of gradient descent.
	\item AIS-Impute: Accelerated and Inexact Soft-Impute \citep{yao2018accelerated} --- This algorithm significantly accelerates the Soft-Impute, a state-of-the-art matrix completion method.
\end{itemize}

To comprehensively evaluate the recovery performance and scalability of these algorithms, we record their RMSE and runtime on synthetic data under different scenarios, including various sampling rates ($ p = 5\% \sim 20\% $), data sizes ($ m, n  = 5\times 10^3 \sim  10^5 $), and noise-free or noisy observations ($ \sigma = 0/0.1 $). We randomly select 20\% of those elements observed to serve as a validation set. The rank, step size and regularization parameters of these methods  are selected using the validation set. The results are shown in Table \ref{tab1}. Overall, we see that GSGD achieves significantly  superior recovery results on all data sets evaluated, while requiring considerably less time than other methods, and this advantage becomes even more significant in cases with large data sizes (e.g., at the scale of $ m = 10^4, n = 5\times 10^4 $ and $ m = 10^4, n = 10^5 $). Specifically, we can make the following observations:
\begin{table}[t]\footnotesize
	\caption{Comparison with State-of-the-Art Algorithms on Synthetic Data.}
	\center
	%	\vspace{-3mm}
	\setlength{\extrarowheight}{3pt}  % 为每一行增加3pt的高度
	\setlength{\tabcolsep}{4.0pt} % 通过减小这个值来压缩列间距
	\begin{tabular}{cccccccccccccc}
		%\hline
		\toprule
		\vspace{-2.5mm}
		& & & & \multicolumn{2}{c}{GSGD} & \multicolumn{2}{c}{GRALS}& \multicolumn{2}{c}{RGD}& \multicolumn{2}{c}{ScaledGD}& \multicolumn{2}{c}{AIS-Impute}\\ %\cline{5-14}
		&&&& \hspace{0.05cm}\hrulefill\hspace{-0.2cm} & \hspace{-0.2cm}\hrulefill\hspace{0.05cm} &\hspace{0.05cm}\hrulefill\hspace{-0.2cm} & \hspace{-0.2cm}\hrulefill\hspace{0.05cm} &\hspace{0.05cm}\hrulefill\hspace{-0.2cm} & \hspace{-0.2cm}\hrulefill\hspace{0.05cm} &\hspace{0.05cm}\hrulefill\hspace{-0.2cm} & \hspace{-0.2cm}\hrulefill\hspace{0.05cm} &\hspace{0.05cm}\hrulefill\hspace{-0.2cm} & \hspace{-0.2cm}\hrulefill\hspace{0.05cm}   \\
		p &$ \sigma $ & m & n &  RMSE& Time(s)& RMSE& Time(s)& RMSE& Time(s)& RMSE& Time(s)& RMSE& Time(s)		
		\\\hline
		\multirow{8}{*}{5\%} & \multirow{4}{*}{0} & $ 5\times10^3 $ & $ 5\times10^3 $ & 0.0009 & 19.1 & 0.0061& 14.1&0.0031 &30.1 & 0.0501 &45.2&0.0913&7.8\\
		&  &$ 10^4 $ & $ 10^4 $ &0.0009  & 100.6 &0.0036 &203.2 &0.0099 & 199.0&0.0111 & 209.0& 0.0531&59.3\\
		&  & $ 10^4 $ & $ 5\times10^4 $  & 0.0009& 428.8 &0.0704&582.6&0.1199&989.2 &0.2912 &1254.0 &0.2480 &456.1\\
		&  & $ 10^4 $ & $ 10^5 $  & 0.0009  & 617.1 & 0.0317&655.0 &0.0838 &1489.2 &0.1691 &1232.5 & 0.1273&1010.1
		\\\cline{2-14}
		& \multirow{4}{*}{0.1} & $ 5\times10^3 $ & $ 5\times10^3 $  & 0.0066 & 27.7 &0.0100 &30.6& 0.0088&28.3 & 0.0555& 47.9&0.0928 &7.9\\
		&  &$ 10^4 $ & $ 10^4 $& 0.0066 & 146.5 &0.0093 &143.4 &0.0290 & 280.9&0.0136 &183.2 & 0.0533&58.0\\
		&  & $ 10^4 $ & $ 5\times10^4 $  & 0.0067 & 336.9 &0.0991 &873.9 & 0.1611& 1013.3& 0.2974&625.9 & 0.2720&453.2\\
		&  & $ 10^4 $ & $ 10^5 $ &0.0053  & 593.0 &0.1374 & 677.4&0.1565 &1355.5 & 0.2763& 1701.5& 0.1926&999.4
		\\ \hline
		\multirow{8}{*}{10\%} & \multirow{4}{*}{0} & $ 5\times10^3 $ & $ 5\times10^3 $  & 0.0008 & 13.7 &0.0033 &30.2 &0.0022 &40.7 & 0.0101&56.9 &0.0146 &17.0\\
		&  &$ 10^4 $ & $ 10^4 $ &0.0008  &48.4  &0.0024 & 97.6&0.0025 &69.3 & 0.0082& 109.4&0.0071 &120.3\\
		&  & $ 10^4 $ & $ 5\times10^4 $  &0.0009  &212.6  &0.0630 &402.3 &0.0442 &813.5 &0.1089 &997.3 &0.1741 &961.7\\
		&  & $ 10^4 $ & $ 10^5 $  & 0.0007 & 269.7 & 0.0263& 490.1&0.0292 &836.6 & 0.0637& 1830.8& 0.0460&1609.6
		\\\cline{2-14}
		& \multirow{4}{*}{0.1} & $ 5\times10^3 $ & $ 5\times10^3 $  & 0.0037 & 14.4 &0.0073 & 30.3& 0.0077& 25.6&0.0113 &53.1 &0.0155 &17.3\\
		&  &$ 10^4 $ & $ 10^4 $ & 0.0037 & 54.0 &0.0062 & 204.5& 0.0053& 65.4& 0.0099&105.9 &0.0078 &122.7\\
		&  & $ 10^4 $ & $ 5\times10^4 $  &0.0031  & 379.4 &0.0708 &490.3 &0.0985 &794.2 &0.2087 &1550.6 &0.0927 &647.8\\
		&  & $ 10^4 $ & $ 10^5 $  & 0.0027 & 467.4 & 0.1195& 749.8&0.1276 &1110.6 & 0.1712& 1612.9& 0.0864&1455.0
		\\ \hline
		\multirow{8}{*}{20\%} & \multirow{4}{*}{0} & $ 5\times10^3 $ & $ 5\times10^3 $  & 0.0008 &9.5  & 0.0019&19.8 &0.0013 &16.2&0.0048 &39.4&0.0049&14.8\\
		&  &$ 10^4 $ & $ 10^4 $ & 0.0008 & 32.5 & 0.0019& 55.7&0.0017 & 47.2&0.0029 &60.3 & 0.0039&153.1\\
		&  & $ 10^4 $ & $ 5\times10^4 $  &0.0007 & 106.2 &0.0394&416.6&0.0272& 469.1&0.0547 & 413.7& 0.0846&1745.0\\
		&  & $ 10^4 $ & $ 10^5 $  & 0.0007 &179.5  & 0.0211& 449.9&0.0222 &499.7 & 0.0094& 1194.6& 0.0330&3030.5
		\\\cline{2-14}
		& \multirow{4}{*}{0.1} & $ 5\times10^3 $ & $ 5\times10^3 $  &0.0026  & 10.9 &0.0055 & 15.4& 0.0045&14.8 &0.0063 &39.3 &0.0092 &17.6\\
		&  &$ 10^4 $ & $ 10^4 $ & 0.0026 & 34.9 &0.0048 &95.1 &0.0042 & 58.0&0.0053 &52.6 & 0.0047&145.0\\
		&  & $ 10^4 $ & $ 5\times10^4 $  & 0.0020 & 174.5 &0.0401 &392.8 & 0.0665&481.3 &0.0602 &541.4 &0.0489 &1509.2\\
		&  & $ 10^4 $ & $ 10^5 $  & 0.0020 & 314.9 &0.0336 & 618.9&0.0352 & 878.0& 0.0451&1433.3 & 0.0223&2529.9
		\\ \hline
		\toprule
	\end{tabular}
	\label{tab1}
	%	\vspace{-4mm}
\end{table} 
\begin{itemize}
	\item As the observed data increases (with 
	$ p $ rising from $ 5\%  $ to $ 20\% $), the RMSE of all methods gradually decreases, which is consistent with expectations. As for runtime, the required time for GSGD, GRALS, RGD and ScaledGD generally tends to decreases when $ p $ is larger, as more observed data reduces the number of iterations needed. One exception is AIS-Impute, whose runtime increases significantly with larger $ p $, which is mainly because more observed data considerably increases the computation time for its approximate singular value thresholding scheme. 
	\item In most cases, graph regularized methods outperform graph-agnostic ones, with the advantage becoming more pronounced when $ p $ is small. This is because the severely limited amount of observed data significantly increases the challenge of matrix recovery, and at this point, the additional structural information provided by graph regularization plays a crucial role.
	\item GSGD demonstrates significant superiority over other methods in both recovery accuracy and runtime. In terms of recovery accuracy, GSGD achieves significantly lower RMSE compared to other methods for small data sizes (at the scale of $ m = 5\times10^3, n = 5\times 10^3 $ and $ m = 10^4, n = 10^4 $). For large data sets ($ m = 10^4, n = 5\times 10^4 $ and $ m = 10^4, n = 10^5 $), we observe that the performance of other methods deteriorates significantly—with RMSE increasing by an order of magnitude compared to small datasets—while GSGD maintains stable performance. Remarkably, the RMSE of GSGD remains similar to its performance on small datasets and is an order of magnitude smaller than that of other methods. This represents a substantial improvement of GSGD in recovery performance, highlighting its capability and stability in exploiting graph information. In terms of runtime, we observe that  GSGD consistently outperforms other methods in most cases, often requiring only a fraction of the time. It is worth noting that all four algorithms we compared are designed with scalability in mind and have demonstrated strong computational efficiency. Building on this, the speed advantage of GSGD  clearly underscores its superior scalability.
\end{itemize}

\section{Real-World Experiments}
\label{section realworld data}
We  shall report on the performance of GSGD on two real-world data sets: MovieLens1M\citep{harper2015movielens} and Epinions\citep{hamedani2021trustrec}. 
MovieLens1M is a well-known movie rating data set containing $ 1, 000, 000  $ movie ratings from
$ 6040 $ users on $ 3883 $ movies along with the user/movie
features.  Here we employ the user and movie features to construct 10-nearest neighbor similarity graphs using
the Euclidean distance metric, respectively.  We randomly withhold 30\% of the ratings as a test set, using the remaining 70\% to impute the complete
matrix, where cross validation is performed on the appropriate hyperparameters.
Epinions contains users’ ratings on items and explicit trust/distrust relationships between users collected from the general consumer review site Epinions.com, giving rise to a large-scale data set widely used to evaluate recommender systems in the literature. For this experiment, more than $ 13.3 $ million ratings scatter across the rating matrix of $ 32577 $ users and $ 674932 $ items, while  a similarity graph of users can be constructed based on the trust networks. We randomly mask $ 10\% $ of the ground truth values and use the remaining $ 90\% $ as observations to evaluate the recovery performance. The results are illustrated in Table \ref{tab2}.
\begin{table}[t]\footnotesize
	\caption{Comparison with State-of-the-Art Algorithms on Real-World Data Sets.}
	\center
	%	\vspace{-3mm}
	\setlength{\extrarowheight}{3pt}  % 为每一行增加3pt的高度
	\setlength{\tabcolsep}{3.5pt} % 通过减小这个值来压缩列间距
	\begin{tabular}{ccccccccccccc}
		%\hline
		\toprule
		\vspace{-2.5mm}
		& &  & \multicolumn{2}{c}{GSGD} & \multicolumn{2}{c}{GRALS}& \multicolumn{2}{c}{RGD}& \multicolumn{2}{c}{ScaledGD}& \multicolumn{2}{c}{AIS-Impute}\\ 

		&&& \hspace{0.05cm}\hrulefill\hspace{-0.2cm} & \hspace{-0.2cm}\hrulefill\hspace{0.05cm} &\hspace{0.05cm}\hrulefill\hspace{-0.2cm} & \hspace{-0.2cm}\hrulefill\hspace{0.05cm} &\hspace{0.05cm}\hrulefill\hspace{-0.2cm} & \hspace{-0.2cm}\hrulefill\hspace{0.05cm} &\hspace{0.05cm}\hrulefill\hspace{-0.2cm} & \hspace{-0.2cm}\hrulefill\hspace{0.05cm} &\hspace{0.05cm}\hrulefill\hspace{-0.2cm} & \hspace{-0.2cm}\hrulefill\hspace{0.05cm}   \\
		Dataset  & m & n &  RMSE& Time(s)& RMSE& Time(s)& RMSE& Time(s)& RMSE& Time(s)& RMSE& Time(s)		
		\\\hline
		MovieLens1M  & $ 6040 $ & $ 3883 $ & 0.868 & 13.8  &0.881 & 45.3&0.886 & 54.1&0.901 &60.9& 0.905 & 53.4
		\\
		Epinions  & $ 32577 $ & $ 674932 $  & 0.694 & 382.5 &0.747 & 765.3&0.752 &1586.3 &0.823 & 1671.4 & 0.801&1839.4
		\\ \hline
		\toprule
	\end{tabular}
	\label{tab2}
	%	\vspace{-4mm}
\end{table} 

We can see that GSGD consistently outperforms the other algorithms in recovery accuracy, while significantly reducing the required runtime. Basically, the three graph regularized methods demonstrate superiority over the two graph-agnostic ones in both recovery accuracy and runtime, which is primarily attributed to the exploitation of additional graph information. Among these methods, GSGD shows greater advantages in both effectiveness and efficiency.  Its accuracy advantage is particularly pronounced, reflecting GSGD's superior capability and  stability in extracting graph information. Besides, GSGD requires significantly less time than other methods. Specifically, on the large-scale data set Epinions, the runtime of GSGD is only half that of GRALS and a fraction of the other methods. Considering that these competing methods have been proven to be among the most efficient algorithms, this time advantage further ensures the superior scalability of GSGD.

\section{Conclusions}
\label{section conclusion}
We provided a new scalable and provable nonconvex optimization algorithm called GSGD for matrix completion problem with graph information.  Breaking away from the conventional graph Laplacian regularization framework, GSGD derives a preconditioned projected gradient descent algorithm incorporating higher-order graph information to enhance the recovery performance, which also demonstrates superior robustness and stability against false edges in the graph. Theoretically, we prove that  GSGD linearly converges to the global optimum at a rate independent of the condition number of the low-rank matrix  with near-optimal sample complexity, and high-quality graph information can effectively reduce the sample complexity. This establishes the first theoretical guarantee in terms of statistical and iteration complexities in the perspective of nonconvex optimization. Experimental results on synthetic and real-world data sets highlight the superior recovery accuracy and scalability of GSGD over several state-of-the-art methods for large-scale matrix completion tasks.  Extending GSGD to other graph regularized matrix recovery problems, such as matrix sensing and robust PCA, presents 
an intriguing research direction.

%\THEEndNotes
%\begingroup \parindent 0pt \parskip 0.0ex \def\enotesize{\normalsize} \theendnotes \endgroup

% Appendix here
% Options are (1) APPENDIX (with or without general title) or
%             (2) APPENDICES (if it has more than one unrelated sections)
% Outcomment the appropriate case if necessary
%
% \begin{APPENDIX}{<Title of the Appendix>}
% \end{APPENDIX}
%
%   or
%
% \begin{APPENDICES}
% \section{<Title of Section A>}
% \section{<Title of Section B>}
% etc
% \end{APPENDICES}

% Acknowledgments here
%\ACKNOWLEDGMENT{We would like to express our sincere gratitude to [acknowledge individuals, organizations, or institutions] for their invaluable contributions to this research. We are also grateful to [mention any additional acknowledgements, such as technical assistance, data providers, or colleagues] for their support and assistance throughout the course of this work.}

% References here (outcomment the appropriate case)

% CASE 1: BiBTeX used to constantly update the references
%   (while the paper is being written).
%\bibliographystyle{informs2014} % outcomment this and next line in Case 1
%\bibliography{<your bib file(s)>} % if more than one, comma separated

%\bibliographystyle{informs2014} % outcomment this and next line in Case 1
%\bibliography{sample} % if more than one, comma separated

% CASE 2: BiBTeX used to generate mypaper.bbl (to be further fine tuned)
%\input{mypaper.bbl} % outcomment this line in Case 2

%If you don't use BiBTex, you can manually itemize references as shown below.

%\bibliographystyle{nonumber}

\bibliographystyle{informs2014}
\bibliography{ref1}

\newpage
\RUNAUTHOR{Wang, Y., et al.}

% Title or shortened title suitable for running heads. Sample:
% \RUNTITLE{Predictive Maintenance in Manufacturing}
% Enter the (shortened) title:
\RUNTITLE{Supplemental Material for ``Matrix Completion with Graph Information: Nonconvex Optimization Approach''}
\setcounter{equation}{0}
\setcounter{theorem}{0}  % 重新开始定理编号
\setcounter{lemma}{0}     % 重新开始引理编号
\setcounter{definition}{0}
\begin{APPENDICES}
	{\setlength{\baselineskip}{30pt} 
	\begin{flushleft}
		\textbf{\LARGE Supplemental Material for ``Matrix Completion with Graph Information: A Provable Nonconvex Optimization Approach''}
	\end{flushleft}}

 \noindent	\rule{\textwidth}{1.0pt}

  This supplemental material contains details regarding the implementation and
 computational complexity of the proposed GSGD algorithm, and proofs of the Proposition 1 and Theorem 1-3 in the main manuscript, where the proofs of some intermediate technical lemmas are presented at the end.
 
  \noindent	\rule{\textwidth}{0.5pt}
 
	\section{Implementation and Computational Complexity of GSGD}\label{computational complexity}
	To accelerate the computation of $ L_W $ and $ L_H $, we use Incomplete Cholesky Decomposition to achieve fast inversion of sparse symmetric positive definite matrices while preserving sparsity.
	In practice, we find that the iterates of our algorithm remain graph incoherent, so that one may drop the projection step $ \mathcal{P}_B(\cdot) $. We perform the update rules without projections in our experiments. It can be verified that the computational complexity of the update rules is $ O\big (\lvert \Omega \rvert r + nnz(L_W)r + nnz(L_H)r + (m+n)r^2 + r^3 \big ) $, where $ nnz(\cdot) $ is the
	number of non zeros. Considering that $ r $ is much smaller than $ m $ and $ n $, and $ L_W $, $ L_H $ are usually quite sparse, the per-iteration cost of GSGD is very cheap, on the same order as gradient descent.
	
	\section{Proof of Proposition 1}
	\begin{proof}{Proof of Proposition 1}
		The optimization of $ W $ and $ H $  can be decoupled and done separately, thus in the following we focus on the optimization of  $ W $, and $ H $ can be obtained in similar way. $ W $ is solved by the following form:
		\begin{equation}\label{solve W}
			\begin{aligned}
				W= \argmin_{W\in \mathbb{R}^{m\times r}} \big\|L_W^{\frac{1}{2}}(W - \widetilde{W})(\widetilde{H}^T\widetilde{H})^{\frac{1}{2}}\big\|_F^2,
				~~~\text{s.t.} ~ \sqrt{m}\big\|L_W^{\frac{1}{2}}W(\widetilde{H}^T\widetilde{H})^{\frac{1}{2}}\big\|_{2,\infty} \leq B.
			\end{aligned}
		\end{equation}
		Denote $ G := L_W^{\frac{1}{2}}W(\widetilde{H}^T\widetilde{H})^{\frac{1}{2}} $ and $ \widetilde{G} := L_W^{\frac{1}{2}}\widetilde{W}(\widetilde{H}^T\widetilde{H})^{\frac{1}{2}} $, then (\ref{solve W}) can be equivalently rewritten as:
		\begin{equation}
			\begin{aligned}
				G= \argmin_{G\in \mathbb{R}^{m\times r}} \big\|G - \widetilde{G}\big\|_F^2, ~~~\text{s.t.} ~ \sqrt{m}\big\|G\big\|_{2,\infty} \leq B,
			\end{aligned}
		\end{equation}
		which can be solved  by the following closed-form solution \citep{chen2015fast}:
		\begin{equation}\label{G solve}
			G_{i:} = \bigg(1 \wedge \frac{B}{\sqrt{m}\|\widetilde{G}_{i:}\|_2}\bigg)\widetilde{G}_{i:}, ~ i=1,2,\cdots,m.
		\end{equation}
		Let $ \mathcal{W} := L_W^{\frac{1}{2}}W $, $ \widetilde{\mathcal{W}} := L_W^{\frac{1}{2}}\widetilde{W} $, then we have 
		$
		G_{i:} = (\mathcal{W}(\widetilde{H}^T\widetilde{H})^{\frac{1}{2}})_{i:} = \mathcal{W}_{i:} (\widetilde{H}^T\widetilde{H})^{\frac{1}{2}},
		\widetilde{G}_{i:} = (\widetilde{\mathcal{W}}(\widetilde{H}^T\widetilde{H})^{\frac{1}{2}})_{i:} = \widetilde{\mathcal{W}}_{i:} (\widetilde{H}^T\widetilde{H})^{\frac{1}{2}},
		$
		then (\ref{G solve}) implies that
		$
		\mathcal{W}_{i:} = \bigg(1 \wedge \frac{B}{\sqrt{m}\|\widetilde{\mathcal{W}}_{i:}\widetilde{\mathcal{H}}^T\|_2}\bigg)\widetilde{\mathcal{W}}_{i:}, ~ i=1,2,\cdots,m,
		$
		where we employ the equality $ \|\widetilde{\mathcal{W}}_{i:}\widetilde{\mathcal{H}}^T\|_2 = \|\widetilde{\mathcal{W}}_{i:} (\widetilde{H}^T\widetilde{H})^{\frac{1}{2}}\|_2. $
		After obtaining $ \mathcal{W} $, $ W $ can be directly calculated by $ W = L_W^{-\frac{1}{2}}\mathcal{W} $, which leads to the solution in the proposition. 
		\Halmos
	\end{proof}	
	
	\section{Optimal Alignment Matrix}
	For the convenience of subsequent proofs, we  first introduce the definition of optimal alignment matrix $ Q $ as follows.
	\begin{definition}[Optimal alignment matrix]
		For any factor matrix $ F := 
		[W^T,H^T]^T
		\in \mathbb{R}^{(m+n)\times r} $ and graph-aware error metric 
		$ \text{dist}(F,F_\star) = \sqrt{\inf_{Q\in \text{GL}(r)} \|L_W^{1/2}(WQ-W_\star)\Sigma_\star^{1/2}\|_F^2 + \|L_H^{1/2}(HQ^{-T}-H_\star)\Sigma_\star^{1/2}\|_F^2} $,
		the optimal alignment matrix $ Q $ between $ F $ and $ F_\star $ is defined as
		$ Q := \arg\min_{Q\in \text{GL}(r)} \|L_W^{1/2}(WQ-W_\star)\Sigma_\star^{1/2}\|_F^2 + \|L_H^{1/2}(HQ^{-T}-H_\star)\Sigma_\star^{1/2}\|_F^2 $,
		whenever the minimum is achieved.
	\end{definition}
	
	It is worth noting that $ Q $ is well-defined meaning that with proper initialization the optimal alignment matrix $ Q_t $ is guaranteed to exist for the iterates $ F_t $, which is ensured by the following lemma.
	\begin{lemma}[Existence of optimal alignment matrix]\label{Q exist}
		A sufficient condition for the existence of the optimal alignment matrix $ Q $ between $ F $ and $ F_\star $ is that $ \text{dist}(F, F_\star) < \sigma_r(X_\star) $.
	\end{lemma}

	\section{Proof of Theorem 1}
	\begin{proof}{Proof of Theorem 1}
		First, the condition $ \text{dist}(\widetilde{F}, F_\star) < \sigma_r(X_\star) $ and Lemma \ref{Q exist} ensures the existence of the optimal alignment matrix between $ \widetilde{F} $ and  $ F_\star $, which we denote as $ \widetilde{Q} $. Denote $ \mathcal{P}_B(\widetilde{F}) = [W^T, H^T]^T $,  $ \mathcal{W} = L_W^{\frac{1}{2}}W $, $ \widetilde{\mathcal{W}} = L_W^{\frac{1}{2}}\widetilde{W} $, $ \mathcal{H} := L_H^{\frac{1}{2}}H $, and $ \widetilde{\mathcal{H}} := L_H^{\frac{1}{2}}\widetilde{H} $, then it can be easily verified that
		\begin{equation}\label{dist PB F}
			\begin{aligned}
				\text{dist}^2(\mathcal{P}_B(\widetilde{F}), F_\star) & \leq \big\| \mathcal{W} \widetilde{Q}\Sigma_\star^{\frac{1}{2}} - L_W^{\frac{1}{2}} W_\star\Sigma_\star^{\frac{1}{2}} \big\|_F^2 + \big\| \mathcal{H} \widetilde{Q}^{-T}\Sigma_\star^{\frac{1}{2}} - L_H^{\frac{1}{2}} H_\star\Sigma_\star^{\frac{1}{2}} \big\|_F^2\\
				& = \sum_{i=1}^{m}\big\| \mathcal{W}_{i:} \widetilde{Q}\Sigma_\star^{\frac{1}{2}} - \big(L_W^{\frac{1}{2}} W_\star\Sigma_\star^{\frac{1}{2}}\big)_{i:} \big\|_F^2 + \sum_{j=1}^{n}\big\| \mathcal{H}_{j:} \widetilde{Q}^{-T}\Sigma_\star^{\frac{1}{2}} - \big(L_H^{\frac{1}{2}} H_\star\Sigma_\star^{\frac{1}{2}}\big)_{j:} \big\|_F^2,
			\end{aligned}
		\end{equation}
		\begin{equation}\label{dist F}
			\begin{aligned}
				\text{dist}^2(\widetilde{F}, F_\star) = \sum_{i=1}^{m}\big\| \widetilde{\mathcal{W}}_{i:} \widetilde{Q}\Sigma_\star^{\frac{1}{2}} - \big(L_W^{\frac{1}{2}} W_\star\Sigma_\star^{\frac{1}{2}}\big)_{i:} \big\|_F^2 + \sum_{j=1}^{n}\big\| \widetilde{\mathcal{H}}_{j:} \widetilde{Q}^{-T}\Sigma_\star^{\frac{1}{2}} - \big(L_H^{\frac{1}{2}} H_\star\Sigma_\star^{\frac{1}{2}}\big)_{j:} \big\|_F^2.
			\end{aligned}
		\end{equation}
		Condition $ \text{dist}(\widetilde{F}, F_\star) \leq \epsilon\sigma_r(X_\star) $ implies
		$
		\|L_W^{1/2}(\widetilde{W}\widetilde{Q} - W_\star)\Sigma_\star^{-1/2}\|_\text{op} \vee \|L_H^{1/2}(\widetilde{H}\widetilde{Q}^{-T} - H_\star)\Sigma_\star^{-1/2}\|_\text{op} \leq \epsilon,
		$
		then we have
		\begin{equation}\label{B1}
			\begin{aligned}
				&\big\|\widetilde{\mathcal{W}}_{i:}\widetilde{H}^T\big\|_2
				\leq \big\|\widetilde{\mathcal{W}}_{i:}\widetilde{Q}\Sigma_\star^{\frac{1}{2}}\big\|_2\big\|\widetilde{H}\widetilde{Q}^{-T}\Sigma_\star^{-\frac{1}{2}}\big\|_\text{op}
				\leq \big\|\widetilde{\mathcal{W}}_{i:}\widetilde{Q}\Sigma_\star^{\frac{1}{2}}\big\|_2\big\|H_\star\Sigma_\star^{-\frac{1}{2}} +( \widetilde{H}\widetilde{Q}^{-T} - H_\star)\Sigma_\star^{-\frac{1}{2}}\big\|_\text{op}\\
				& \leq \big\|\widetilde{\mathcal{W}}_{i:}\widetilde{Q}\Sigma_\star^{\frac{1}{2}}\big\|_2\big\|V_\star +( \widetilde{H}\widetilde{Q}^{-T} - H_\star)\Sigma_\star^{-\frac{1}{2}}\big\|_\text{op}
				\leq \big\|\widetilde{\mathcal{W}}_{i:}\widetilde{Q}\Sigma_\star^{\frac{1}{2}}\big\|_2\Big(\big\|V_\star\big\|_\text{op} +\big\|( \widetilde{H}\widetilde{Q}^{-T} - H_\star)\Sigma_\star^{-\frac{1}{2}}\big\|_\text{op}\Big)\\
				& \leq \big\|\widetilde{\mathcal{W}}_{i:}\widetilde{Q}\Sigma_\star^{\frac{1}{2}}\big\|_2\Big(1 +\big\|L_H^{\frac{1}{2}}( \widetilde{H}\widetilde{Q}^{-T} - H_\star)\Sigma_\star^{-\frac{1}{2}}\big\|_\text{op}\Big)
				\leq (1+\epsilon)\big\|\widetilde{\mathcal{W}}_{i:}\widetilde{Q}\Sigma_\star^{\frac{1}{2}}\big\|_2.
			\end{aligned}
		\end{equation}
		Meanwhile, the condition $ B \geq (1+\epsilon)\sqrt{\mu r(1+\beta)}\sigma_1(X_\star) \geq (1+\epsilon)\sqrt{\mu r}\sigma_1(X_\star)\sqrt{\|L_W\|_\text{op} \vee \|L_H\|_\text{op}}   $ and graph incoherence of $ X_\star $ implies that
		\begin{equation}\label{B2}
			\begin{aligned}
				\sqrt{m}\big\| \big(L_W^{\frac{1}{2}}W_\star\Sigma_\star^{\frac{1}{2}}\big)_{i:} \big\|_2 &\leq \sqrt{m}\big\| \big(L_W^{\frac{1}{2}}U_\star\big)_{i:} \big\|_2\big\|\Sigma_\star \big\|_\text{op} \leq \sqrt{m}\big\| L_W^{\frac{1}{2}}U_\star \big\|_{2,\infty}\big\|\Sigma_\star \big\|_\text{op} \leq \sqrt{\mu r}\sigma_1(X_\star)\\
				& \leq \frac{B}{(1+\epsilon)\sqrt{\|L_W\|_\text{op} \vee \|L_H\|_\text{op}}} \leq \frac{B}{(1+\epsilon)\sqrt{\|L_H\|_\text{op}}}.
			\end{aligned}
		\end{equation}
		Combining inequalities (\ref{B1}) and (\ref{B2}) gives rise to the following inequality:
		$
		\frac{B}{\sqrt{m\|L_H\|_\text{op}}\big\|\widetilde{\mathcal{W}}_{i:}\widetilde{H}^T\big\|_2} \geq \frac{\big\| \big(L_W^{\frac{1}{2}}W_\star\Sigma_\star^{\frac{1}{2}}\big)_{i:} \big\|_2 }{\big\|\widetilde{\mathcal{W}}_{i:}\widetilde{Q}\Sigma_\star^{\frac{1}{2}}\big\|_2}.
		$
		Then we record the following useful claim.
		\begin{claim}[\citep{tong2021accelerating}, Claim 5]\label{claim1}
			For $ \boldsymbol{u}, \boldsymbol{u}_\star \in \mathbb{R}^n $  and $ \lambda \geq \frac{\|\boldsymbol{u}_\star\|_2}{\|\boldsymbol{u}\|_2} $, it holds that
			$
			\|(1\wedge \lambda)\boldsymbol{u}-\boldsymbol{u}_\star\|_2 \leq \|\boldsymbol{u}-\boldsymbol{u}_\star\|_2.
			$
		\end{claim}
		Take the calculation rule in Proposition 1 and Claim \ref{claim1} with $ \boldsymbol{u} := \widetilde{\mathcal{W}}_{i:}\widetilde{Q}\Sigma_\star^{\frac{1}{2}} $, $ \boldsymbol{u}_\star :=  \big(L_W^{\frac{1}{2}}W_\star\Sigma_\star^{\frac{1}{2}}\big)_{i:} $, and $ \lambda := 	\frac{B}{\sqrt{m\|L_H\|_\text{op}}\big\|\widetilde{\mathcal{W}}_{i:}\widetilde{H}^T\big\|_2} $ collectively to reach 
		\begin{equation}\label{W inequality}
			\begin{aligned}
				\Big\| \mathcal{W}_{i:}\widetilde{Q}\Sigma_\star^{\frac{1}{2}} - \big(L_W^{\frac{1}{2}}W_\star\Sigma_\star^{\frac{1}{2}}\big)_{i:} \Big\|_2^2  \leq \Big\| \widetilde{\mathcal{W}}_{i:}\widetilde{Q}\Sigma_\star^{\frac{1}{2}} - \big(L_W^{\frac{1}{2}}W_\star\Sigma_\star^{\frac{1}{2}}\big)_{i:} \Big\|_2^2.
			\end{aligned}
		\end{equation}
		A similar inequality for $ \mathcal{H} $ and $ \widetilde{\mathcal{H}} $ can also be reached as follows:
		\begin{equation}\label{H inequality}
			\Big\| \mathcal{H}_{j:}\widetilde{Q}^{-T}\Sigma_\star^{\frac{1}{2}} - \big(L_H^{\frac{1}{2}}H_\star\Sigma_\star^{\frac{1}{2}}\big)_{j:} \Big\|_2^2 \leq \Big\| \widetilde{\mathcal{H}}_{j:}\widetilde{Q}^{-T}\Sigma_\star^{\frac{1}{2}} - \big(L_H^{\frac{1}{2}}H_\star\Sigma_\star^{\frac{1}{2}}\big)_{j:} \Big\|_2^2.
		\end{equation}
		Combining (\ref{W inequality}), (\ref{H inequality}) with (\ref{dist PB F}), (\ref{dist F}) leads to the conclusion
		$
		\text{dist}(\mathcal{P}_B(\widetilde{F}) , F_\star) \leq \text{dist}(\widetilde{F}, F_\star),
		$
		which proves the non-expansiveness of the new projection operator.
		
		To prove the graph incoherence condition, we record the definition of $ \|\cdot\|_{2,\infty} $ as follows:
		$
		\|L_W^{\frac{1}{2}}WH^T\|_{2,\infty} = \max_{i} \|\mathcal{W}_{i:}H^T\|_2,~\|L_H^{\frac{1}{2}}HW^T\|_{2,\infty} = \max_{j} \|\mathcal{H}_{j:}W^T\|_2,
		$
		thus we focus on the bound of $ \|\mathcal{W}_{i:}H^T\|_2 $ and $ \|\mathcal{H}_{j:}W^T\|_2 $.
		\begin{equation}\label{W incoherence }
			\begin{aligned}
				&\|\mathcal{W}_{i:}H^T\|_2^2 = \|\mathcal{W}_{i:}\mathcal{H}^TL_H^{\frac{1}{2}}\|_2^2  \leq \|\mathcal{W}_{i:}\mathcal{H}^T\|_2^2  =  \sum_{j=1}^{n} \langle \mathcal{W}_{i:}, \mathcal{H}_{j:} \rangle^2\\
				& = \sum_{j=1}^{n} \bigg(1 \wedge \frac{B}{\sqrt{m\|L_H\|_\text{op}}\|\widetilde{\mathcal{W}}_{i:}\widetilde{\mathcal{H}}^T\|_2}\bigg)^2 \langle \widetilde{\mathcal{W}}_{i:}, \widetilde{\mathcal{H}}_{j:} \rangle^2 \bigg(1 \wedge \frac{B}{\sqrt{n\|L_W\|_\text{op}}\|\widetilde{\mathcal{H}}_{j:}\widetilde{\mathcal{W}}^T\|_2}\bigg)^2\\
				& \leq  \bigg(1 \wedge \frac{B}{\sqrt{m\|L_H\|_\text{op}}\|\widetilde{\mathcal{W}}_{i:}\widetilde{\mathcal{H}}^T\|_2}\bigg)^2\sum_{j=1}^{n} \langle \widetilde{\mathcal{W}}_{i:}, \widetilde{\mathcal{H}}_{j:} \rangle^2 
				= \bigg(1 \wedge \frac{B}{\sqrt{m\|L_H\|_\text{op}}\|\widetilde{\mathcal{W}}_{i:}\widetilde{\mathcal{H}}^T\|_2}\bigg)^2\|\widetilde{\mathcal{W}}_{i:}\widetilde{\mathcal{H}}^T\|_2^2\\
				& = \bigg(1 \wedge \frac{B}{\sqrt{m\|L_H\|_\text{op}}\|\widetilde{\mathcal{W}}_{i:}\widetilde{\mathcal{H}}^T\|_2}\bigg)^2\|\widetilde{\mathcal{W}}_{i:}\widetilde{H}^TL_H^{\frac{1}{2}}\|_2^2
				\leq \|L_H\|_\text{op} \|\widetilde{\mathcal{W}}_{i:}\widetilde{H}^T\|_2^2 \bigg(1 \wedge \frac{B}{\sqrt{m\|L_H\|_\text{op}}\|\widetilde{\mathcal{W}}_{i:}\widetilde{\mathcal{H}}^T\|_2}\bigg)^2
				\leq \frac{B^2}{m},
			\end{aligned}
		\end{equation}
		which implies that $ \|L_W^{\frac{1}{2}}WH^T\|_{2,\infty}^2 \leq \frac{B^2}{m} $. The bound $ \|L_H^{\frac{1}{2}}HW^T\|_{2,\infty}^2 \leq \frac{B^2}{n} $ can also be achieved in similar way. Combining the two bounds gives  the graph incoherence condition.
		Now we complete the proof of Theorem 1.
		\Halmos 
	\end{proof}	
	
	\section{Proof of Theorem 2}
	\begin{proof}{Proof of Theorem 2}
		As in \citep{tong2021accelerating}, we first introduce two lemmas stating the computational properties of the orthogonal projection operator $ \mathcal{P}_\Omega(\cdot) $, where we use $ \mathcal{I}(\cdot) $ to denote the identity projection meaning that $ \mathcal{I}(X) = X $. 
		\begin{lemma}[\citep{zheng2016convergence}, Lemma 4; \citep{tong2021accelerating}, Lemma 35]\label{lemma proof1}
			Suppose that $ X_\star $ is $ (\mu; \beta, \lambda) $-graph incoherent, and $ p \gtrsim  \mu r \log(m \vee n)/(m \wedge n) $, then the following bound holds with overwhelming probability:
			\begin{equation*}
				\begin{aligned}
					&\left |  \big \langle (p^{-1}\mathcal{P}_\Omega - \mathcal{I})(W_\star H_A^T + W_AH_\star ^T), W_\star H_B^T + W_BH_\star ^T \big \rangle  \right | \\
					&\leq C_1\sqrt{\frac{\mu r \log(m \vee n)}{p(m \wedge n)}}\|W_\star H_A^T + W_AH_\star ^T\|_F\|W_\star H_B^T + W_BH_\star ^T\|_F,
				\end{aligned}
			\end{equation*}
			simultaneously for all $ W_A, W_B \in \mathbb{R}^{m\times r} $ and $ H_A, H_B \in \mathbb{R}^{n\times r} $, where $ c_1 > 0 $ is some universal constant.
		\end{lemma}
		
		\begin{lemma}[\citep{chen2019model}, Lemma 8;  \citep{tong2021accelerating}, Lemma 36]\label{lemma proof2}
			Suppose that $ p \gtrsim  \log(m \vee n)/(m \wedge n) $, then the following bound holds with overwhelming probability:
			\begin{equation*}
				\begin{aligned}
					&\left |  \big \langle (p^{-1}\mathcal{P}_\Omega - \mathcal{I})(W_AH_A ^T), W_B H_B^T  \big \rangle  \right | \\
					&\leq C_2\sqrt{\frac{m \vee n}{p}}\big(\|W_A\|_F\|W_B\|_{2,\infty} \wedge \|W_A\|_{2,\infty}\|W_B\|_F\big)\big(\|H_A\|_F\|H_B\|_{2,\infty} \wedge \|H_A\|_{2,\infty}\|H_B\|_F\big),
				\end{aligned}
			\end{equation*}
			simultaneously for all $ W_A, W_B \in \mathbb{R}^{m\times r} $ and $ H_A, H_B \in \mathbb{R}^{n\times r} $, where $ c_1 > 0 $ is some universal constant.
		\end{lemma}
		
		We then define a event $ \mathcal{E} $ as that the two bounds in Lemma \ref{lemma proof1} and Lemma \ref{lemma proof2} hold simultaneously, which happens with overwhelming probability. The rest of the proof is performed under the event $ \mathcal{E} $, as stated in Theorem 2.
		
		Based on the condition $ \text{dist}(F_t,F_\star) \leq 0.02(1+\beta)\sigma_r(X_\star) $, Lemma \ref{Q exist}  guarantees the existence of the optimal alignment matrix $ Q_t $ between $ F_t $ and $ F_\star $. We denote $ W := W_tQ_t $, $ H := H_tQ_t^{-T} $, $ \Delta_W := W - W_\star $, $ \Delta_H := H - H_\star $ and $ \epsilon := 0.02(1+\beta) $. Let $ \widetilde{F}_{t+1} =  	
		[\widetilde{W}_{t+1}^T, \widetilde{H}_{t+1}^T]^T $ as the update before projection,
		then we have $ F_{t+1} =  \mathcal{P}_B(\widetilde{F}_{t+1}) $. It is worth noting that in the rest of the proof we first concentrate on proving the following conclusion:
		$
		\text{dist}(\widetilde{F}_{t+1},F_\star) \leq (1-\gamma\eta)\text{dist}(F_t,F_\star),
		$
		based on which Theorem 1 guarantees the relation $ \text{dist}(F_{t+1},F_\star) \leq (1-\gamma\eta)\text{dist}(F_t,F_\star) $
		and  the graph incoherence condition 
		$ 
		\sqrt{m}\|L_W^{\frac{1}{2}}W_{t+1}H_{t+1}^T\|_{2,\infty} \vee \sqrt{n}\|L_H^{\frac{1}{2}}H_{t+1}W_{t+1}^T\|_{2,\infty} \leq B. $ We first list some useful bounds in the following lemma.
		\begin{lemma} \label{useful bounds}
			Under the conditions $ \text{dist}(F_t,F_\star) \leq \epsilon\sigma_r(X_\star) $ and 
			$ \sqrt{m}\|L_W^{\frac{1}{2}}WH^T\|_{2,\infty} \vee \sqrt{n}\|L_H^{\frac{1}{2}}HW^T\|_{2,\infty} \leq  C_B\sqrt{\mu r (1+\beta)}\sigma_1(X_\star)  $, the following bounds hold:
			\vspace{-5mm}
			\begin{subequations}
				\renewcommand{\theequation}{\theparentequation\alph{equation}}
				\begin{align}
					\|\Delta_W\Sigma_\star^{-1/2}\|_\text{op} \vee \|\Delta_H\Sigma_\star^{-1/2}\|_\text{op} \leq  \epsilon; ~~
					\|L_W^{1/2}\Delta_W\Sigma_\star^{-1/2}\|_\text{op} &\vee \|L_H^{1/2}\Delta_H\Sigma_\star^{-1/2}\|_\text{op} \leq \epsilon; \label{eq:1b} \\
					\|H(H^TH)^{-1}\Sigma_\star^{1/2}\|_\text{op}  \leq \frac{1}{1-\epsilon}; ~~
					\|\Sigma_\star^{1/2}(H^TH)^{-1}\Sigma_\star^{1/2}\|_\text{op}  &\leq \frac{1}{(1-\epsilon)^2}; \label{eq:1d}\\
					\sqrt{m}\|L_W^{\frac{1}{2}}W\Sigma_\star^{\frac{1}{2}}\|_{2,\infty} \vee \sqrt{n}\|L_H^{\frac{1}{2}}H\Sigma_\star^{\frac{1}{2}}\|_{2,\infty} &\leq \frac{\sqrt{1+\beta}}{1-\epsilon}C_B\sqrt{\mu r}\sigma_1(X_\star); \label{eq:1e}\\
					\sqrt{m}\|L_W^{\frac{1}{2}}W\Sigma_\star^{-\frac{1}{2}}\|_{2,\infty} \vee \sqrt{n}\|L_H^{\frac{1}{2}}H\Sigma_\star^{-\frac{1}{2}}\|_{2,\infty}  &\leq \frac{\sqrt{1+\beta}}{1-\epsilon}\kappa C_B\sqrt{\mu r}; \label{eq:1f}\\
					\sqrt{m}\|L_W^{\frac{1}{2}}\Delta_W\Sigma_\star^{\frac{1}{2}}\|_{2,\infty} \vee \sqrt{n}\|L_H^{\frac{1}{2}}\Delta_H\Sigma_\star^{\frac{1}{2}}\|_{2,\infty} &\leq \Big(1+ \frac{C_B\sqrt{1+\beta}}{1-\epsilon}\Big)\sqrt{\mu r}\sigma_1(X_\star); \label{eq:1g}
				\end{align}
			\end{subequations}		
		\end{lemma}
		\vspace{-5mm}
		
		Denote $ Q_t $ as the optimal alignment matrix between $ F_t $ and $ F_\star $, then we have 
		\begin{equation}\label{error metric comletion}
			\begin{aligned}
				\text{dist}^2(\widetilde{F}_{t+1},F_\star) \leq \|L_W^{1/2}(\widetilde{W}_{t+1}Q_t-W_\star)\Sigma_\star^{1/2}\|_F^2 + \|L_H^{1/2}(\widetilde{H}_{t+1}Q_t^{-T}-H_\star)\Sigma_\star^{1/2}\|_F^2.
			\end{aligned}
		\end{equation} 
		We first bound the first term $ \|L_W^{1/2}(\widetilde{W}_{t+1}Q_t-W_\star)\Sigma_\star^{1/2}\|_F^2 $. 
		Based on the update rules, we have
		\begin{equation*}
			\begin{aligned}
				&L_W^{1/2}(\widetilde{W}_{t+1}Q_t-W_\star)\Sigma_\star^{1/2}
				= L_W^{1/2}\left (\{W_t - \eta p^{-1} L_W \mathcal{P}_\Omega(W_tH_t^T-X_\star)H_t(H_t^TH_t)^{-1} \}Q_t - W_\star\right )\Sigma_\star^{1/2}\\
				& = L_W^{1/2}\left (W - \eta p^{-1} L_W \mathcal{P}_\Omega(WH^T-X_\star)H(H^TH)^{-1}  - W_\star\right )\Sigma_\star^{1/2}\\
				& = L_W^{1/2}\left (\Delta_W - \eta p^{-1} L_W \mathcal{P}_\Omega(WH^T-X_\star)H(H^TH)^{-1} \right )\Sigma_\star^{1/2}\\
				& = L_W^{1/2}\left (\Delta_W - \eta L_W (WH^T-X_\star)H(H^TH)^{-1} - \eta L_W (p^{-1}\mathcal{P}_\Omega-\mathcal{I})(WH^T-X_\star)H(H^TH)^{-1} \right )\Sigma_\star^{1/2}\\
				& \overset{(\text{\romannumeral1})}{=} L_W^{1/2}( I - \eta L_W)\Delta_W \Sigma_\star^{1/2} - \eta L_W^{3/2} W_\star\Delta_H^TH(H^TH)^{-1}\Sigma_\star^{1/2}  -\eta L_W^{3/2} (p^{-1}\mathcal{P}_\Omega-\mathcal{I})(WH^T-X_\star)H(H^TH)^{-1}\Sigma_\star^{1/2},
			\end{aligned}
		\end{equation*}
		where in $ (\text{\romannumeral1}) $ we utilize the decomposition $ WH^T-X_\star = \Delta_WH^T + W_\star\Delta_H^T $. Then the first term of (\ref{error metric comletion}) can be expanded as
		\begin{equation}\label{first term completion}
			\begin{aligned}
				&\|L_W^{1/2}(\widetilde{W}_{t+1}Q_t-W_\star)\Sigma_\star^{1/2}\|_F^2
				= \underbrace{ \|L_W^{1/2}( I - \eta L_W)\Delta_W \Sigma_\star^{1/2} - \eta L_W^{3/2} W_\star\Delta_H^TH(H^TH)^{-1}\Sigma_\star^{1/2} \|_F^2}_{\mathfrak{R}_1}\\
				& ~~~~~~~~~~~~~~~ - 2\eta\underbrace{ \text{tr}\big(L_W^{3/2} (p^{-1}\mathcal{P}_\Omega-\mathcal{I})(WH^T-X_\star)H(H^TH)^{-1}\Sigma_\star\Delta_W^T( I - \eta L_W)L_W^{1/2}\big)}_{\mathfrak{R}_2}\\
				& ~~~~~~~~~~~~~~~ + 2\eta^2\underbrace{\text{tr}\big(L_W^{3/2} (p^{-1}\mathcal{P}_\Omega-\mathcal{I})(WH^T-X_\star)H(H^TH)^{-1}\Sigma_\star (H^TH)^{-1}H^T\Delta_H W_\star^T L_W^{3/2}\big)}_{\mathfrak{R}_3}\\
				& ~~~~~~~~~~~~~~~ + \eta^2\underbrace{\|L_W^{3/2} (p^{-1}\mathcal{P}_\Omega-\mathcal{I})(WH^T-X_\star)H(H^TH)^{-1}\Sigma_\star^{1/2}\|_F^2}_{\mathfrak{R}_4}.
			\end{aligned}
		\end{equation}
		Next we bound the four terms in sequence.
		
		1. Controlling $ \mathfrak{R}_1 $: 
		It is easy to see that $ \mathfrak{R}_1 $ can be decomposed as
		\begin{equation}
			\begin{aligned}
				\mathfrak{R}_1 &=  \underbrace{\text{tr}\big(L_W^{1/2}( I - \eta L_W)\Delta_W\Sigma_\star \Delta_W^T( I - \eta L_W)L_W^{1/2}\big)}_{\mathfrak{F}_1}+ \eta^2\underbrace{\big\|L_W^{3/2}W_\star \Delta_H^TH(H^TH)^{-1}\Sigma_\star^{1/2}\big\|_F^2}_{\mathfrak{F}_2} \\
				&~~~-2\eta\underbrace{\text{tr}\big(W_\star\Delta_H^TH(H^TH)^{-1}\Sigma_\star\Delta_W^T( I - \eta L_W)L_W^{2}\big)}_{\mathfrak{F}_3}.
			\end{aligned}
		\end{equation}
		In the following we first build a useful lemma, and then focus on controlling $ \mathfrak{F}_1 $, $ \mathfrak{F}_2 $ and $ \mathfrak{F}_3 $ in sequence.
		\begin{lemma}\label{lemma Q obeys}
			For any stacked factor matrix $ F := 
			[W^T,H^T]^T
			\in \mathbb{R}^{(m+n)\times r} $, if the optimal alignment matrix $ Q $ between $ F $ and $ F_\star $ exists, then $ Q $ satisfies
			$
			Q^TW^TL_W(WQ-W_\star)\Sigma_\star = \Sigma_\star(HQ^{-T}-H_\star)^TL_HHQ^{-T}.
			$
		\end{lemma}
		
		(1) Controlling $ \mathfrak{F}_1 $: 
		we decompose $ \mathfrak{F}_1 $ into several items as follows:
		\begin{equation}
			\begin{aligned}
				\mathfrak{F}_1 &= \text{tr}\big(L_W^{1/2}\Delta_W\Sigma_\star \Delta_W^TL_W^{1/2}\big) - 2\eta \text{tr}\big(L_W^{1/2}L_W\Delta_W\Sigma_\star \Delta_W^TL_W^{1/2}\big)+ \eta^2 \text{tr}\big(L_W^{1/2}L_W\Delta_W\Sigma_\star \Delta_W^TL_WL_W^{1/2}\big)\\
				& = \text{tr}\big(L_W^{1/2}\Delta_W\Sigma_\star \Delta_W^TL_W^{1/2}\big) - 2\eta \text{tr}\big(L_W\Delta_W\Sigma_\star \Delta_W^TL_W\big)  + \eta^2 \text{tr}\big(L_W^{3/2}\Delta_W\Sigma_\star \Delta_W^TL_W^{3/2}\big),
			\end{aligned}
		\end{equation}
		which will be further analyzed in subsequent parts.
		
		(2) Controlling $ \mathfrak{F}_2 $:
		from the hypothesis 
		$
		\text{dist}(F_{t},F_\star) =  \sqrt{\|L_W^{1/2}\Delta_W\Sigma_\star^{1/2}\|_F^2 + \|L_H^{1/2}\Delta_H\Sigma_\star^{1/2}\|_F^2}\leq \epsilon\sigma_r(X_\star),
		$
		we have
		\begin{equation}\label{dist epsilon}
			\begin{aligned}
				&\sigma_r(X_\star)\sqrt{\|L_W^{1/2}\Delta_W\Sigma_\star^{-1/2}\|_F^2 + \|L_H^{1/2}\Delta_H\Sigma_\star^{-1/2}\|_F^2}
				\leq \sqrt{\|L_W^{1/2}\Delta_W\Sigma_\star^{-1/2}\Sigma_\star\|_F^2 + \|L_H^{1/2}\Delta_H\Sigma_\star^{-1/2}\Sigma_\star\|_F^2}\\
				&~~~~~~~~~~\leq \sqrt{\|L_W^{1/2}\Delta_W\Sigma_\star^{1/2}\|_F^2 + \|L_H^{1/2}\Delta_H\Sigma_\star^{1/2}\|_F^2}
				\leq \epsilon\sigma_r(X_\star),
			\end{aligned}
		\end{equation}
		which implies
		$
		\|L_W^{1/2}\Delta_W\Sigma_\star^{-1/2}\|_F \vee \|L_H^{1/2}\Delta_H\Sigma_\star^{-1/2}\|_F \leq \epsilon,
		$
		and thus
		$
		\|L_W^{1/2}\Delta_W\Sigma_\star^{-1/2}\|_\text{op} \vee \|L_H^{1/2}\Delta_H\Sigma_\star^{-1/2}\|_\text{op} \leq \epsilon
		$
		due to the relation $ \|A\|_\text{op} \leq \|A\|_F $. Taking $ 1 \leq \sigma(L_W) \leq 1+\beta$ and $ 1 \leq \sigma(L_H) \leq1+\beta$ into account, it can be  verified that
		\begin{equation}\label{alpha t}
			\begin{aligned}
				\frac{1}{1+\beta} \leq \frac{ \|\Delta_W\Sigma_\star^{1/2}\|_F^2 + \|\Delta_H\Sigma_\star^{1/2}\|_F^2}{ \|L_W^{1/2}\Delta_W\Sigma_\star^{1/2}\|_F^2 + \|L_H^{1/2}\Delta_H\Sigma_\star^{1/2}\|_F^2} \leq 1.
			\end{aligned}
		\end{equation}
		Substituting (\ref{alpha t}) into (\ref{dist epsilon}) to get that 
		\begin{equation*}
			\begin{aligned}
				&\sigma_r(X_\star)\sqrt{\|\Delta_W\Sigma_\star^{-1/2}\|_F^2 + \|\Delta_H\Sigma_\star^{-1/2}\|_F^2}
				\leq\sqrt{\|\Delta_W\Sigma_\star^{-1/2}\Sigma_\star\|_F^2 + \|\Delta_H\Sigma_\star^{-1/2}\Sigma_\star\|_F^2}\\
				&~~~~~~~~~~ =\sqrt{\|\Delta_W\Sigma_\star^{1/2}\|_F^2 + \|\Delta_H\Sigma_\star^{1/2}\|_F^2}
				\leq \sqrt{\|L_W^{1/2}\Delta_W\Sigma_\star^{1/2}\|_F^2 + \|L_H^{1/2}\Delta_H\Sigma_\star^{1/2}\|_F^2}
				\leq \epsilon\sigma_r(X_\star),
			\end{aligned}
		\end{equation*}
		which implies
		$
		\|\Delta_W\Sigma_\star^{-1/2}\|_F \vee \|\Delta_H\Sigma_\star^{-1/2}\|_F \leq \epsilon,
		$
		and
		$
		\|\Delta_W\Sigma_\star^{-1/2}\|_\text{op} \vee \|\Delta_H\Sigma_\star^{-1/2}\|_\text{op} \leq \epsilon.
		$
		Then we can bound $ \mathfrak{F}_2 $ as
		\begin{equation*}
			\begin{aligned}
				\mathfrak{F}_2 &= \|L_W^{3/2}W_\star \Delta_H^TH(H^TH)^{-1}\Sigma_\star^{1/2}\|_F^2
				\leq  (1+\beta)^\frac{3}{2}\text{tr}\big(W_\star \Delta_H^TH(H^TH)^{-1}\Sigma_\star(H^TH)^{-1}H^T\Delta_HW_\star^T\big)\\
				&\overset{(\text{\romannumeral1})}{=}(1+\beta)^\frac{3}{2} \text{tr}\big( H(H^TH)^{-1}\Sigma_\star(H^TH)^{-1}H^T\Delta_H\Sigma_\star\Delta_H^T\big)\\
				& \overset{(\text{\romannumeral2})}{=}(1+\beta)^\frac{3}{2} \text{tr}\big( H(H^TH)^{-1}H^T\Delta_H\Sigma_\star\Delta_H^T\big) - (1+\beta)^\frac{3}{2}\text{tr}\big( H(H^TH)^{-1}(H^TH - \Sigma_\star)(H^TH)^{-1}H^T\Delta_H\Sigma_\star\Delta_H^T\big)\\
				& \overset{(\text{\romannumeral3})}{\leq}(1+\beta)^\frac{3}{2} \Big(\underbrace{\text{tr}\big( H(H^TH)^{-1}H^TL_H\Delta_H\Sigma_\star\Delta_H^T\big)}_{\mathfrak{F}_2^{(\text{\romannumeral1})}}  -  \underbrace{\text{tr}\big( H(H^TH)^{-1}(H^TH - \Sigma_\star)(H^TH)^{-1}H^T\Delta_H\Sigma_\star\Delta_H^T\big)}_{\mathfrak{F}_2^{(\text{\romannumeral2})}}\Big),
			\end{aligned}
		\end{equation*}
		where in equations $ (\text{\romannumeral1}) $, $ (\text{\romannumeral2}) $ and $ (\text{\romannumeral3}) $ we utilize $ W_\star^TW_\star = \Sigma_\star $, $ \Sigma_\star = H^TH - (H^TH -\Sigma_\star ) $, and the maximum singular value of $ L_H $, $ \sigma_{\text{max}}(L_H) < 1+\beta $, respectively. For $ \mathfrak{F}_2^{(\text{\romannumeral1})} $, it is easy to verify that  $  H(H^TH)^{-1}H^TL_H\Delta_H\Sigma_\star\Delta_H^T $ is a positive semi-definite matrix, and thus we have $ \mathfrak{F}_2^{(\text{\romannumeral1})} \geq 0$.
		$ \mathfrak{F}_2^{(\text{\romannumeral2})} $ can be controlled by
		\begin{equation}
			\begin{aligned}
				\lvert \mathfrak{F}_2^{(\text{\romannumeral2})}\rvert &\leq \|H(H^TH)^{-1}(H^TH - \Sigma_\star)(H^TH)^{-1}H^T\|_\text{op}\text{tr}\big( \Delta_H\Sigma_\star\Delta_H^T\big)\\
				& \leq \|H(H^TH)^{-1}\Sigma_\star^{1/2}\|_\text{op}^2\|\Sigma_\star^{-1/2}(H^TH - \Sigma_\star)\Sigma_\star^{-1/2}\|_\text{op}\text{tr}\big( \Delta_H\Sigma_\star\Delta_H^T\big), 
			\end{aligned}
		\end{equation}
		and we can then bound the three terms in the following. Based on the
		notice that
		$
		\|H(H^TH)^{-1}\Sigma_\star^{1/2}\|_\text{op} = \frac{1}{\sigma_r(H\Sigma_\star^{-1/2})},
		$
		utilizing the Weyl's inequality $ \lvert \sigma_r(A) -  \sigma_r(B)\rvert \leq \| A - B \|_\text{op}$ and the fact $ \sigma_r(H_\star\Sigma_\star^{-1/2}) = \sigma_r(V_\star) = 1 $, we can obtain
		\begin{equation}
			\sigma_r(H\Sigma_\star^{-1/2}) \geq \sigma_r(H_\star\Sigma_\star^{-1/2}) - \|\Delta_H\Sigma_\star^{-1/2}\|_\text{op}\geq 1-\|\Delta_H\Sigma_\star^{-1/2}\|_\text{op},
		\end{equation}
		which gives  a bound of the first term:
		$
		\|H(H^TH)^{-1}\Sigma_\star^{1/2}\|_\text{op} \leq \frac{1}{1 -\|\Delta_H\Sigma_\star^{-1/2}\|_\text{op}} \leq \frac{1}{1-\epsilon}.
		$
		The second term is controlled by
		\begin{equation}
			\begin{aligned}
				&\|\Sigma_\star^{-1/2}(H^TH - \Sigma_\star)\Sigma_\star^{-1/2}\|_\text{op}= \|\Sigma_\star^{-1/2}(H_\star^T\Delta_H+\Delta_H^TH_\star+\Delta_H^T\Delta_H)\Sigma_\star^{-1/2}\|_\text{op}\\
				& \leq \|U_\star^T\Delta_H\Sigma_\star^{-1/2}\|_\text{op}  + \|\Sigma_\star^{-1/2}\Delta_H^TU_\star\|_\text{op} + \|\Sigma_\star^{-1/2}\Delta_H^T\Delta_H\Sigma_\star^{-1/2}\|_\text{op}
				= 2\|\Delta_H\Sigma_\star^{-1/2}\|_\text{op} + \|\Delta_H\Sigma_\star^{-1/2}\|_\text{op}^2
				\leq 2\epsilon+\epsilon^2.
			\end{aligned}
		\end{equation}
		Combining the above gives
		$
		\lvert \mathfrak{F}_2^{(\text{\romannumeral2})}\rvert \leq \frac{2\epsilon+\epsilon^2}{(1-\epsilon)^2}\text{tr}\big( \Delta_H\Sigma_\star\Delta_H^T\big).
		$
		
		(3) Controlling $ \mathfrak{F}_3 $:
		to bound $ \mathfrak{F}_3 $, we first invoke the
		decomposition $ W_\star = W - \Delta_W $ to get
		\begin{equation}
			\begin{aligned}
				&\text{tr}\big(W_\star\Delta_H^TH(H^TH)^{-1}\Sigma_\star \Delta_W^T( I - \eta L_W)L_W^2\big)
				= \text{tr}(H(H^TH)^{-1}\Sigma_\star \Delta_W^T( I - \eta L_W)L_W^2W_\star\Delta_H^T)\\
				& = \underbrace{\text{tr}(H(H^TH)^{-1}\Sigma_\star \Delta_W^T( I - \eta L_W)L_W^2W\Delta_H^T)}_{\mathfrak{F}_3^{(\text{\romannumeral1})}} - \underbrace{\text{tr}(H(H^TH)^{-1}\Sigma_\star \Delta_W^T( I - \eta L_W)L_W^2\Delta_W\Delta_H^T)}_{\mathfrak{F}_3^{(\text{\romannumeral2})}},
			\end{aligned}
		\end{equation}
		then $ \mathfrak{F}_3^{(\text{\romannumeral1})} $ and $ \mathfrak{F}_3^{(\text{\romannumeral2})} $ can be bounded as follows.  
		For $ \mathfrak{F}_3^{(\text{\romannumeral1})} $, invoke Lemma \ref{lemma Q obeys} to get $ \Sigma_\star \Delta_W^TL_WW = H^TL_H\Delta_H\Sigma_\star $, then $ H(H^TH)^{-1}H^TL_H\Delta_H\Sigma_\star\Delta_H^T = H(H^TH)^{-1}\Sigma_\star \Delta_W^TL_WW\Delta_H^T $. Obviously, $ H(H^TH)^{-1}H^TL_H\Delta_H\Sigma_\star\Delta_H^T $ is  positive semi-definite, and thus $ H(H^TH)^{-1}\Sigma_\star \Delta_W^TL_WW\Delta_H^T $ is also positive semi-definite. On the condition $ \eta \leq \frac{1}{\sigma_\text{max}(L_W)} $, we have
		\begin{equation}
			\begin{aligned}
				\mathfrak{F}_3^{(\text{\romannumeral1})} &= \text{tr}(H(H^TH)^{-1}\Sigma_\star \Delta_W^T( I - \eta L_W)L_W^2W\Delta_H^T)
				\geq \sigma_\text{min}\big(L_W( I - \eta L_W)\big)\text{tr}(H(H^TH)^{-1}\Sigma_\star \Delta_W^TL_WW\Delta_H^T)\\
				& = \sigma_\text{min}\big(L_W( I - \eta L_W)\big) \text{tr}(H(H^TH)^{-1}H^TL_H\Delta_H\Sigma_\star\Delta_H^T)
				\geq 0,
			\end{aligned}
		\end{equation}
		where $ \sigma_\text{min}(\cdot) $ denotes the minimum singular value.
		Denote $ \zeta := \sigma_\text{min}\big(L_W( I - \eta L_W)\big) $, then we analysis the value of $ \zeta $. Let $ \sigma_1 \geq \sigma_2 \geq \cdots  \geq \sigma_m=0 $ be the singular values of Laplacian matrix $ \widetilde{L}_W $ in descending order. $ L_W =  (1+\beta)I_m - \beta(I_m + \lambda\widetilde{L}_W)^{-1} $ implies that the singular values of $ L_W $ consist of  $ 1+\beta-\frac{\beta}{1+\lambda\sigma_1} \geq 1+\beta-\frac{\beta}{1+\lambda\sigma_2} \geq \cdots  \geq 1+\beta-\frac{\beta}{1+\lambda\sigma_m} = 1 $, giving rise to that $ \sigma_\text{min}(L_W) = 1 $ and $ \sigma_\text{max}(L_W) =  1+\beta-\frac{\beta}{1+\lambda\sigma_1} \leq 1+\beta $. Denote $ \sigma $ as one of the singular values of $ L_W $, then  $ \sigma\in [1,1+\beta] $, and the singular value of matrix $ L_W( I - \eta L_W) $ at the corresponding position is $ \sigma - \eta\sigma^2 $. Denote $ \widehat{\zeta} $ is the minimum value of objective $ \sigma - \eta\sigma^2 $ on the interval $ [1,1+\beta] $, then we have $ \zeta \leq \widehat{\zeta} $ . Considering that objective $ \sigma - \eta\sigma^2 $ is a downward parabola, its minimum value must be obtained at $ \sigma = 1 $ or $ \sigma = 1+\beta $, and thus we have $  \widehat{\zeta} = \minimize \{1-\eta, (1+\beta)-\eta(1+\beta)^2\} $. Meanwhile, we let $ \eta \leq \frac{1}{1+\beta} $, then the condition $ \eta \leq \frac{1}{\sigma_\text{max}(L_W)} $ can be satisfied.
		
		For $ \mathfrak{F}_3^{(\text{\romannumeral2})} $, we have
		\begin{equation*}
			\begin{aligned}
				\lvert \mathfrak{F}_3^{(\text{\romannumeral2})} \rvert  
				& = \lvert \text{tr}(\Sigma_\star^{-1/2}\Delta_H^TH(H^TH)^{-1}\Sigma_\star \Delta_W^T( I - \eta L_W)L_W^2\Delta_W\Sigma_\star^{1/2})\rvert \\
				& \leq \|\Sigma_\star^{-1/2}\Delta_H^TH(H^TH)^{-1}\Sigma_\star^{1/2}\|_\text{op} \text{tr}(\Sigma_\star^{1/2} \Delta_W^T( I - \eta L_W)L_W^2\Delta_W\Sigma_\star^{1/2})\\
				& \leq \|\Delta_H\Sigma_\star^{-1/2}\|_\text{op}\|H(H^TH)^{-1}\Sigma_\star^{1/2}\|_\text{op}\big(\text{tr}(L_W\Delta_W\Sigma_\star \Delta_W^TL_W) - \eta \text{tr}(L_W^{3/2}\Delta_W\Sigma_\star \Delta_W^TL_W^{3/2})\big).
			\end{aligned} 
		\end{equation*}
		Invoking $ \|\Delta_H\Sigma_\star^{-1/2}\|_\text{op} \leq \alpha_t\|H(H^TH)^{-1}\Sigma_\star^{1/2}\|_\text{op} \leq \frac{1}{1-\epsilon} $ in the above, $ \lvert \mathfrak{F}_3^{(\text{\romannumeral2})} \rvert $ can be bounded by
		\begin{equation}
			\lvert \mathfrak{R}_3^{(\text{\romannumeral2})} \rvert \leq \frac{\epsilon}{1-\epsilon}\big(\text{tr}(L_W\Delta_W\Sigma_\star \Delta_W^TL_W) - \eta \text{tr}(L_W^{3/2}\Delta_W\Sigma_\star \Delta_W^TL_W^{3/2})\big).
		\end{equation}
		
		(4) Combination:
		combining the bounds for $ \mathfrak{F}_1 $, $ \mathfrak{F}_2 $, $ \mathfrak{F}_3 $, we can obtain 
		\begin{equation}\label{combining matrix decomposition}
			\begin{aligned}
				\mathfrak{R}_1
				& \leq  \text{tr}\big(L_W^{1/2}\Delta_W\Sigma_\star \Delta_W^TL_W^{1/2}\big) - 2\eta \text{tr}\big(L_W\Delta_W\Sigma_\star \Delta_W^TL_W\big)  + \eta^2 \text{tr}\big(L_W^{3/2}\Delta_W\Sigma_\star \Delta_W^TL_W^{3/2}\big)  \\
				& ~~~+ (\eta^2(1+\beta)^\frac{3}{2}-2\eta\zeta)\text{tr}\big( H(H^TH)^{-1}H^TL_H\Delta_H\Sigma_\star\Delta_H^T\big)+2\eta\frac{\epsilon}{1-\epsilon}\text{tr}\big(L_W\Delta_W\Sigma_\star \Delta_W^TL_W\big) \\
				& ~~~ + \eta^2(1+\beta)^\frac{3}{2} \frac{2\epsilon+\epsilon^2}{(1-\epsilon)^2}\text{tr}\big( \Delta_H\Sigma_\star\Delta_H^T\big)  - 2\eta^2\frac{\epsilon}{1-\epsilon} \text{tr}\big(L_W^{3/2}\Delta_W\Sigma_\star \Delta_W^TL_W^{3/2}\big).
			\end{aligned}
		\end{equation}
		Considering that $ \text{tr}\big( H(H^TH)^{-1}H^TL_H\Delta_H\Sigma_\star\Delta_H^T\big) $ is a positive semi-definite matrix, we let $ \eta^2(1+\beta)^\frac{3}{2}-2\eta\zeta \leq 0 $.
		As previously analyzed, $ \zeta $ should satisfies $  \zeta \leq \widehat{\zeta} = \minimize \{1-\eta, (1+\beta)-\eta(1+\beta)^2\} $, thus we only need to ensure that the inequality $ \eta^2(1+\beta)^\frac{3}{2}-2\eta\zeta \leq 0 $ holds for both $ \zeta = 1-\eta $ and $ \zeta =  (1+\beta)-\eta(1+\beta)^2 $ simultaneously.
		The first condition implies that $ \eta^2(1+\beta)^\frac{3}{2}-2\eta(1-\eta) \leq 0 $, then we have
		$
		\eta \leq \frac{2}{2+(1+\beta)^\frac{3}{2}}.
		$
		The second condition is equivalent to that $ \sqrt{1+\beta}\eta - 2(1 - (1+\beta)\eta) \leq 0 $, leading to that 
		$
		\eta \leq \frac{2}{2(1+\beta)+\sqrt{(1+\beta)}}.
		$
		Combining these conditions together, we can get that $ \eta $ should satisfies 
		$
		\eta \leq \minimize \Big\{\frac{2}{2+(1+\beta)^\frac{3}{2}}, \frac{2}{2(1+\beta)+\sqrt{(1+\beta)}}, \frac{1}{1+\beta}\Big\}.
		$
		Obviously,  for $ 0 < \beta \leq 1 $,  $ \frac{2}{2+(1+\beta)^\frac{3}{2}} \geq \frac{2}{2(1+\beta)+\sqrt{(1+\beta)}} $ and $ \frac{1}{1+\beta} \geq \frac{2}{2(1+\beta)+\sqrt{(1+\beta)}} $ hold, and thus the bound of $ \eta $ can be simplified by
		$
		\eta \leq \frac{2}{2(1+\beta)+\sqrt{(1+\beta)}}.
		$
		Thus, with $ 0 < \beta \leq 1 $ and $  0 < \eta \leq \frac{2}{2(1+\beta)+\sqrt{(1+\beta)}} $, we have
		$
		(\eta^2-2\eta\zeta)\text{tr}(H(H^TH)^{-1}H^TL_H\Delta_H\Sigma_\star\Delta_H^T) \leq 0,
		$
		and thus
		(\ref{combining matrix decomposition}) can be simplified as
		\begin{equation}\label{combining matrix decomposition2}
			\begin{aligned}
				\mathfrak{R}_1 
				& \leq  \big\|L_W^{\frac{1}{2}}\Delta_W\Sigma_\star^{\frac{1}{2}}\big\|_F^2  + \Big(\eta^2- 2\eta^2\frac{\epsilon}{1-\epsilon}\Big) \big\|L_W^{\frac{3}{2}}\Delta_W\Sigma_\star^{\frac{1}{2}}\big\|_F^2 \\
				& ~~~ +\Big(2\eta\frac{\epsilon}{1-\epsilon}-2\eta\Big)\big\|L_W\Delta_W\Sigma_\star^{\frac{1}{2}}\big\|_F^2  + \eta^2(1+\beta)^\frac{3}{2} \frac{2\epsilon+\epsilon^2}{(1-\epsilon)^2}\big\|\Delta_H\Sigma_\star^{\frac{1}{2}}\big\|_F^2.
			\end{aligned}
		\end{equation}

		2. Controlling $ \mathfrak{R}_2 $:\\
		\begin{equation}\label{key}
			\begin{aligned}
				|\mathfrak{R}_2| 
				& \overset{(\text{\romannumeral1})}{=} \big|\text{tr}\big(L_W^{3/2} (p^{-1}\mathcal{P}_\Omega-\mathcal{I})( \Delta_WH_\star^T + W\Delta_H^T)H(H^TH)^{-1}\Sigma_\star\Delta_W^T( I - \eta L_W)L_W^{1/2}\big)\big|\\
				&\overset{(\text{\romannumeral2})}{\leq} \underbrace{\big|\text{tr}\big( (p^{-1}\mathcal{P}_\Omega-\mathcal{I}) \Delta_WH_\star^TH_\star(H^TH)^{-1}\Sigma_\star\Delta_W^T( I - \eta L_W)L_W^2\big)\big|}_{\mathfrak{R}_2^{(\text{\romannumeral1})}}\\
				&+ \underbrace{\big|\text{tr}\big( (p^{-1}\mathcal{P}_\Omega-\mathcal{I}) \Delta_WH_\star^T\Delta_H(H^TH)^{-1}\Sigma_\star\Delta_W^T( I - \eta L_W)L_W^2\big)\big|}_{\mathfrak{R}_2^{(\text{\romannumeral2})}}\\
				&+ \underbrace{\big|\text{tr}\big( (p^{-1}\mathcal{P}_\Omega-\mathcal{I}) W\Delta_H^TH(H^TH)^{-1}\Sigma_\star\Delta_W^T( I - \eta L_W)L_W^2\big)\big|}_{\mathfrak{R}_2^{(\text{\romannumeral3})}},
			\end{aligned}
		\end{equation}
		where in  $ (\text{\romannumeral1}) $ we utilize the decomposition $ WH^T-X_\star = \Delta_WH_\star^T + W\Delta_H^T $, and in  $ (\text{\romannumeral2}) $ we employ triangle inequality and $ H = H_\star + \Delta_H $. For $ \mathfrak{R}_2^{(\text{\romannumeral1})} $, invoking Lemma \ref{lemma proof1} by $ W_A = \Delta_W $, $ W_B = L_W^2( I - \eta L_W)\Delta_W\Sigma_\star(H^TH)^{-1} $, $ H_A = H_B = 0 $, we have
		\begin{equation*}
			\begin{aligned}
				\mathfrak{R}_2^{(\text{\romannumeral1})} &\leq	C_1\sqrt{\frac{\mu r \log(m \vee n)}{p(m \wedge n)}}\|\Delta_WH_\star ^T\|_F\|L_W^2( I - \eta L_W)\Delta_W\Sigma_\star(H^TH)^{-1}H_\star ^T\|_F\\
				& \leq C_1\sqrt{\frac{\mu r \log(m \vee n)}{p(m \wedge n)}}\|\Delta_W\Sigma_\star ^\frac{1}{2}\|_F\|L_W^2( I - \eta L_W)\Delta_W\Sigma_\star^\frac{1}{2}\|_F\|\Sigma_\star^\frac{1}{2}(H^TH)^{-1}\Sigma_\star ^\frac{1}{2}\|_\text{op}\\
				& \leq C_1\sqrt{\frac{\mu r \log(m \vee n)}{p(m \wedge n)}}\|\Delta_W\Sigma_\star ^\frac{1}{2}\|_F\| I - \eta L_W\|_\text{op}\|L_W^2\Delta_W\Sigma_\star^\frac{1}{2}\|_F\|\Sigma_\star^\frac{1}{2}(H^TH)^{-1}\Sigma_\star ^\frac{1}{2}\|_\text{op}\frac{1}{(1-\epsilon)^2}\\
				& \leq C_1\sqrt{\frac{\mu r \log(m \vee n)}{p(m \wedge n)}}\frac{1-\eta}{(1-\epsilon)^2}\|\Delta_W\Sigma_\star ^\frac{1}{2}\|_F\|L_W^2\Delta_W\Sigma_\star^\frac{1}{2}\|_F.
			\end{aligned}
		\end{equation*}
		
		For $ \mathfrak{R}_2^{(\text{\romannumeral2})} $, we can invoke Lemma \ref{lemma proof2} by $ W_A = \Delta_W\Sigma_\star^\frac{1}{2} $, $ H_A = H_\star\Sigma_\star^{-\frac{1}{2}} $,  $ W_B = L_W^2( I - \eta L_W)\Delta_W\Sigma_\star^\frac{1}{2} $, $ H_B = \Delta_H(H^TH)^{-1}\Sigma_\star ^\frac{1}{2} $, leading to the following bound:
		\begin{equation*}
			\begin{aligned}
				\mathfrak{R}_2^{(\text{\romannumeral2})}	&\leq C_2\sqrt{\frac{m \vee n}{p}} \|\Delta_W\Sigma_\star^\frac{1}{2}\|_{2,\infty}\|L_W^2( I - \eta L_W)\Delta_W\Sigma_\star^\frac{1}{2}\|_F \|H_\star\Sigma_\star^{-\frac{1}{2}}\|_{2,\infty}\|\Delta_H(H^TH)^{-1}\Sigma_\star ^\frac{1}{2}\|_F\\
				&\leq C_2\sqrt{\frac{m \vee n}{p}} \|L_W^\frac{1}{2}\Delta_W\Sigma_\star^\frac{1}{2}\|_{2,\infty}(1-\eta)\|L_W^2\Delta_W\Sigma_\star^\frac{1}{2}\|_F \|L_W^\frac{1}{2}V_\star\|_{2,\infty}\|\Delta_H\Sigma_\star ^{-\frac{1}{2}}\|_F\|\Sigma_\star ^\frac{1}{2}(H^TH)^{-1}\Sigma_\star ^\frac{1}{2}\|_\text{op}\\
				& \leq C_2\sqrt{\frac{m \vee n}{p}}\frac{1}{\sqrt{m}}\Big(1+ \frac{C_B\sqrt{1+\beta}}{1-\epsilon}\Big)\sqrt{\mu r}\sigma_1(X_\star)(1-\eta)\|L_W^2\Delta_W\Sigma_\star^\frac{1}{2}\|_F\sqrt{\frac{\mu r}{n}}\|\Delta_H\Sigma_\star ^{\frac{1}{2}}\|_F\frac{1}{\sigma_r(X_\star)}\frac{1}{(1-\epsilon)^2}\\
				&\leq \frac{(1-\eta)\mu r}{\sqrt{p(m \wedge n)}}\frac{C_2\kappa}{(1-\epsilon)^2}\Big(1+ \frac{C_B\sqrt{1+\beta}}{1-\epsilon}\Big)\|\Delta_H\Sigma_\star ^{\frac{1}{2}}\|_F\|L_W^2\Delta_W\Sigma_\star^\frac{1}{2}\|_F.
			\end{aligned}
		\end{equation*}
		
		Invoking Lemma \ref{lemma proof2} by $ W_A = W\Sigma_\star^{-\frac{1}{2}} $, $ H_A = \Delta_H\Sigma_\star^{\frac{1}{2}} $,  $ W_B = L_W^2( I - \eta L_W)\Delta_W\Sigma_\star^\frac{1}{2} $, $ H_B = H(H^TH)^{-1}\Sigma_\star ^\frac{1}{2} $, $ \mathfrak{R}_2^{(\text{\romannumeral3})} $ can then be controlled by:
		\begin{equation*}
			\begin{aligned}
				\mathfrak{R}_2^{(\text{\romannumeral3})}	
				&\leq C_2\sqrt{\frac{m \vee n}{p}} \|W\Sigma_\star^{-\frac{1}{2}} \|_{2,\infty}\|L_W^2( I - \eta L_W)\Delta_W\Sigma_\star^\frac{1}{2}\|_F \|\Delta_H\Sigma_\star^{\frac{1}{2}}\|_F\|H(H^TH)^{-1}\Sigma_\star ^\frac{1}{2}\|_{2,\infty}\\
				&\leq C_2\sqrt{\frac{m \vee n}{p}} \|L_W^\frac{1}{2}W\Sigma_\star^{-\frac{1}{2}} \|_{2,\infty}( 1 - \eta)\|L_W^2\Delta_W\Sigma_\star^\frac{1}{2}\|_F \|\Delta_H\Sigma_\star^{\frac{1}{2}}\|_F\|H\Sigma_\star ^{-\frac{1}{2}}\|_{2,\infty}\|\Sigma_\star ^\frac{1}{2}(H^TH)^{-1}\Sigma_\star ^\frac{1}{2}\|_\text{op}\\
				&\leq C_2\sqrt{\frac{m \vee n}{p}} \frac{1}{\sqrt{m}}\frac{\sqrt{1+\beta}}{1-\epsilon}\kappa C_B\sqrt{\mu r}( 1 - \eta)\|L_W^2\Delta_W\Sigma_\star^\frac{1}{2}\|_F \|\Delta_H\Sigma_\star^{\frac{1}{2}}\|_F\frac{1}{\sqrt{n}}\frac{\sqrt{1+\beta}}{1-\epsilon}\kappa C_B\sqrt{\mu r}\frac{1}{(1-\epsilon)^2}\\
				&\leq \frac{(1-\eta)\mu r}{\sqrt{p(m \wedge n)}}\frac{C_2C_B^2\kappa^2(1+\beta)}{(1-\epsilon)^4}\|\Delta_H\Sigma_\star ^{\frac{1}{2}}\|_F\|L_W^2\Delta_W\Sigma_\star^\frac{1}{2}\|_F.
			\end{aligned}
		\end{equation*}
		
		Combining $ \mathfrak{R}_2^{(\text{\romannumeral1})} $, $ \mathfrak{R}_2^{(\text{\romannumeral2})} $ and $ \mathfrak{R}_2^{(\text{\romannumeral3})} $, we have
		\begin{equation}\label{R2 bound 1}
			\begin{aligned}
				\mathfrak{R}_2 &\leq C_1\sqrt{\frac{\mu r \log(m \vee n)}{p(m \wedge n)}}\frac{1-\eta}{(1-\epsilon)^2}\|\Delta_W\Sigma_\star ^\frac{1}{2}\|_F\|L_W^2\Delta_W\Sigma_\star^\frac{1}{2}\|_F \\
				&~~~~~+ \frac{(1-\eta)\mu r}{\sqrt{p(m \wedge n)}}\bigg(\frac{C_2\kappa}{(1-\epsilon)^2}\Big(1+ \frac{C_B\sqrt{1+\beta}}{1-\epsilon}\Big) + \frac{C_2C_B^2\kappa^2(1+\beta)}{(1-\epsilon)^4}\bigg)\|\Delta_H\Sigma_\star ^{\frac{1}{2}}\|_F\|L_W^2\Delta_W\Sigma_\star^\frac{1}{2}\|_F.
			\end{aligned}
		\end{equation}
		Denote 
		$
		\delta_1 := C_1\sqrt{\frac{\mu r \log(m \vee n)}{p(m \wedge n)}}\frac{1}{(1-\epsilon)^2}	,~~ \delta_2 := \frac{\mu r}{\sqrt{p(m \wedge n)}}\frac{C_2\kappa}{(1-\epsilon)^2}\Big(1+ \frac{C_B\sqrt{1+\beta}}{1-\epsilon} + \frac{C_B^2\kappa(1+\beta)}{(1-\epsilon)^2}\Big),
		$
		then (\ref{R2 bound 1}) can be rewritten as the following bound of $ \mathfrak{R}_2 $:
		\begin{equation}\label{R2 bound 2}
			\begin{aligned}
				\mathfrak{R}_2  &\leq \delta_1(1-\eta)\|\Delta_W\Sigma_\star ^\frac{1}{2}\|_F\|L_W^2\Delta_W\Sigma_\star^\frac{1}{2}\|_F + \delta_2(1-\eta)\|\Delta_H\Sigma_\star ^{\frac{1}{2}}\|_F\|L_W^2\Delta_W\Sigma_\star^\frac{1}{2}\|_F\\
				&\leq \frac{\delta_1(1-\eta)}{2}\big(\|\Delta_W\Sigma_\star ^\frac{1}{2}\|_F^2 + \|L_W^2\Delta_W\Sigma_\star^\frac{1}{2}\|_F^2\big) + \frac{\delta_2(1-\eta)}{2}\big(\|\Delta_H\Sigma_\star ^{\frac{1}{2}}\|_F^2 + \|L_W^2\Delta_W\Sigma_\star^\frac{1}{2}\|_F^2\big)\\
				&\leq \frac{\delta_1(1-\eta)}{2}\|\Delta_W\Sigma_\star ^\frac{1}{2}\|_F^2 + \frac{\delta_2(1-\eta)}{2}\|\Delta_H\Sigma_\star ^{\frac{1}{2}}\|_F^2 + \frac{(\delta_1+\delta_2)(1-\eta)}{2} \|L_W^2\Delta_W\Sigma_\star^\frac{1}{2}\|_F^2.
			\end{aligned}
		\end{equation}
		
		3. Controlling $ \mathfrak{R}_3 $ and $ \mathfrak{R}_4 $:\\	
		The bounds of $ \mathfrak{R}_3 $ and $ \mathfrak{R}_4 $ can be obtained by a similar argument for controlling $ \mathfrak{R}_2 $ (i.e. repeatedly using Lemmas \ref{lemma proof1} and \ref{lemma proof2}). Due to page limitations, we summarize these results in the following lemma.
		\begin{lemma}[Controlling $ \mathfrak{R}_3 $ and $ \mathfrak{R}_4 $]\label{bounding R3 R4}
			Under the event $ \mathcal{E} $, $ \mathfrak{R}_3 $ and $ \mathfrak{R}_4 $ can be controlled by 
			\begin{equation}\label{R3R4}
				\begin{aligned}
					|\mathfrak{R}_3| &\leq  (1+\beta)^3\frac{\delta_2}{2}\|\Delta_W\Sigma_\star^{\frac{1}{2}}\|_F^2 + (1+\beta)^3(\delta_1 + \frac{\delta_2}{2})\|\Delta_H\Sigma_\star^{\frac{1}{2}}\|_F^2,\\
					\mathfrak{R}_4 & \leq (1+\beta)^3\delta_1(\delta_1+\delta_2)\|\Delta_W\Sigma_\star ^{\frac{1}{2}}\|_F^2 + (1+\beta)^3\delta_2(\delta_1+\delta_2)\|\Delta_H\Sigma_\star ^{\frac{1}{2}}\|_F^2.
				\end{aligned}
			\end{equation}	
		\end{lemma}
		
		4. Combination:
		Combining the bounds for $ \mathfrak{R}_1 $, $ \mathfrak{R}_2 $, $ \mathfrak{R}_3 $ and $ \mathfrak{R}_4 $, we can obtain 
		\begin{equation*}
			\begin{aligned}
				&\|L_W^{1/2}(\widetilde{W}_{t+1}Q_t-W_\star)\Sigma_\star^{1/2}\|_F^2
				\leq \big\|L_W^{\frac{1}{2}}\Delta_W\Sigma_\star^{\frac{1}{2}}\big\|_F^2  + \Big(\eta^2- 2\eta^2\frac{\epsilon}{1-\epsilon}\Big) \big\|L_W^{\frac{3}{2}}\Delta_W\Sigma_\star^{\frac{1}{2}}\big\|_F^2   \\
				& ~~~~~~~~~~+\Big(2\eta\frac{\epsilon}{1-\epsilon}-2\eta\Big)\big\|L_W\Delta_W\Sigma_\star^{\frac{1}{2}}\big\|_F^2  + \eta^2(1+\beta)^\frac{3}{2} \frac{2\epsilon+\epsilon^2}{(1-\epsilon)^2}\big\|\Delta_H\Sigma_\star^{\frac{1}{2}}\big\|_F^2  \\
				&~~~~~~~~~~+\eta(1-\eta)\big( \delta_1\|\Delta_W\Sigma_\star ^\frac{1}{2}\|_F^2 + \delta_2\|\Delta_H\Sigma_\star ^{\frac{1}{2}}\|_F^2 + (\delta_1+\delta_2) \|L_W^2\Delta_W\Sigma_\star^\frac{1}{2}\|_F^2  \big)\\
				&~~~~~~~~~~+\eta^2(1+\beta)^3\big(  \delta_2\|\Delta_W\Sigma_\star^{\frac{1}{2}}\|_F^2 + (2\delta_1 + \delta_2)\|\Delta_H\Sigma_\star^{\frac{1}{2}}\|_F^2 \big)\\
				&~~~~~~~~~~+\eta^2(1+\beta)^3\big( \delta_1(\delta_1+\delta_2)\|\Delta_W\Sigma_\star ^{\frac{1}{2}}\|_F^2 + \delta_2(\delta_1+\delta_2)\|\Delta_H\Sigma_\star ^{\frac{1}{2}}\|_F^2 \big).
			\end{aligned}
		\end{equation*}
		A similar bound holds for the second term of (\ref{error metric comletion}). Consequently, 	denoting $ \alpha := 1+\beta $, we can obtain
		\begin{equation*}
			\begin{aligned}
				&\|L_W^{1/2}(\widetilde{W}_{t+1}Q_t-W_\star)\Sigma_\star^{1/2}\|_F^2 + \|L_H^{1/2}(\widetilde{H}_{t+1}Q_t^{-T}-H_\star)\Sigma_\star^{1/2}\|_F^2\\
				&\leq \big\{\big\|L_W^{\frac{1}{2}}\Delta_W\Sigma_\star^{\frac{1}{2}}\big\|_F^2 + \big\|L_H^{\frac{1}{2}}\Delta_H\Sigma_\star^{\frac{1}{2}}\big\|_F^2\big\} + \Big(\eta^2- 2\eta^2\frac{\epsilon}{1-\epsilon}\Big) \big\{\big\|L_W^{\frac{3}{2}}\Delta_W\Sigma_\star^{\frac{1}{2}}\big\|_F^2 + \big\|L_H^{\frac{3}{2}}\Delta_H\Sigma_\star^{\frac{1}{2}}\big\|_F^2\big\}
			\end{aligned}
		\end{equation*}
		\begin{equation*}
			\begin{aligned}
				&~~~~~~~~~~~~~~~~-2\eta\frac{1-2\epsilon}{1-\epsilon}\big\{\big\|L_W\Delta_W\Sigma_\star^{\frac{1}{2}}\big\|_F^2 + \big\|L_H\Delta_H\Sigma_\star^{\frac{1}{2}}\big\|_F^2\big\} + \eta^2(1+\beta)^\frac{3}{2} \frac{2\epsilon+\epsilon^2}{(1-\epsilon)^2}\big\{\big\|\Delta_W\Sigma_\star^{\frac{1}{2}}\big\|_F^2 + \big\|\Delta_H\Sigma_\star^{\frac{1}{2}}\big\|_F^2\big\}\\
				&~~~~~~~~~~~~~~~~+\eta(1-\eta)(\delta_1+\delta_2)\big\{\big\|L_W^2\Delta_W\Sigma_\star^{\frac{1}{2}}\big\|_F^2 + \big\|L_H^2\Delta_H\Sigma_\star^{\frac{1}{2}}\big\|_F^2\big\}\\
				&~~~~~~~~~~~~~~~~+\Big(  \eta(1-\eta)(\delta_1+\delta_2)+\eta^2(1+\beta)^3(2\delta_1+2\delta_2)  + \eta^2(1+\beta)^3(\delta_1+\delta_2)^2 \Big)\big\{\big\|\Delta_W\Sigma_\star^{\frac{1}{2}}\big\|_F^2 + \big\|\Delta_H\Sigma_\star^{\frac{1}{2}}\big\|_F^2\big\}\\
				&~~~~~~~~~~~~~\leq \text{dist}(F_t,F_\star)^2 + \alpha^2\eta^2\frac{1-3\epsilon}{1-\epsilon}\text{dist}(F_t,F_\star)^2 -2\alpha\eta\frac{1-2\epsilon}{1-\epsilon}\text{dist}(F_t,F_\star)^2 \\
				&~~~~~~~~~~~~~~~~ +\eta^2(1+\beta)^\frac{3}{2} \frac{2\epsilon+\epsilon^2}{(1-\epsilon)^2} \text{dist}(F_t,F_\star)^2 +\alpha^3 \eta(1-\eta)(\delta_1+\delta_2)\text{dist}(F_t,F_\star)^2 \\
				&~~~~~~~~~~~~~~~~+ \Big(  \eta(1-\eta)(\delta_1+\delta_2)+\eta^2(1+\beta)^3(2\delta_1+2\delta_2)  + \eta^2(1+\beta)^3(\delta_1+\delta_2)^2 \Big)\text{dist}(F_t,F_\star)^2\\
				&~~~~~~~~~~~~~= \rho^2(\epsilon;\eta;\beta)\text{dist}(F_t,F_\star)^2,
			\end{aligned}
		\end{equation*}
		where $ \rho^2(\epsilon;\eta;\beta)   $  is the contraction rate defined as
		\begin{equation}\label{rho completion}
			\begin{aligned}
				\rho^2(\epsilon;\eta;\beta) &:= 1+\frac{1-3\epsilon}{1-\epsilon}\alpha^2\eta^2 - \frac{1-2\epsilon}{1-\epsilon}2\alpha\eta + \frac{2\epsilon+\epsilon^2}{(1-\epsilon)^2}(1+\beta)^\frac{3}{2} \eta^2 + \alpha^3(\delta_1+\delta_2) \eta(1-\eta) \\
				&~~~+ (\delta_1+\delta_2) \eta(1-\eta) + (2\delta_1+2\delta_2)(1+\beta)^3\eta^2 + (\delta_1+\delta_2)^2(1+\beta)^3\eta^2.
			\end{aligned}
		\end{equation}
		It is easy to verified that the definition of $ \delta_1 $ and $ \delta_2 $  guarantees that as long as $ p\geq C\big(\mu r \kappa^4  \vee  \frac{\log(m\vee n)}{1+\beta}    \big)\mu r/(m\wedge n) $ for some sufficiently large constant $ C $, one has $ \delta_1 + \delta_2 \leq 0.1(1+\beta) $.  When  $ 0 < \beta \leq 1 $ and $ 0<\eta\leq \frac{2}{2(1+\beta)+\sqrt{(1+\beta)}} $, it can be further verified that $ \rho^2(\epsilon;\eta;\beta) \leq (1 - \gamma \eta)^2 $ for $ \gamma = -0.96 \beta^2  + 0.35\beta+0.63 $. Thus we can obtain that 
		\begin{equation}\label{first conclusion proof}
			\text{dist}(\widetilde{F}_{t+1},F_\star) \leq \sqrt{\|L_W^{1/2}(\widetilde{W}_{t+1}Q_t-W_\star)\Sigma_\star^{1/2}\|_F^2 + \|L_H^{1/2}(\widetilde{H}_{t+1}Q_t^{-T}-H_\star)\Sigma_\star^{1/2}\|_F^2} \leq (1-\gamma\eta)\text{dist}(F_t,F_\star).
		\end{equation}
		
		Next, we demonstrate the conclusion
		$ \|W_{t+1}H_{t+1}^T - X_\star\|_F \leq 1.5\text{dist}(F_{t+1}, F_\star) $ in the following. Actually, for any $ W\in \mathbb{R}^{m\times r} $, $ H\in \mathbb{R}^{n\times r} $, $ \Delta_W = W-W_\star $ and $ \Delta_H = H-H_\star $, we have
		\begin{equation*}
			\begin{aligned}
				\|WH^T - X_\star\|_F \leq \|\Delta_WH_\star^T\|_F + \|\Delta_HW_\star^T\|_F + \|\Delta_W\Delta_H^T\|_F 
				= \|\Delta_W\Sigma_\star^{1/2}\|_F + \|\Delta_H\Sigma_\star^{1/2}\|_F + \|\Delta_W\Delta_H^T\|_F,
			\end{aligned}
		\end{equation*}
		where the last term can be further bounded by
		\begin{equation}\label{similar 1}
			\begin{aligned}
				\|\Delta_W\Delta_H^T\|_F 
				&= \frac{1}{2}\|\Delta_W\Sigma_\star^{1/2}(\Delta_H\Sigma_\star^{-1/2})^T\|_F + \frac{1}{2}\|\Delta_W\Sigma_\star^{-1/2}(\Delta_H\Sigma_\star^{1/2})^T\|_F\\
				&\leq \frac{1}{2}\|\Delta_W\Sigma_\star^{1/2}\|_F\|\Delta_H\Sigma_\star^{-1/2}\|_\text{op} + \frac{1}{2}\|\Delta_H\Sigma_\star^{1/2}\|_F\|\Delta_W\Sigma_\star^{-1/2}\|_\text{op}\\
				&\leq \frac{1}{2} (\|\Delta_H\Sigma_\star^{-1/2}\|_\text{op} \vee \|\Delta_W\Sigma_\star^{-1/2}\|_\text{op}) (\|\Delta_W\Sigma_\star^{1/2}\|_F + \|\Delta_H\Sigma_\star^{1/2}\|_F).
			\end{aligned}
		\end{equation}
		Substituting $ W = W_{t+1} $, $ H = H_{t+1} $ into the above formulas and considering $ \|\Delta_H\Sigma_\star^{-1/2}\|_\text{op} \vee \|\Delta_W\Sigma_\star^{-1/2}\|_\text{op} \leq \epsilon $, we have
		\begin{equation}\label{similar 2}
			\begin{aligned}
				&\|W_{t+1}H_{t+1}^T - X_\star\|_F \leq (1+\frac{\epsilon}{2}) (\|\Delta_W\Sigma_\star^{1/2}\|_F + \|\Delta_H\Sigma_\star^{1/2}\|_F)
				\leq  (1+\frac{\epsilon}{2})\sqrt{2(\|\Delta_W\Sigma_\star^{1/2}\|_F^2 + \|\Delta_H\Sigma_\star^{1/2}\|_F^2)}\\
				&~~~~~~~~~~\leq  (1+\frac{\epsilon}{2})\sqrt{2(\|L_W^{1/2}\Delta_W\Sigma_\star^{1/2}\|_F^2 + \|L_H^{1/2}\Delta_H\Sigma_\star^{1/2}\|_F^2)}
				= (1+\frac{\epsilon}{2})\sqrt{2}\text{dist}(F_{t+1}, F_\star)
				\leq 1.5\text{dist}(F_{t+1}, F_\star).
			\end{aligned}
		\end{equation}
		
		Now all the conclusions of Theorem 2 can be guaranteed, and we complete the proof.
		\Halmos 	
	\end{proof}

	\section{Proof of Theorem 3}
	\begin{proof}{Proof of Theorem 3}
		To begin with, we give the following two useful lemmas.
		\begin{lemma}\label{initialization lemma1}
			For any factor matrix $ F := [W^T, H^T]^T \in \mathbb{R}^{(m+n)\times r}$, the distance between $ F $ and $ X_
			\star $ is bounded by
			\begin{equation*}
				\text{dist}(F, X_\star) \leq \sqrt{(1+\beta)(\sqrt{2}+1)}\|WH^T - X_\star\|_F.
			\end{equation*}
		\end{lemma}
		\begin{lemma}\label{initialization lemma2}
			For any fixed $ X\in\mathbb{R}^{m\times n} $, suppose $ G_1 $, $ G_2 $ are $ \psi $-smooth on it, then with overwhelming probability, one has
			\begin{equation*}
				\begin{aligned}
					\|(p^{-1}\mathcal{A}\mathcal{P}_\Omega \mathcal{B} - \mathcal{I})(X)\|_\text{op} 
					\leq C_0\frac{\log(m\vee n)}{p}\|X\|_\infty + C_0\sqrt{\frac{\log(m\vee n)}{p}}(\|X\|_{2,\infty} \vee \|X^T\|_{2,\infty}) + \sqrt{\frac{\psi r}{m \wedge n}}\|X\|_\text{op}.
				\end{aligned}
			\end{equation*}
		\end{lemma}
		Due to that the matrix $ U_0\Sigma_0V_0^T - X_\star $ has rank at most $ 2r $, Lemma \ref{initialization lemma1} ensures that
		\begin{equation}\label{inequality1}
			\begin{aligned}
				\text{dist}(\widetilde{F}_0, X_\star) \leq \sqrt{(1+\beta)(\sqrt{2}+1)}\|U_0\Sigma_0V_0^T - X_\star \|_F 
				\leq \sqrt{(1+\beta)(\sqrt{2}+1)2r}\|U_0\Sigma_0V_0^T - X_\star \|_\text{op}.
			\end{aligned}
		\end{equation}
		Considering that $ U_0\Sigma_0V_0^T $ is the best rank-$ r $ approximation to $ p^{-1}\mathcal{A}\mathcal{P}_\Omega(X_\star) \mathcal{B} $, we have 
		\begin{equation*}
			\|p^{-1}\mathcal{A}\mathcal{P}_\Omega(X_\star) \mathcal{B} - U_0\Sigma_0V_0^T\|_\text{op}\leq \|p^{-1}\mathcal{A}\mathcal{P}_\Omega(X_\star) \mathcal{B} - X_\star\|_\text{op},
		\end{equation*}
		leading to the following inequality:
		\begin{equation}\label{inequality2}
			\begin{aligned}
				\|U_0\Sigma_0V_0^T-X_\star\|_\text{op} &\leq \|p^{-1}\mathcal{A}\mathcal{P}_\Omega(X_\star) \mathcal{B} - U_0\Sigma_0V_0^T\|_\text{op} + \|p^{-1}\mathcal{A}\mathcal{P}_\Omega(X_\star) \mathcal{B} - X_\star\|_\text{op} \\
				&\leq 2\|p^{-1}\mathcal{A}\mathcal{P}_\Omega(X_\star) \mathcal{B} - X_\star\|_\text{op}.
			\end{aligned}
		\end{equation}
		Combining (\ref{inequality1}) and (\ref{inequality2}) gives that 
		\begin{equation}\label{inequality4}
			\begin{aligned}
				\text{dist}(\widetilde{F}_0, X_\star) \leq 2\sqrt{(1+\beta)(\sqrt{2}+1)2r}\|p^{-1}\mathcal{A}\mathcal{P}_\Omega(X_\star) \mathcal{B} - X_\star\|_\text{op} \leq 5\sqrt{(1+\beta)r}\|(p^{-1}\mathcal{A}\mathcal{P}_\Omega \mathcal{B} -\mathcal{I}) (X_\star)\|_\text{op} .
			\end{aligned}
		\end{equation}
		The graph incoherence assumption of $ X_\star $ gives the following bounds:
		\begin{equation*}
			\begin{aligned}
				\|X_\star\|_\infty &\leq \|L_W^\frac{1}{2}U_\star\|_{2,\infty}\|\Sigma_\star\|_\text{op}\|L_H^\frac{1}{2}V_\star\|_{2,\infty} \leq \frac{\mu r}{\sqrt{mn}}\kappa\sigma_r(X_\star),\\
				\|X_\star\|_{2,\infty} &\leq \|L_W^\frac{1}{2}U_\star\|_{2,\infty}\|\Sigma_\star\|_\text{op}\|L_H^\frac{1}{2}V_\star\|_\text{op} \leq \sqrt{\frac{\mu r}{m}}\kappa\sigma_r(X_\star),\\
				\|X_\star^T\|_{2,\infty} &\leq \|L_W^\frac{1}{2}U_\star\|_\text{op}\|\Sigma_\star\|_\text{op}\|L_H^\frac{1}{2}V_\star\|_{2,\infty} \leq \sqrt{\frac{\mu r}{n}}\kappa\sigma_r(X_\star),
			\end{aligned}
		\end{equation*}
		based on which Lemma \ref{initialization lemma2} ensures that, with overwhelming probability, we have 
		\begin{equation}\label{inequality3}
			\begin{aligned}
				&\|(p^{-1}\mathcal{A}\mathcal{P}_\Omega \mathcal{B} - \mathcal{I})(X_\star)\|_\text{op} \\
				&\leq C_0\frac{\log(m\vee n)}{p}\|X_\star\|_\infty + C_0\sqrt{\frac{\log(m\vee n)}{p}}(\|X_\star\|_{2,\infty} \vee \|X_\star^T\|_{2,\infty}) + \sqrt{\frac{\psi r}{m \wedge n}}\|X_\star\|_\text{op}\\
				& \leq \Big(C_0\frac{\log(m\vee n)}{p} \frac{\mu r}{\sqrt{mn}} + C_0\sqrt{\frac{\log(m\vee n)}{p}} \sqrt{\frac{\mu r}{m\wedge n}} + \sqrt{\frac{\psi r}{m \wedge n}}\Big)\kappa\sigma_r(X_\star)\\
				& \leq C\Big(\frac{\mu r\log(m\vee n)}{p\sqrt{mn}} + \sqrt{\frac{\mu r\log(m\vee n)}{p(m\wedge n)}} + \sqrt{\frac{\psi r}{p(m \wedge n)}}\Big)\kappa\sigma_r(X_\star).
			\end{aligned}
		\end{equation}
		Combining (\ref{inequality4}) and (\ref{inequality3}) gives the conclusion of Theorem 3.
		\Halmos 	
	\end{proof}

\newpage
	\begin{center}
		\textbf{\Large Proofs of Technical Lemmas}
	\end{center}
\vspace{0.5cm}
	\begin{flushleft}
		{\fontsize{11pt}{13pt}\selectfont\textbf{ Proof of Lemma \ref{Q exist}}}
	\end{flushleft}
	\begin{proof}{Proof of Lemma \ref{Q exist}}
		According to the definition of $ \text{dist}(F, F_\star) $,
		it is straightforward to verify that if $ \text{dist}(F, F_\star) < \sigma_r(X_\star) $, then there must exist a matrix $ \bar{Q}\in \text{GL}(r) $ such that
		\begin{equation*}
			\sqrt{ \|L_W^{1/2}(W\bar{Q}-W_\star)\Sigma_\star^{1/2}\|_F^2 + \|L_H^{1/2}(H\bar{Q}^{-T}-H_\star)\Sigma_\star^{1/2}\|_F^2} \leq \varepsilon \sigma_r(X_\star)
		\end{equation*}
		for some small $ \varepsilon $ satisfying $ 0< \varepsilon<1 $, which further leads to the following inequality:
		\begin{equation*}
			\|L_W^{1/2}(W\bar{Q}-W_\star)\Sigma_\star^{-1/2}\|_\text{op} \vee \|L_H^{1/2}(H\bar{Q}^{-T}-H_\star)\Sigma_\star^{-1/2}\|_\text{op} \leq \varepsilon.
		\end{equation*}
		Then Weyl's inequality $ \lvert \sigma_r(A) - \sigma_r(B) \rvert \leq \|A - B\|_\text{op} $ tells us that 
		\begin{equation}
			\sigma_r(L_W^{1/2}W\bar{Q}\Sigma_\star^{-1/2}) \geq \sigma_r(L_W^{1/2}W_\star\Sigma_\star^{-1/2}) - \|L_W^{1/2}(W\bar{Q}-W_\star)\Sigma_\star^{-1/2}\|_\text{op}
			\geq \sigma_r(L_W^{1/2}U_\star) - \varepsilon.
		\end{equation}
		Notice that $ \sigma_r(L_W^{1/2}U_\star)   \geq 1 $, we can therefore get 
		\begin{equation}\label{Q exist 1}
			\sigma_r(L_W^{1/2}W\bar{Q}\Sigma_\star^{-1/2}) \geq 1- \varepsilon.
		\end{equation} 
		On the basis of $ \bar{Q} $, we further introduce a new matrix $ P $ considering the following optimization problem:
		\begin{equation}\label{P introduce}
			\inf_{P\in \text{GL}(r)} \|L_W^{1/2}(W\bar{Q}P-W_\star)\Sigma_\star^{1/2}\|_F^2 + \|L_H^{1/2}(H\bar{Q}^{-T}P^{-T}-H_\star)\Sigma_\star^{1/2}\|_F^2.
		\end{equation}
		It is easy to verify that if the minimum of the above problem is attained at some $ P $, then $ \bar{Q}P $ must be the optimal alignment matrix between $ F $ and $ F_\star $, i.e., the existence of $ Q $ is guaranteed. Next we concentrate on demonstrating that the minimum of optimization problem (\ref{P introduce}) is attained at some $ P $.
		
		It is straightforward to see that
		\begin{equation}
			\begin{aligned}
				&\inf_{P\in \text{GL}(r)} \|L_W^{1/2}(W\bar{Q}P-W_\star)\Sigma_\star^{1/2}\|_F^2 + \|L_H^{1/2}(H\bar{Q}^{-T}P^{-T}-H_\star)\Sigma_\star^{1/2}\|_F^2\\
				& \leq \|L_W^{1/2}(W\bar{Q}-W_\star)\Sigma_\star^{1/2}\|_F^2 + \|L_H^{1/2}(H\bar{Q}^{-T}-H_\star)\Sigma_\star^{1/2}\|_F^2,
			\end{aligned}
		\end{equation}
		then for any $ \bar{Q}P $ achieving a smaller distance than $ \bar{Q} $, $ P $ must obey
		\begin{equation}
			\sqrt{\|L_W^{1/2}(W\bar{Q}P-W_\star)\Sigma_\star^{1/2}\|_F^2 + \|L_H^{1/2}(H\bar{Q}^{-T}P^{-T}-H_\star)\Sigma_\star^{1/2}\|_F^2} \leq \varepsilon \sigma_r(X_\star),
		\end{equation}
		which further implies that 
		\begin{equation}
			\|L_W^{1/2}(W\bar{Q}P-W_\star)\Sigma_\star^{-1/2}\|_\text{op} \vee \|L_H^{1/2}(H\bar{Q}^{-T}P^{-T}-H_\star)\Sigma_\star^{-1/2}\|_\text{op} \leq \varepsilon.
		\end{equation}
		Then Weyl's inequality $ \lvert \sigma_1(A) - \sigma_1(B) \rvert \leq \|A - B\|_\text{op} $ tells us that 
		\begin{equation}\label{Q exist 2}
			\begin{aligned}
				\sigma_1(L_W^{1/2}W\bar{Q}P\Sigma_\star^{-1/2}) &\leq 	\sigma_1(L_W^{1/2}W_\star\Sigma_\star^{-1/2}) + \|L_W^{1/2}(W\bar{Q}P-W_\star)\Sigma_\star^{-1/2}\|_\text{op} \\
				& \leq \sqrt{\sigma_1(U_\star^TL_WU_\star)} + \varepsilon = \sqrt{1+\beta} + \varepsilon.
			\end{aligned}
		\end{equation}
		Invoking the relation $ \sigma_r(A)\sigma_1(B) \leq \sigma_1(AB) $, we can get
		\begin{equation}
			\sigma_r(L_W^{1/2}W\bar{Q}\Sigma_\star^{-1/2})\sigma_1(\Sigma_\star^{1/2}P\Sigma_\star^{-1/2}) \leq \sigma_1(L_W^{1/2}W\bar{Q}P\Sigma_\star^{-1/2}),
		\end{equation}
		which implies that $ \sigma_1(\Sigma_\star^{1/2}P\Sigma_\star^{-1/2}) \leq \frac{\sqrt{1+\beta} + \varepsilon}{1 - \varepsilon} $. Similarly, we can also get $ \sigma_1(\Sigma_\star^{1/2}P^{-T}\Sigma_\star^{-1/2}) \leq  \frac{\sqrt{1+\beta} + \varepsilon}{1 - \varepsilon} $, which is equivalent to $ \sigma_r(\Sigma_\star^{1/2}P\Sigma_\star^{-1/2}) \geq \frac{1 - \varepsilon}{\sqrt{1+\beta}+ \varepsilon} $. Consequently, the problem (\ref{P introduce}) is equivalent to the following constrained optimization problem:
		\begin{equation}
			\begin{aligned}
				&\min_{P\in \text{GL}(r)} \|L_W^{1/2}(W\bar{Q}P-W_\star)\Sigma_\star^{1/2}\|_F^2 + \|L_H^{1/2}(H\bar{Q}^{-T}P^{-T}-H_\star)\Sigma_\star^{1/2}\|_F^2 \\
				& ~~~\text{s.t.} \frac{1 - \varepsilon}{\sqrt{1+\beta}+ \varepsilon} \leq \sigma_r(\Sigma_\star^{1/2}P\Sigma_\star^{-1/2}) \leq \sigma_1(\Sigma_\star^{1/2}P\Sigma_\star^{-1/2}) \leq \frac{\sqrt{1+\beta} + \varepsilon}{1 - \varepsilon},
			\end{aligned}
		\end{equation}
		which is a continuous optimization problem over a compact set, and thus the Weierstrass extreme value theorem guarantees the existence of $ P $. 
		The proof is now completed. 
		\Halmos
	\end{proof}
	
	\vspace{0.2cm}
	\begin{flushleft}
		{\fontsize{11pt}{13pt}\selectfont\textbf{ Proof of Lemma \ref{useful bounds}}}
	\end{flushleft}
	\begin{proof}{Proof of Lemma \ref{useful bounds}}
		First, bounds (\ref{eq:1b}) and the first part of (\ref{eq:1d}) are exactly the existing consequences. The second part of(\ref{eq:1d}) can be easily obtained by:
		\begin{equation*}
			\begin{aligned}
				&\|\Sigma_\star^{1/2}(H^TH)^{-1}\Sigma_\star^{1/2}\|_\text{op}  \\
				&= \|\Sigma_\star^{1/2}(H^TH)^{-1}H^TH(H^TH)^{-1}\Sigma_\star^{1/2}\|_\text{op} \\
				&= \|H(H^TH)^{-1}\Sigma_\star^{1/2}\|_\text{op}^2 \\
				&\leq \frac{1}{(1-\epsilon)^2}.
			\end{aligned}
		\end{equation*}
		From the following derivation
		\begin{equation*}
			\begin{aligned}
				&\|L_W^{\frac{1}{2}}WH^T\|_{2,\infty} \\
				&\geq \sigma_r(H\Sigma_\star^{-\frac{1}{2}})	\|L_W^{\frac{1}{2}}W\Sigma_\star^{\frac{1}{2}}\|_{2,\infty}\\
				&\geq \big(\sigma_r(H_\star\Sigma_\star^{-\frac{1}{2}}) - \|\Delta_H\Sigma_\star^{-\frac{1}{2}}\|_\text{op}\big)\|L_W^{\frac{1}{2}}W\Sigma_\star^{\frac{1}{2}}\|_{2,\infty}\\
				& \geq 
				(1-\epsilon)\|L_W^{\frac{1}{2}}W\Sigma_\star^{\frac{1}{2}}\|_{2,\infty},
			\end{aligned}
		\end{equation*}
		we can get that
		\begin{equation*}
			\|L_W^{\frac{1}{2}}W\Sigma_\star^{\frac{1}{2}}\|_{2,\infty} \leq \frac{1}{1-\epsilon}\|L_W^{\frac{1}{2}}WH^T\|_{2,\infty} \leq \frac{\sqrt{1+\beta}}{(1-\epsilon)\sqrt{m}}C_B\sqrt{\mu r}\sigma_1(X_\star).
		\end{equation*}
		Similarly, we can also get 
		\begin{equation*}
			\|L_H^{\frac{1}{2}}H\Sigma_\star^{\frac{1}{2}}\|_{2,\infty} \leq \frac{1}{1-\epsilon}\|L_H^{\frac{1}{2}}HW^T\|_{2,\infty} \leq \frac{\sqrt{1+\beta}}{(1-\epsilon)\sqrt{n}}C_B\sqrt{\mu r}\sigma_1(X_\star),
		\end{equation*}
		and thus (\ref{eq:1e}) can be obtained.
		Take (\ref{eq:1e})  together with the relation $ 	\|L_W^{\frac{1}{2}}W\Sigma_\star^{-\frac{1}{2}}\|_{2,\infty} \leq 	\|L_W^{\frac{1}{2}}W\Sigma_\star^{\frac{1}{2}}\|_{2,\infty} /\sigma_1(X_\star) $ and $ 	\|L_H^{\frac{1}{2}}H\Sigma_\star^{-\frac{1}{2}}\|_{2,\infty} \leq 	\|L_H^{\frac{1}{2}}H\Sigma_\star^{\frac{1}{2}}\|_{2,\infty} /\sigma_1(X_\star) $ to obtain 
		(\ref{eq:1f}). Finally, (\ref{eq:1g}) can be obtained by the following derivation
		\begin{equation*}
			\begin{aligned}
				&\sqrt{m}\|L_W^{\frac{1}{2}}\Delta_W\Sigma_\star^{-\frac{1}{2}}\|_{2,\infty}\\
				&\leq \sqrt{m}\big(\|L_W^{\frac{1}{2}}W\Sigma_\star^{-\frac{1}{2}}\|_{2,\infty} + \|L_W^{\frac{1}{2}}W_\star\Sigma_\star^{-\frac{1}{2}}\|_{2,\infty}\big)\\
				&\leq \sqrt{m}\bigg(\frac{\sqrt{1+\beta}}{(1-\epsilon)\sqrt{m}}C_B\sqrt{\mu r}\sigma_1(X_\star) + \|L_W^{\frac{1}{2}}U_\star\|_{2,\infty}\|\Sigma_\star\|_\text{op}\bigg)\\
				&\leq \sqrt{m}\bigg(\frac{\sqrt{1+\beta}}{(1-\epsilon)\sqrt{m}}C_B\sqrt{\mu r}\sigma_1(X_\star) + \frac{\sqrt{\mu r}}{\sqrt{m}}\sigma_1(X_\star)\bigg)\\
				&\leq \Big(1+ \frac{C_B\sqrt{1+\beta}}{1-\epsilon}\Big)\sqrt{\mu r}\sigma_1(X_\star)
			\end{aligned}
		\end{equation*}
		together with a similar bound obtained in the same way: 
		$$ \sqrt{n}\|L_H^{\frac{1}{2}}\Delta_H\Sigma_\star^{-\frac{1}{2}}\|_{2,\infty} \leq \Big(1+ \frac{C_B\sqrt{1+\beta}}{1-\epsilon}\Big)\sqrt{\mu r}\sigma_1(X_\star).
		$$
		
		Now we complete the proof of Lemma \ref{useful bounds}.
		\Halmos 
	\end{proof}
	
   \vspace{0.2cm}
	\begin{flushleft}
		{\fontsize{11pt}{13pt}\selectfont\textbf{ Proof of Lemma \ref{lemma Q obeys}}}
	\end{flushleft}
	\begin{proof}{Proof of Lemma \ref{lemma Q obeys}}
		According to the definition of optimal alignment matrix between $ F $ and $ F_\star $, $ Q $ has the following  form:
		\begin{equation*}
			Q := \arg\min_{Q\in \text{GL}(r)} \|L_W^{1/2}(WQ-W_\star)\Sigma_\star^{1/2}\|_F^2 + \|L_H^{1/2}(HQ^{-T}-H_\star)\Sigma_\star^{1/2}\|_F^2,
		\end{equation*}
		which is equivalent to
		\begin{equation*}
			Q = \arg\min_{Q\in \text{GL}(r)}\text{tr}\big((WQ-W_\star)^TL_W(WQ-W_\star)\Sigma_\star\big) + \text{tr}\big((HQ^{-T}-H_\star)^TL_H(HQ^{-T}-H_\star)\Sigma_\star\big).
		\end{equation*}
		According to the first order necessary condition, the gradient of the objective function with respect to $ Q $ is zero, i.e.,
		\begin{equation*}
			2W^TL_W(WQ-W_\star)\Sigma_\star - 2Q^{-T}\Sigma_\star (HQ^{-T}-H_\star)^TL_HH_Q^{-T} = 0,
		\end{equation*}
		which further implies that
		\begin{equation*}
			Q^TW^TL_W(WQ-W_\star)\Sigma_\star = \Sigma_\star(HQ^{-T}-H_\star)^TL_HH_Q^{-T}.
		\end{equation*}
		This completes the proof of Lemma \ref{lemma Q obeys}.
		\Halmos
	\end{proof}
	
	\vspace{0.2cm}
	\begin{flushleft}
		{\fontsize{11pt}{13pt}\selectfont\textbf{ Proof of Lemma \ref{bounding R3 R4}}}
	\end{flushleft}
	\begin{proof}{Proof of Lemma \ref{bounding R3 R4}}
		We derive the bounds of $ \mathfrak{R}_3 $ and $ \mathfrak{R}_4 $ respectively.
		1. Controlling $ \mathfrak{R}_3 $:\\	
		Plugging in the decomposition $ WH^T-X_\star = \Delta_WH^T + W_\star\Delta_H^T $, we can obtain that
		\begin{equation*}
			\begin{aligned}
				|\mathfrak{R}_3| &= \big|\text{tr}\big(L_W^{3/2} (p^{-1}\mathcal{P}_\Omega-\mathcal{I})(WH^T-X_\star)H(H^TH)^{-1}\Sigma_\star (H^TH)^{-1}H^T\Delta_H W_\star^T L_W^{3/2}\big)\big|\\
				&\leq \underbrace{\big|\text{tr}\big( (p^{-1}\mathcal{P}_\Omega-\mathcal{I})\Delta_WH^TH(H^TH)^{-1}\Sigma_\star (H^TH)^{-1}H^T\Delta_H W_\star^T L_W^{3}\big)\big|}_{\mathfrak{R}_3^{(\text{\romannumeral1})}}\\
				&~~~ + \underbrace{\big|\text{tr}\big( (p^{-1}\mathcal{P}_\Omega-\mathcal{I})W_\star\Delta_H^TH(H^TH)^{-1}\Sigma_\star (H^TH)^{-1}H^T\Delta_H W_\star^T L_W^{3}\big)\big|}_{\mathfrak{R}_3^{(\text{\romannumeral2})}}.
			\end{aligned}
		\end{equation*}
		For $ \mathfrak{R}_3^{(\text{\romannumeral1})} $, invoking Lemma \ref{lemma proof2} by $ W_A = \Delta_W\Sigma_\star^\frac{1}{2} $, $ H_A = H\Sigma_\star^{-\frac{1}{2}} $, $ W_B = L_W^3W_\star\Sigma_\star^{-\frac{1}{2}} $, $ H_B = H(H^TH)^{-1}\Sigma_\star (H^TH)^{-1}H^T\Delta_H\Sigma_\star^{\frac{1}{2}} $, we can get the following bound:
		\begin{equation*}
			\begin{aligned}
				\mathfrak{R}_3^{(\text{\romannumeral1})}
				&\leq C_2\sqrt{\frac{m \vee n}{p}} \|L_W^3W_\star\Sigma_\star^{-\frac{1}{2}} \|_{2,\infty}\|\Delta_W\Sigma_\star^\frac{1}{2}\|_F \|H(H^TH)^{-1}\Sigma_\star (H^TH)^{-1}H^T\Delta_H\Sigma_\star^{\frac{1}{2}}\|_F\|H\Sigma_\star^{-\frac{1}{2}}\|_{2,\infty}\\	
				&\leq C_2\sqrt{\frac{m \vee n}{p}} (1+\beta)^\frac{5}{2}\|L_W^\frac{1}{2}W_\star\Sigma_\star^{-\frac{1}{2}} \|_{2,\infty}\|\Delta_W\Sigma_\star^\frac{1}{2}\|_F \|L_H^\frac{1}{2}H\Sigma_\star^{-\frac{1}{2}}\|_{2,\infty}\|H(H^TH)^{-1}\Sigma_\star^{\frac{1}{2}}\|_\text{op}^2\|\Delta_H\Sigma_\star^{\frac{1}{2}}\|_F\\
				&\leq C_2\sqrt{\frac{m \vee n}{p}} (1+\beta)^\frac{5}{2}\sqrt{\frac{\mu r}{m}}\frac{\sqrt{1+\beta}}{\sqrt{n}(1-\epsilon)}\kappa C_B\sqrt{\mu r}\frac{1}{(1-\epsilon)^2}\|\Delta_W\Sigma_\star^{\frac{1}{2}}\|_F\|\Delta_H\Sigma_\star^{\frac{1}{2}}\|_F\\
				&\leq \frac{\mu r}{\sqrt{p(m \wedge n)}}\frac{C_2C_B\kappa(1+\beta)^3}{(1-\epsilon)^3}\|\Delta_W\Sigma_\star^{\frac{1}{2}}\|_F\|\Delta_H\Sigma_\star^{\frac{1}{2}}\|_F.
			\end{aligned}
		\end{equation*}
		For $ \mathfrak{R}_3^{(\text{\romannumeral2})} $, we can invoke Lemma \ref{lemma proof1} by $ W_A = W_B = 0 $, $ H_A = \Delta_H $, $ H_B = H(H^TH)^{-1}\Sigma_\star (H^TH)^{-1}H^T\Delta_H $, leading to the following bound:
		\begin{equation*}
			\begin{aligned}
				\mathfrak{R}_3^{(\text{\romannumeral2})} &= \big|\text{tr}\big( (p^{-1}\mathcal{P}_\Omega-\mathcal{I})W_\star\Delta_H^TH(H^TH)^{-1}\Sigma_\star (H^TH)^{-1}H^T\Delta_H W_\star^T L_W^{3}\big)\big|\\
				&\leq (1+\beta)^3\big|\text{tr}\big( (p^{-1}\mathcal{P}_\Omega-\mathcal{I})W_\star\Delta_H^TH(H^TH)^{-1}\Sigma_\star (H^TH)^{-1}H^T\Delta_H W_\star^T \big)\big|\\
				&\leq (1+\beta)^3C_1\sqrt{\frac{\mu r \log(m \vee n)}{p(m \wedge n)}}\|W_\star\Delta_H^T\|_F\|W_\star\Delta_H^TH(H^TH)^{-1}\Sigma_\star (H^TH)^{-1}H^T\|_F\\
				&\leq (1+\beta)^3C_1\sqrt{\frac{\mu r \log(m \vee n)}{p(m \wedge n)}}\|\Delta_H\Sigma_\star^\frac{1}{2}\|_F^2\|H(H^TH)^{-1}\Sigma_\star^\frac{1}{2}\|_\text{op}^2\\
				&\leq (1+\beta)^3C_1\sqrt{\frac{\mu r \log(m \vee n)}{p(m \wedge n)}}\frac{1}{(1-\epsilon)^2}\|\Delta_H\Sigma_\star^\frac{1}{2}\|_F^2.
			\end{aligned}
		\end{equation*}
		We then combine the bound of $ \mathfrak{R}_3^{(\text{\romannumeral1})} $ and $ \mathfrak{R}_3^{(\text{\romannumeral2})} $ to control $ |\mathfrak{R}_3|  $ as follows:
		\begin{equation*}
			\begin{aligned}
				|\mathfrak{R}_3| &\leq \frac{\mu r}{\sqrt{p(m \wedge n)}}\frac{C_2C_B\kappa(1+\beta)^3}{(1-\epsilon)^3}\|\Delta_W\Sigma_\star^{\frac{1}{2}}\|_F\|\Delta_H\Sigma_\star^{\frac{1}{2}}\|_F + C_1\sqrt{\frac{\mu r \log(m \vee n)}{p(m \wedge n)}}\frac{(1+\beta)^3}{(1-\epsilon)^2}\|\Delta_H\Sigma_\star^\frac{1}{2}\|_F^2\\
				& \leq  (1+\beta)^3\delta_1\|\Delta_H\Sigma_\star^\frac{1}{2}\|_F^2 + (1+\beta)^3\delta_2\|\Delta_W\Sigma_\star^{\frac{1}{2}}\|_F\|\Delta_H\Sigma_\star^{\frac{1}{2}}\|_F \\
				& \leq  (1+\beta)^3\delta_1\|\Delta_H\Sigma_\star^\frac{1}{2}\|_F^2 + (1+\beta)^3\frac{\delta_2}{2}\big(\|\Delta_W\Sigma_\star^{\frac{1}{2}}\|_F^2 + \|\Delta_H\Sigma_\star^{\frac{1}{2}}\|_F^2\big) \\
				&= (1+\beta)^3\frac{\delta_2}{2}\|\Delta_W\Sigma_\star^{\frac{1}{2}}\|_F^2 + (1+\beta)^3(\delta_1 + \frac{\delta_2}{2})\|\Delta_H\Sigma_\star^{\frac{1}{2}}\|_F^2.
			\end{aligned}
		\end{equation*}
		
		2. Controlling $ \mathfrak{R}_4 $:\\
		$ \sqrt{\mathfrak{R}_4} $ can be decomposed as
		\begin{equation*}
			\begin{aligned}
				\sqrt{\mathfrak{R}_4} &= \|L_W^{\frac{3}{2}} (p^{-1}\mathcal{P}_\Omega-\mathcal{I})(WH^T-X_\star)H(H^TH)^{-1}\Sigma_\star^{\frac{1}{2}}\|_F\\&
				\overset{(\text{\romannumeral1})}{\leq} \big|\text{tr}\big(L_W^{\frac{3}{2}} (p^{-1}\mathcal{P}_\Omega-\mathcal{I})(WH^T-X_\star)H(H^TH)^{-1}\Sigma_\star^{\frac{1}{2}}Z^T\big)\big|\\
				&\leq \underbrace{\big|\text{tr}\big( (p^{-1}\mathcal{P}_\Omega-\mathcal{I})\Delta_WH_\star^TH_\star(H^TH)^{-1}\Sigma_\star^{\frac{1}{2}}Z^TL_W^{\frac{3}{2}}\big)\big|}_{\mathfrak{R}_4^{(\text{\romannumeral1})}}+ \underbrace{\big|\text{tr}\big( (p^{-1}\mathcal{P}_\Omega-\mathcal{I})\Delta_WH_\star^T\Delta_H(H^TH)^{-1}\Sigma_\star^{\frac{1}{2}}Z^TL_W^{\frac{3}{2}}\big)\big|}_{\mathfrak{R}_4^{(\text{\romannumeral2})}}\\
				&~~~+ \underbrace{\big|\text{tr}\big( (p^{-1}\mathcal{P}_\Omega-\mathcal{I})W\Delta_H^TH(H^TH)^{-1}\Sigma_\star^{\frac{1}{2}}Z^TL_W^{\frac{3}{2}}\big)\big|}_{\mathfrak{R}_4^{(\text{\romannumeral3})}},
			\end{aligned}
		\end{equation*}
		where in inequality $ (\text{\romannumeral1}) $ we employ the variational representation of the Frobenius norm for some $ Z\in \mathbb{R}^{m\times r} $ obeying$  \|Z\|_F = 1 $. For the first term $ \mathfrak{R}_4^{(\text{\romannumeral1})} $, under the event $ \mathcal{E} $, invoking Lemma \ref{lemma proof1} by $ W_A = \Delta_W $, $ W_B = L_W^{\frac{3}{2}}Z\Sigma_\star^{\frac{1}{2}}(H^TH)^{-1} $, $ H_A = H_B = 0 $, we can get
		\begin{equation*}
			\begin{aligned}
				\mathfrak{R}_4^{(\text{\romannumeral1})} &\leq	C_1\sqrt{\frac{\mu r \log(m \vee n)}{p(m \wedge n)}}\|\Delta_WH_\star ^T\|_F\|H_\star(H^TH)^{-1}\Sigma_\star^{\frac{1}{2}}Z^TL_W^{\frac{3}{2}}\|_F\\
				& \leq C_1\sqrt{\frac{\mu r \log(m \vee n)}{p(m \wedge n)}}\|\Delta_W\Sigma_\star ^\frac{1}{2}\|_F\|\Sigma_\star^\frac{1}{2}(H^TH)^{-1}\Sigma_\star ^\frac{1}{2}\|_\text{op}\|Z^T\|_F\|L_W^{\frac{3}{2}}\|_\text{op}\\
				& \leq C_1\sqrt{\frac{\mu r \log(m \vee n)}{p(m \wedge n)}}\frac{(1+\beta)^\frac{3}{2}}{(1-\epsilon)^2}\|\Delta_W\Sigma_\star ^\frac{1}{2}\|_F.
			\end{aligned}
		\end{equation*}
		Invoking Lemma \ref{lemma proof2} by $ W_A = \Delta_W\Sigma_\star^\frac{1}{2} $, $ H_A = H_\star\Sigma_\star^{-\frac{1}{2}} $,  $ W_B = L_W^\frac{3}{2}Z $, $ H_B = \Delta_H(H^TH)^{-1}\Sigma_\star ^\frac{1}{2} $, $ \mathfrak{R}_4^{(\text{\romannumeral2})} $ can be controlled by:
		\begin{equation*}
			\begin{aligned}
				\mathfrak{R}_4^{(\text{\romannumeral2})} 	&\leq C_2\sqrt{\frac{m \vee n}{p}} \|\Delta_W\Sigma_\star^\frac{1}{2}\|_{2,\infty}\|L_W^\frac{3}{2}Z\|_F \|H_\star\Sigma_\star^{-\frac{1}{2}}\|_{2,\infty}\|\Delta_H(H^TH)^{-1}\Sigma_\star ^\frac{1}{2}\|_F\\
				&\leq C_2\sqrt{\frac{m \vee n}{p}} \|L_W^\frac{1}{2}\Delta_W\Sigma_\star^\frac{1}{2}\|_{2,\infty}\|L_W^\frac{3}{2}\|_\text{op}\|Z\|_F \|L_W^\frac{1}{2}V_\star\|_{2,\infty}\|\Delta_H\Sigma_\star ^{-\frac{1}{2}}\|_F\|\Sigma_\star ^\frac{1}{2}(H^TH)^{-1}\Sigma_\star ^\frac{1}{2}\|_\text{op}\\
				& \leq C_2\sqrt{\frac{m \vee n}{p}}\frac{1}{\sqrt{m}}\Big(1+ \frac{C_B\sqrt{1+\beta}}{1-\epsilon}\Big)\sqrt{\mu r}\sigma_1(X_\star)(1+\beta)^\frac{3}{2}\sqrt{\frac{\mu r}{n}}\|\Delta_H\Sigma_\star ^{\frac{1}{2}}\|_F\frac{1}{\sigma_r(X_\star)}\frac{1}{(1-\epsilon)^2}\\
				&\leq \frac{\mu r}{\sqrt{p(m \wedge n)}}\frac{C_2\kappa}{(1-\epsilon)^2}(1+\beta)^\frac{3}{2}\Big(1+ \frac{C_B\sqrt{1+\beta}}{1-\epsilon}\Big)\|\Delta_H\Sigma_\star ^{\frac{1}{2}}\|_F.
			\end{aligned}
		\end{equation*}
		For $ \mathfrak{R}_4^{(\text{\romannumeral3})} $, we can invoke Lemma \ref{lemma proof2} by $ W_A = W\Sigma_\star^{-\frac{1}{2}} $, $ H_A = \Delta_H\Sigma_\star^{\frac{1}{2}} $,  $ W_B = L_W^\frac{3}{2}Z  $, $ H_B = H(H^TH)^{-1}\Sigma_\star ^\frac{1}{2} $, leading to the following bound:
		\begin{equation*}
			\begin{aligned}
				\mathfrak{R}_4^{(\text{\romannumeral3})}	
				&\leq C_2\sqrt{\frac{m \vee n}{p}} \|W\Sigma_\star^{-\frac{1}{2}} \|_{2,\infty}\|L_W^\frac{3}{2}Z \|_F \|\Delta_H\Sigma_\star^{\frac{1}{2}}\|_F\|H(H^TH)^{-1}\Sigma_\star ^\frac{1}{2}\|_{2,\infty}\\
				&\leq C_2\sqrt{\frac{m \vee n}{p}} \|L_W^\frac{1}{2}W\Sigma_\star^{-\frac{1}{2}} \|_{2,\infty}\|L_W^\frac{3}{2}\|_\text{op}\|Z\|_F \|\Delta_H\Sigma_\star^{\frac{1}{2}}\|_F\|H\Sigma_\star ^{-\frac{1}{2}}\|_{2,\infty}\|\Sigma_\star ^\frac{1}{2}(H^TH)^{-1}\Sigma_\star ^\frac{1}{2}\|_\text{op}\\
				&\leq C_2\sqrt{\frac{m \vee n}{p}} \frac{1}{\sqrt{m}}\frac{\sqrt{1+\beta}}{1-\epsilon}\kappa C_B\sqrt{\mu r}(1+\beta)^\frac{3}{2} \|\Delta_H\Sigma_\star^{\frac{1}{2}}\|_F\frac{1}{\sqrt{n}}\frac{\sqrt{1+\beta}}{1-\epsilon}\kappa C_B\sqrt{\mu r}\frac{1}{(1-\epsilon)^2}\\
				&\leq \frac{\mu r}{\sqrt{p(m \wedge n)}}\frac{C_2C_B^2\kappa^2(1+\beta)^\frac{5}{2}}{(1-\epsilon)^4}\|\Delta_H\Sigma_\star ^{\frac{1}{2}}\|_F.
			\end{aligned}
		\end{equation*}
		
		Combining $ \mathfrak{R}_4^{(\text{\romannumeral1})} $, $ \mathfrak{R}_4^{(\text{\romannumeral2})} $ and $ \mathfrak{R}_4^{(\text{\romannumeral3})} $, we have
		\begin{equation}\label{R4 bound 1}
			\begin{aligned}
				\sqrt{\mathfrak{R}_4} &\leq C_1\sqrt{\frac{\mu r \log(m \vee n)}{p(m \wedge n)}}\frac{(1+\beta)^\frac{3}{2}}{(1-\epsilon)^2}\|\Delta_W\Sigma_\star ^\frac{1}{2}\|_F \\
				&~~~~~+ \frac{(1+\beta)^\frac{3}{2}\mu r}{\sqrt{p(m \wedge n)}}\bigg(\frac{C_2\kappa}{(1-\epsilon)^2}\Big(1+ \frac{C_B\sqrt{1+\beta}}{1-\epsilon}\Big) + \frac{C_2C_B^2\kappa^2(1+\beta)}{(1-\epsilon)^4}\bigg)\|\Delta_H\Sigma_\star ^{\frac{1}{2}}\|_F\\
				& = (1+\beta)^\frac{3}{2}\delta_1\|\Delta_W\Sigma_\star ^{\frac{1}{2}}\|_F + (1+\beta)^\frac{3}{2}\delta_2\|\Delta_H\Sigma_\star ^{\frac{1}{2}}\|_F,
			\end{aligned}
		\end{equation}	
		and thus we have 
		\begin{equation}\label{R4 bound 2}
			\mathfrak{R}_4 \leq (1+\beta)^3\delta_1(\delta_1+\delta_2)\|\Delta_W\Sigma_\star ^{\frac{1}{2}}\|_F^2 + (1+\beta)^3\delta_2(\delta_1+\delta_2)\|\Delta_H\Sigma_\star ^{\frac{1}{2}}\|_F^2.
		\end{equation}	
		Now we complete the proof of Lemma \ref{bounding R3 R4}. \Halmos
	\end{proof}
	
	\vspace{0.2cm}
	\begin{flushleft}
		{\fontsize{11pt}{13pt}\selectfont\textbf{ Proof of Lemma \ref{initialization lemma1}}}
	\end{flushleft}
	\begin{proof}{Proof of Lemma \ref{initialization lemma1}}
		This Lemma is a slight modification to the following lemma:
		\begin{lemma}[\citep{tong2021accelerating}, Lemma 24]\label{lemma proof3}
			For any factor matrix $ F := \begin{bmatrix}
				W \\ 
				H
			\end{bmatrix} \in \mathbb{R}^{(m+n)\times r}$, the following bound holds:
			\begin{equation*}
				\inf_{Q\in \text{GL}(r)} \|(WQ-W_\star)\Sigma_\star^{1/2}\|_F^2 + \|(HQ^{-T}-H_\star)\Sigma_\star^{1/2}\|_F^2 \leq (\sqrt{2}+1)\|WH^T - X_\star\|_F^2.
			\end{equation*}
		\end{lemma}
		Denote
		\begin{equation*}
			Q_1 := \arg\min_{Q\in \text{GL}(r)} \|(WQ-W_\star)\Sigma_\star^{1/2}\|_F^2 + \|(HQ^{-T}-H_\star)\Sigma_\star^{1/2}\|_F^2,
		\end{equation*}
		then Lemma \ref{lemma proof3} ensures that
		\begin{equation*}
			\sqrt{\|(WQ_1-W_\star)\Sigma_\star^{1/2}\|_F^2 + \|(HQ_1^{-T}-H_\star)\Sigma_\star^{1/2}\|_F^2} \leq \sqrt{(\sqrt{2}+1)}\|WH^T - X_\star\|_F. 
		\end{equation*}
		On this basis, it can be verified that 
		\begin{equation*}
			\begin{aligned}
				\text{dist}(F, X_\star) &=\inf_{Q\in \text{GL}(r)} \|L_W^{1/2}(WQ-W_\star)\Sigma_\star^{1/2}\|_F^2 + \|L_H^{1/2}(HQ^{-T}-H_\star)\Sigma_\star^{1/2}\|_F^2\\
				&\leq \sqrt{\|L_W^{1/2}(WQ_1-W_\star)\Sigma_\star^{1/2}\|_F^2 + \|L_H^{1/2}(HQ_1^{-T}-H_\star)\Sigma_\star^{1/2}\|_F^2}\\
				&\leq \sqrt{\|L_W\|_\text{op}\|(WQ_1-W_\star)\Sigma_\star^{1/2}\|_F^2 + \|L_H\|_\text{op} \|(HQ_1^{-T}-H_\star)\Sigma_\star^{1/2}\|_F^2}\\
				&\leq \sqrt{(1+\beta)(\|(WQ_1-W_\star)\Sigma_\star^{1/2}\|_F^2 +  \|(HQ_1^{-T}-H_\star)\Sigma_\star^{1/2}\|_F^2)}\\
				&\leq \sqrt{(1+\beta)(\sqrt{2}+1)}\|WH^T - X_\star\|_F,
			\end{aligned}
		\end{equation*}
		which is exactly the conclusion of Lemma \ref{initialization lemma1}.	
		\Halmos 
	\end{proof}
	
	\vspace{0.2cm}
	\begin{flushleft}
		{\fontsize{11pt}{13pt}\selectfont\textbf{ Proof of Lemma \ref{initialization lemma2}}}
	\end{flushleft}
	\begin{proof}{Proof of Lemma \ref{initialization lemma2}}
		We start by recording a useful lemma as follows:
		\begin{lemma}[\citep{chen2015incoherence}, Lemma 2; \citep{tong2021accelerating}, Lemma 37]\label{lemma proof4}
			For any fixed matrix $ X\in\mathbb{R}^{m\times n} $, with overwhelming probability, one has
			\begin{equation*}
				\begin{aligned}
					\|(p^{-1}\mathcal{P}_\Omega - \mathcal{I})(X)\|_\text{op} \leq C_0\frac{\log(m\vee n)}{p}\|X\|_\infty + C_0\sqrt{\frac{\log(m\vee n)}{p}}(\|X\|_{2,\infty} \vee \|X^T\|_{2,\infty}).
				\end{aligned}
			\end{equation*}
		\end{lemma}
		The condition $ G_1 $ and $ G_2 $ are $ \psi $-smooth on matrix $ X $ means that
		\begin{equation*}
			\frac{\|\mathcal{A}X\mathcal{B} - X
				\|_\text{op}}{\|X\|_\text{op}} \leq \sqrt{\frac{\psi r}{m\wedge n}},
		\end{equation*}
		based on which
		$ \|(p^{-1}\mathcal{A}\mathcal{P}_\Omega \mathcal{B} - \mathcal{I})(X)\|_\text{op} $ can be decomposed as follows:
		\begin{equation}\label{inequality5}
			\begin{aligned}
				&\|(p^{-1}\mathcal{A}\mathcal{P}_\Omega \mathcal{B} - \mathcal{I})(X)\|_\text{op}\\
				&\leq \|p^{-1}\mathcal{A}\mathcal{P}_\Omega(X) \mathcal{B}-\mathcal{A}X \mathcal{B} +\mathcal{A}X \mathcal{B} -X\|_\text{op}\\
				&\leq \|p^{-1}\mathcal{A}\mathcal{P}_\Omega(X) \mathcal{B}-\mathcal{A}X \mathcal{B} \|_\text{op} + \|\mathcal{A}X \mathcal{B} -X\|_\text{op}\\
				&\leq \|p^{-1}\mathcal{P}_\Omega(X) -X \|_\text{op} + \sqrt{\frac{\psi r}{m\wedge n}}\|X\|_\text{op}.			
			\end{aligned}
		\end{equation}
		Combining (\ref{inequality5}) and Lemma \ref{lemma proof4} gives the conclusion of Lemma \ref{initialization lemma2}.
		\Halmos 
	\end{proof} 
\end{APPENDICES}

%%%%%%%%%%%%%%%%%
\end{document}